\pdfoutput=1
\documentclass[11pt]{article}
\ifdefined\pdfinclusioncopyfonts \pdfinclusioncopyfonts=1 \fi
\usepackage{juice_preprint}
\usepackage{amsmath,amsfonts}
\usepackage{graphicx,booktabs,tabularx,xcolor,colortbl}
\usepackage{enumitem}
\usepackage{pifont}
\usepackage{placeins}
\usepackage{algorithm}
\usepackage{algorithmic}
\usepackage{hyperref}
\usepackage[most]{tcolorbox}
\usetikzlibrary{arrows.meta,shadings}

\definecolor{paperblue}{HTML}{0072B2}
\definecolor{papergray}{HTML}{F4F5F6}
\definecolor{sigcolor}{HTML}{1F6F8B}
\definecolor{dinocolor}{HTML}{B8761A}
\newcolumntype{R}{>{\raggedleft\arraybackslash\leavevmode}X}
\newcolumntype{C}{>{\centering\arraybackslash\leavevmode}X}
\newlength{\caseimg}
\newcommand{\qacell}[7]{%
\begin{minipage}[t]{0.238\linewidth}
\includegraphics[width=\linewidth]{#1}\par\vspace{3pt}
{\fontsize{6.8}{8}\selectfont\textsc{#2} \textcolor{black!55}{(#3)}\par}\vspace{1pt}
{\fontsize{7.2}{8.6}\selectfont\raggedright #4\par}\vspace{2pt}
{\fontsize{7.2}{8.6}\selectfont\raggedright\textcolor{sigcolor}{\textbf{SigLIP2:}}~#5\quad\textcolor{dinocolor}{\textbf{DINO:}}~#6\par
\textcolor{black!55}{Reference:~#7}\par}
\end{minipage}}
\newcommand{\jointcell}[3]{%
\begin{minipage}[t]{0.236\linewidth}
{\footnotesize\textbf{#2}\par}\vspace{2pt}
\includegraphics[width=\linewidth]{#1}\par\vspace{3pt}
{\fontsize{6.8}{8.1}\selectfont\raggedright #3\par}
\end{minipage}}

\newcommand{\codeurl}{\url{https://github.com/Ced3-han/JUICE-Code}}
\newcommand{\weightsurl}{\url{https://huggingface.co/collections/CedLJH/juice}}
\newcommand{\blogurl}{\url{https://ced3-han.github.io/JUICE-Website/}}

\makeatletter
\long\def\@makecaption#1#2{%
  \vskip\abovecaptionskip
  \begingroup\small\raggedright\textbf{#1}: #2\par\endgroup
  \vskip\belowcaptionskip}
\makeatother

\hypersetup{pdftitle={JUICE: Unified Multimodal Modeling in Pretrained Representation Spaces},
  pdfauthor={Jiahan Li, Huangyuan Su, Shengbang Tong, Shusheng Yang, Ruihan Guo, Suozhi Huang, Hexu Zhao, Pratyusha Sharma, Sherry Yang, Jiatao Gu, Rob Fergus, Kyunghyun Cho}}

\newcommand{\method}{JUICE}
\newcommand{\keytakeaway}[4]{{\sffamily\bfseries T#1.} & {\sffamily\bfseries #2}: #3 & {\color{paperblue}\S\ref{#4}}\\[1.5pt]}
\newtcolorbox{takeawaybox}{colback=papergray,colframe=paperblue,
  boxrule=0pt,leftrule=1.5pt,arc=1pt,left=5pt,right=5pt,top=2pt,bottom=2pt,
  before skip=4pt,after skip=4pt}

\newcommand{\smalltable}{\small\setlength{\tabcolsep}{4pt}\renewcommand{\arraystretch}{1.0}}

\title{JUICE: Unified Multimodal Modeling\\in Pretrained Representation Spaces}
\newcommand{\vclogo}[2]{\raisebox{-0.5\height}{\includegraphics[#1]{#2}}}
\titlelogos{%
\vclogo{height=18pt}{Figures/logos/nyu}\hfill
\vclogo{height=21pt,trim=7.5bp 7bp 7bp 7bp,clip}{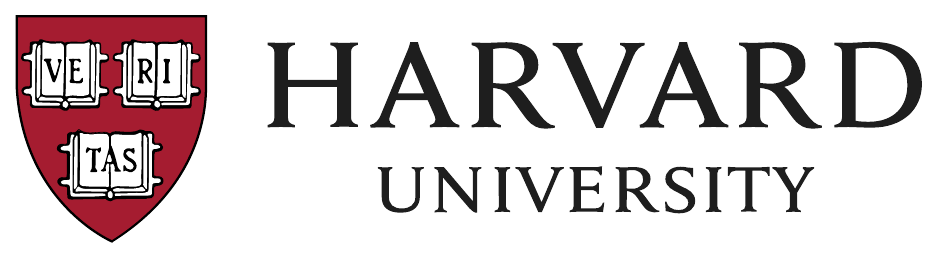}\hfill
\vclogo{height=19pt}{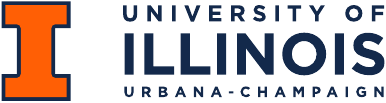}\hfill
\vclogo{height=20pt}{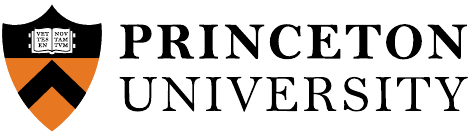}\hfill
\mbox{\vclogo{height=20pt}{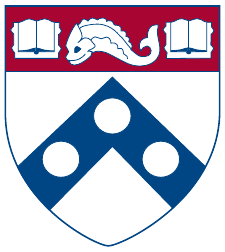}\hspace{3pt}\vclogo{height=20pt}{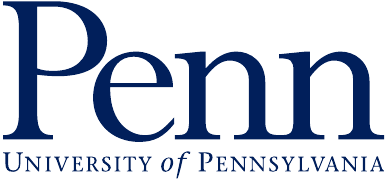}}}
\newcommand{\linkicon}[2]{\raisebox{-0.22\height}{\includegraphics[height=#1]{#2}}}
\author{%
Jiahan Li\textsuperscript{1,*}, Huangyuan Su\textsuperscript{2}, Shengbang Tong\textsuperscript{1}, Shusheng Yang\textsuperscript{1}, Ruihan Guo\textsuperscript{3}\\
Suozhi Huang\textsuperscript{4}, Hexu Zhao\textsuperscript{1}, Pratyusha Sharma\textsuperscript{1}, Sherry Yang\textsuperscript{1}, Jiatao Gu\textsuperscript{5}\\
Rob Fergus\textsuperscript{1,\textdagger}, Kyunghyun Cho\textsuperscript{1,\textdagger}\\[6pt]
{\normalfont\rmfamily\small
\textsuperscript{1}New York University\quad
\textsuperscript{2}Harvard University\quad
\textsuperscript{3}University of Illinois Urbana-Champaign\\
\textsuperscript{4}Princeton University\quad
\textsuperscript{5}University of Pennsylvania\\[4pt]
\textsuperscript{*}Project lead: \href{mailto:jl17319@nyu.edu}{\texttt{jl17319@nyu.edu}}\qquad\textsuperscript{\textdagger}Equal advising\\[8pt]
\begin{tabular}{@{}c@{\hspace{5pt}}l@{\hspace{8pt}}l@{}}
\linkicon{10pt}{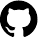} & \textsf{\textbf{Code}} & \codeurl\\
\linkicon{10.5pt}{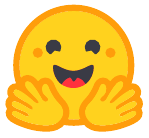} & \textsf{\textbf{Weights}} & \weightsurl\\
\raisebox{-0.22\height}{\includegraphics[height=10pt,trim=0 0 216.5bp 0,clip]{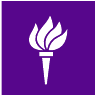}} & \textsf{\textbf{Blog}} & \blogurl
\end{tabular}}}

\begin{document}
\raggedbottom
\maketitle
\begin{abstract}
Most unified multimodal models treat images and text asymmetrically: a pretrained language model generates text, an attached diffusion head generates images, and outputs that contain both are produced in a fixed order, so the first output cannot adapt to the second. We ask whether both modalities can instead be generated by one process that is trained on them jointly from scratch. Pretrained encoder spaces hold both modalities in one continuous form and carry knowledge learned at scale, and recent work finds them effective as generative targets for vision and language. We therefore model the joint distribution directly in these spaces. We introduce JUICE (Joint Unified multimodal modeling In Continuous Encoder spaces), a single flow matching Transformer over frozen image and text representations, in which each modality has its own noise level. Which modalities are clean and which are generated then decides the task: image generation, captioning, visual question answering, or joint generation. Controlled pretraining runs test each design choice. At equal guidance, generating text as continuous states leaves image generation intact (48.41 GenEval, against 47.86 for a model trained only for text-to-image generation), while captions lose some exact wording. Training on joint and conditional denoising together beats either alone, and semantic encoder spaces beat VAE latents and raw pixels as targets. At 2.1B parameters, with no language model initialization and at $224{\times}224$, JUICE reaches up to 0.88 on GenEval and 73.1 on VQAv2. By sampling both modalities from the joint distribution (coupled sampling), it can also draw a route on a map while writing out the matching moves. On mazes with several valid solutions, joint success rises from 54.6\% with independent sampling to 93.7\%, and no modality has to go first. On visual spatial planning, JUICE solves 99.8\% of maps at its training sizes.
\end{abstract}

\begin{tcolorbox}[colback=papergray,colframe=paperblue!75!black,boxrule=0.6pt,arc=3pt,
  left=9pt,right=9pt,top=4pt,bottom=4pt,before skip=2pt,after skip=0pt]
\setlength{\parskip}{0pt}%
{\sffamily\bfseries Key Takeaways}\par\vspace{1pt}
{\color{paperblue!45}\rule{\linewidth}{0.4pt}}\par\vspace{2pt}
\fontsize{9.6}{11.5}\selectfont
\setlength{\tabcolsep}{0pt}\renewcommand{\arraystretch}{1}%
\begin{tabularx}{\linewidth}{@{}p{2.3em}X@{\hspace{8pt}}r@{}}
\keytakeaway{1}{Unified model}{one flow matching model generates both images and text; with its own noise level for each modality, the same weights handle every task.}{sec:method}
\keytakeaway{2}{Pretrained encoder spaces}{both modalities are generated as continuous states of frozen encoders; at equal guidance, image quality holds and captions lose only exact wording.}{sec:paradigm}
\keytakeaway{3}{Joint training}{adding joint denoising, with both modalities noisy, to the conditional tasks improves image generation and captioning alike.}{sec:mixture}
\keytakeaway{4}{Encoder choice}{semantic encoders beat VAE latents and pixels; SigLIP2, aligned with language, helps captioning most, and Flan-T5 gives the best text space.}{sec:representation}
\keytakeaway{5}{Scaling behavior}{every metric improves with model size, but rankings can flip: DINO leads image generation at 179M and SigLIP2 at 2.1B.}{sec:scaling}
\keytakeaway{6}{Joint reasoning}{sampled together, image and text adjust to each other, so a drawn route and its written moves agree without a fixed order.}{sec:joint-reasoning}
\end{tabularx}
\end{tcolorbox}

\clearpage
\begingroup
\setlength{\parskip}{0pt}
\makeatletter\patchcmd{\l@section}{1.0em}{0.45em}{}{}\makeatother
\small
\tableofcontents
\endgroup
\clearpage
\suppressfloats[t]
\section{Introduction}\label{sec:introduction}

Unified multimodal models aim to understand and generate both images and text within a single model. Most current systems, however, treat the two modalities asymmetrically. In the dominant design, a pretrained autoregressive language model generates text and drives a diffusion or flow matching head that generates images~\citep{showo2,tuna2}; other models quantize images and generate both modalities as discrete tokens~\citep{mmada,muddit}. Either way, the two modalities are generated by different processes or through a quantizer, and the joint distribution is factorized in a fixed order: an output that contains both an image and text is produced one modality after the other, and the first cannot be revised in light of the second. Vision is also usually added to a model that was first pretrained on language, although controlled studies of multimodal pretraining find that late integration makes models lean on language priors and that training the modalities jointly from the start works better~\citep{physicsmm}. This suggests a different design: a model that treats both modalities alike, generates them with one process, and learns them together from the start.

Such a design needs two things: a space in which one process can generate both modalities, and a source of knowledge for a backbone trained without a language model. Frozen pretrained encoders provide both. Their spaces are continuous for images and text alike and carry visual and linguistic knowledge learned at scale, and recent studies show that they are good generative targets for vision~\citep{zheng2026diffusion,scalingrae,blm} and for language~\citep{elf}, although they were never built for generation. In these spaces, one flow matching process generates both modalities with the same objective, the same features serve as targets and as context, and a model of the joint distribution also contains its conditionals. Give each modality its own noise level and the two levels span a square (Figure~\ref{fig:time-domain}). Text-to-image generation, captioning, and question answering hold one modality clean and move along an edge. Joint generation cuts through the middle, where image and text take shape together. To build a working model on this idea, we had to settle four questions that existing unified models leave open: whether text should be generated as continuous states or kept as tokens, which denoising tasks to train on, which encoder spaces to read and write, and whether these answers still hold as the model grows.

\begin{figure}[t]
\centering
\setbox0=\hbox{\includegraphics[width=.94\linewidth]{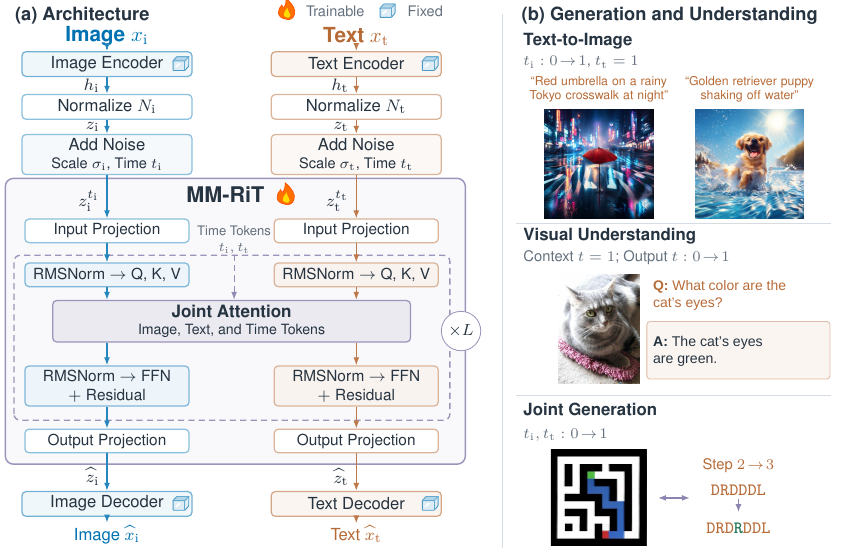}}
\setbox2=\hbox{\includegraphics[width=.97\linewidth]{Figures/juice_teaser.pdf}}
\setlength{\abovecaptionskip}{\dimexpr\abovecaptionskip+\ht0-\ht2\relax}
\leavevmode\box2
\caption{\textbf{\method{} overview.} (a)~Continuous image and text representations from frozen encoders share one MM-RiT backbone via bidirectional joint attention. Appendix~\ref{app:mmrit-block-details} explains the details of each block. (b)~Conditional tasks (text-to-image, visual question answering) and joint reasoning (maze solving) use the same interface by assigning clean context ($t{=}1$) and generation targets ($t{<}1$).}
\label{fig:teaser}
\end{figure}

We introduce \method{} (\textbf{J}oint \textbf{U}nified multimodal modeling \textbf{I}n \textbf{C}ontinuous \textbf{E}ncoder spaces; Figure~\ref{fig:teaser}), a unified model that works directly in the continuous spaces of frozen pretrained image and text encoders. A single flow matching backbone, MM-RiT, trained from scratch, connects the image and text streams through bidirectional joint attention~\citep{sd3}, and each stream has its own noise scale and schedule. We train MM-RiT on both the edges and the interior of the square, with conditional and joint denoising. At inference, a task is a path through the square: keeping some streams clean ($t=1$) and generating the others gives text-to-image generation, captioning, visual question answering, or coupled joint generation (Section~\ref{sec:method}).

Our pretraining studies (Section~\ref{sec:pretraining-study}) change one design choice at a time and keep the rest of the recipe fixed.
\emph{(i)~Generative frameworks.} At a shared guidance scale, \method{} scores 48.41 GenEval and a model trained only for text-to-image generation scores 47.86, while all three frameworks that write text as tokens fall below that model. Captions generated as continuous states lose some exact wording but match the images as well as autoregressive captions do.
\emph{(ii)~Training task mixtures.} Training should cover the regions that sampling visits: joint and conditional denoising together outperform either one alone, beyond the effect of supervising more targets.
\emph{(iii)~Representation spaces.} The VAE reconstructs better than the semantic encoders, yet generation and captioning both work better in the semantic spaces, and Flan-T5 states beat larger text embedding models.
\emph{(iv)~Scaling.} From 179M to 2.1B backbone parameters, SigLIP2~\citep{siglip2}, which is aligned with language, writes better captions at every size, while the early lead of WebSSL-DINO~\citep{webssl} (DINO) in image generation disappears.

With these answers in hand, we train two 2.1B models that pair Flan-T5 with SigLIP2 or with DINO (Section~\ref{sec:unified}). Working at $224{\times}224$ and starting from no language model, each uses one set of weights for compositional image generation, generation that needs world knowledge, and visual question answering. The two reach 0.87 and 0.88 GenEval, and the SigLIP2 model reaches 85.22 on DPG-Bench, 0.45 on WISE~\citep{wise}, and 73.1 on VQAv2.

Because \method{} models the joint distribution, it can also reason jointly across modalities: with coupled sampling, the image and its text are generated together, each conditioning on the other's evolving state. We adapt \method{} to two planning tasks, mazes with multiple solutions~\citep{ivanitskiy2023configurable} and visual spatial planning~\citep{vsp}, in which it draws a route on the map and writes the matching moves. Sampled independently, the two outputs often disagree on mazes: the model draws one valid route and writes out a different one, and joint success is only 54.6\%. Generating one modality first repairs this, but the result then depends on the order (95.2\% with the image first, 85.3\% with the text first), and nothing says in advance which order to use. Coupled sampling reaches 93.7\% without making that choice. In the two cases we trace, whichever output settles later rewrites itself to match the other. On visual spatial planning, \method{} solves 99.8\% of maps at the training sizes, and 94\% and 69\% at two larger sizes it never saw in training.

Our main contributions are:
\begin{enumerate}[leftmargin=*,itemsep=2pt,topsep=0pt,parsep=0pt]
\item \textbf{Joint modeling in pretrained representation spaces.} We propose \method{}, which models the joint distribution of images and text in the continuous spaces of frozen pretrained encoders with one flow matching Transformer trained from scratch. With a noise level for each modality, every task is a path through one square, with no fixed generation order.
\item \textbf{Systematic design studies.} Section~\ref{sec:pretraining-study} reports controlled pretraining runs on how text is generated, which tasks are trained, and which encoder spaces are used, plus a scaling study from 179M to 2.1B parameters. We summarize what they show as four takeaways for continuous joint modeling.
\item \textbf{Scaling and joint reasoning.} A single set of 2.1B weights handles compositional image generation, generation that needs world knowledge, and visual question answering (Section~\ref{sec:experiments}). On two planning tasks, coupled sampling gets the drawn route and the written moves to agree without a fixed generation order (Section~\ref{sec:joint-reasoning}).
\end{enumerate}

\section{Related Work}
\label{sec:related-work}\label{app:related-work}

\begin{table}[!t]
\caption{\textbf{Generation spaces and processes in multimodal models.} AR: autoregressive; FM: flow matching; repr.: representations.}
\label{tab:related}
\centering\small\setlength{\tabcolsep}{4pt}
\begin{tabular*}{\linewidth}{@{\extracolsep{\fill}}lllll@{}}
\toprule
Model & Text representation & Text process & Image representation & Image process \\
\midrule
Chameleon~\citep{chameleon} & Discrete tokens & AR & Discrete codes & AR \\
Transfusion~\citep{transfusion} & Discrete tokens & AR & VAE latents & Diffusion \\
Show-o2~\citep{showo2} & Discrete tokens & AR & VAE latents & FM \\
Tuna-2~\citep{tuna2} & Discrete tokens & AR & Pixels & FM \\
MMaDA~\citep{mmada} & Discrete tokens & Masked diffusion & Discrete codes & Masked diffusion \\
Muddit~\citep{muddit} & Discrete tokens & Masked diffusion & Discrete codes & Masked diffusion \\
D-DiT~\citep{dualdiffusion} & Discrete tokens & Masked diffusion & VAE latents & FM \\
UniDiffuser~\citep{unidiffuser} & CLIP text features & Diffusion & VAE + CLIP features & Diffusion \\
BLM~\citep{blm} & Discrete tokens & AR & Pretrained repr. & FM \\
Scale-RAE~\citep{scalingrae} & Discrete tokens & AR & Pretrained repr. & FM \\
\midrule
\textbf{\method{}} & Pretrained repr. & FM & Pretrained repr. & FM \\
\bottomrule
\end{tabular*}
\end{table}

\paragraph{Unified multimodal models.}
Most unified models are hybrids that combine different representation spaces and generative processes (Table~\ref{tab:related}).
Chameleon~\citep{chameleon} models discrete text tokens and quantized image codes autoregressively.
Transfusion~\citep{transfusion} pairs autoregressive text with diffusion over continuous image latents, and Show-o~\citep{showo} pairs autoregressive text with diffusion over discrete image tokens.
MMaDA~\citep{mmada} and Muddit~\citep{muddit} use discrete masked diffusion for both modalities, while D-DiT~\citep{dualdiffusion} combines discrete masked diffusion for text with continuous flow for images.
D-DiT and Muddit start from pretrained image generators, Show-o and MMaDA from pretrained language models, and Chameleon and Transfusion are trained from scratch.
In most of these models, the two modalities are generated by different processes, and outputs that contain both are produced in a fixed order. \method{} instead generates both with one flow matching process in frozen representation spaces and is trained from scratch; Section~\ref{sec:paradigm} compares it with representative formulations under the same recipe.

\paragraph{Generation in pretrained representation spaces.}
Representation autoencoders (RAEs)~\citep{zheng2026diffusion,scalingrae} show that diffusion works well on frozen pretrained visual features, whose semantic structure makes them good targets.
Beyond Language Modeling (BLM)~\citep{blm} takes up this idea and trains a unified backbone from scratch with flow matching over frozen visual features, while language is still predicted autoregressively as discrete tokens.
Other unified models pick different visual targets. Show-o2~\citep{showo2} still generates VAE latents, which it processes semantically. TUNA~\citep{tuna} aligns unified features built from VAE latents, and Tuna-2~\citep{tuna2} removes the visual encoders altogether and generates pixels.
For text, ELF~\citep{elf} applies continuous flow matching directly in pretrained text representation spaces, though only to generate text on its own.
Each of these works generates only one modality in a continuous representation space: BLM still predicts text as tokens, and ELF generates no images. \method{} generates both modalities in pretrained spaces with one process, so the same representations serve as generation targets and as cross-modal context, and Sections~\ref{sec:representation} and~\ref{sec:scaling} study which spaces suit this joint use and how the answer changes with scale.

\paragraph{Joint and conditional multimodal modeling.}
Separate noise schedules for each modality~\citep{unidiffuser,rojas2025diffuse} provide a mathematical framework for a single model to capture the marginal, conditional, and joint distributions of images and text.
Existing implementations, however, need specialized components. UniDiffuser~\citep{unidiffuser} diffuses CLIP embeddings and needs a separate autoregressive GPT-2 to decode text; OneFlow~\citep{oneflow} models text with discrete edit flows.
OmniFlow~\citep{omniflow} uses rectified flow for generation from any modality to any other. Its MMDiT backbone starts from Stable Diffusion~3, and its text latent comes from a VAE built on Flan-T5 and TinyLlama.
\method{} keeps both modalities as continuous encoder states with one state per token, decodes text with a lightweight parallel decoder, and trains one backbone on them from scratch. In this setting, joint denoising improves conditional generation by more than its extra targets explain (Section~\ref{sec:mixture}), and coupled sampling lets the two streams settle on the same one of several valid solutions (Section~\ref{sec:joint-reasoning}).

\paragraph{Continuous diffusion for language.}
Early work diffused learned word embeddings and rounded them back to tokens (Diffusion-LM~\citep{li2022diffusion}, DiffuSeq~\citep{gong2023diffuseq}); later work improved the likelihood objectives (CDCD~\citep{dieleman2022cdcd}, Plaid~\citep{gulrajani2023plaid}). Flow matching over categorical distributions, which places each token on the probability simplex, offers another continuous route to discrete data such as text~\citep{cheng2024sfm,cheng2025alphaflow}.
A second line of work diffuses directly in the representation spaces of pretrained language models, either after compressing sequences into fixed latents (LD4LG~\citep{lovelace2023ld4lg}) or over contextual encoder states (TEncDM~\citep{shabalin2025tencdm}, ELF~\citep{elf}).
\method{} follows the second line for its text stream and diffuses contextual encoder states. Unlike these text only models, it generates the states jointly with images, and Section~\ref{sec:paradigm} compares this choice with discrete text under one recipe.

\section{Method}
\label{sec:method}
\suppressfloats[t]

As shown in Figure~\ref{fig:teaser}, \method{} operates directly in the continuous representation spaces of frozen pretrained encoders, using a single flow matching Transformer (MM-RiT) trained from scratch.
The same model supports generation, understanding, and joint tasks over images and text simply by varying which representations are provided as clean inputs and which are generated from noise.
Section~\ref{sec:preliminaries} reviews flow matching, Section~\ref{sec:formulation} presents the formulation, and Sections~\ref{sec:architecture} to~\ref{sec:inference} describe the architecture, training procedure, and inference; Appendix~\ref{app:setup} provides implementation details.

\subsection{Preliminaries}
\label{sec:method-preliminaries}
\label{sec:preliminaries}
 
Flow matching~\citep{flowmatching} trains a network to predict a velocity field that, over time, transports a noise distribution to the data distribution.
Given a clean representation $z$ and Gaussian noise $\epsilon\sim\mathcal N(0,\sigma^2\mathbf I)$, we use the linear interpolation
\begin{equation}
  z^t = (1-t)\,\epsilon + t\,z, \qquad t\in[0,1],
  \label{eq:interpolant}
\end{equation}
where $t{=}0$ is pure noise, $t{=}1$ is clean data, and the target velocity is $v^*=z-\epsilon$.
Following JiT~\citep{jit}, MM-RiT predicts the clean state~$\widehat z_\theta$ rather than velocity.
This places the network output directly in the representation space used by the decoders. At any noisy state $z^t$, the predicted clean data $\widehat z_\theta$ defines the velocity field via
\[
  \widehat v_\theta(z^t;\, t) = \frac{\widehat z_\theta - z^t}{d(t)},
\]
where $d(t) = \max(1-t, 0.05)$ floors the denominator near $t{=}1$ for numerical stability. For the linear path, velocity matching is therefore a denoising loss on $\widehat z_\theta$ weighted by $1/d(t)^2$.

\subsection{Formulation}
\label{sec:formulation}

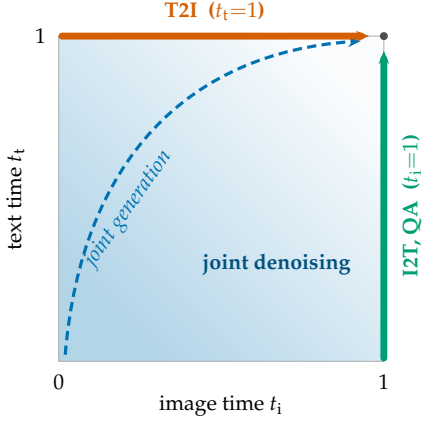
\begin{figure}[t]
\centering
\definecolor{tdT2I}{HTML}{D55E00}\definecolor{tdI2T}{HTML}{009E73}
\sbox0{\begin{tikzpicture}[x=4.3cm,y=4.3cm,font=\scriptsize,>={Stealth[length=5pt,width=4pt]}]
  \shade[shading=axis,bottom color=paperblue!30,top color=paperblue!2,shading angle=-45] (0,0) rectangle (1,1);
  \draw[black!30,line width=0.4pt] (0,0) rectangle (1,1);
  \node[text=paperblue!70!black,align=center,font=\scriptsize\bfseries] at (0.67,0.30) {joint denoising};
  \draw[->,line width=1.2pt,dash pattern=on 3.5pt off 2pt,paperblue] (0.02,0.02) .. controls (0.07,0.62) and (0.40,0.95) .. (0.93,0.985);
  \node[text=paperblue,rotate=55,font=\scriptsize\itshape] at (0.215,0.47) {joint generation};
  \draw[->,line width=2.6pt,tdT2I,line cap=round] (0.01,1) -- (0.955,1);
  \node[above=2pt,text=tdT2I,font=\scriptsize\bfseries] at (0.48,1) {T2I \ ($t_{\mathrm t}{=}1$)};
  \draw[->,line width=2.6pt,tdI2T,line cap=round] (1,0.01) -- (1,0.955);
  \node[rotate=90,anchor=north,yshift=-3pt,text=tdI2T,font=\scriptsize\bfseries] at (1,0.48) {I2T, QA \ ($t_{\mathrm i}{=}1$)};
  \fill[black!70] (1,1) circle (1.5pt);
  \node[below=1pt] at (0,0) {0};
  \node[below=1pt] at (1,0) {1};
  \node[left=1pt] at (0,1) {1};
  \node[below=9pt] at (0.5,0) {image time $t_{\mathrm i}$};
  \node[rotate=90,above=9pt] at (0,0.5) {text time $t_{\mathrm t}$};
\end{tikzpicture}}%
\begin{minipage}[c]{\wd0}\usebox0\end{minipage}%
\hspace{1.2em}%
\begin{minipage}[c]{\dimexpr\linewidth-\wd0-1.2em\relax}
\caption{\textbf{Tasks are paths in the 2D time domain.} Each modality has its own time, from noise ($t{=}0$) to clean ($t{=}1$). Conditional tasks run along the edges, where the condition is clean, and joint generation crosses the interior; its path bends because the image schedule keeps the image noisier for longer. Training covers both regions: conditional denoising on the edges and joint denoising in the interior.}
\label{fig:time-domain}
\end{minipage}
\end{figure}

\method{} models the joint distribution of an image representation $z_{\mathrm i}$ and a text representation $z_{\mathrm t}$, both extracted from frozen pretrained encoders (Section~\ref{sec:architecture}). Each modality has its own time $t_m\in[0,1]$, and a single network predicts both clean states from the pair of noisy states:
\begin{equation}
  (\widehat z_{\mathrm i},\,\widehat z_{\mathrm t})
  = f_\theta\!\bigl(z_{\mathrm i}^{t_{\mathrm i}},\,
                     z_{\mathrm t}^{t_{\mathrm t}};\;
                     t_{\mathrm i},t_{\mathrm t}\bigr).
  \label{eq:joint-prediction}
\end{equation}
The pair of times lives in a unit square, the 2D time domain of Figure~\ref{fig:time-domain}. Setting $t_m{=}1$ makes modality~$m$ a clean condition, and any $t_m{<}1$ means it is still being generated. A task is thus just a path through the square. Text-to-image generation follows the edge $t_{\mathrm t}{=}1$ and captioning the edge $t_{\mathrm i}{=}1$, while joint generation cuts through the interior. Question answering sets times per token, so the image and the question stay at $t{=}1$ while the answer is generated (Appendix~\ref{app:conditioning}). We therefore train on the edges with conditional denoising and on the interior with joint denoising, so that every region sampling passes through is covered.

The formulation introduces three requirements, each addressed by one design component. First, because the two representation spaces differ in dimensionality and feature statistics, each modality has its own projections, noise scale, and noise schedule (Sections~\ref{sec:architecture} and~\ref{sec:training}). Second, because any token can be either a condition or a generation target, time tokens in the sequence and a time embedding for each token tell the network the role of each position (Section~\ref{sec:architecture}). Third, because one set of weights must serve both conditional directions and joint generation, MM-RiT is trained on a mixture of conditional and joint denoising (Section~\ref{sec:training}).

\subsection{Architecture}
\label{sec:method-representations}
\label{sec:architecture}
 
\textbf{Representation Encoders.}
Given a paired image~$x_{\mathrm i}$ and text~$x_{\mathrm t}$, frozen pretrained encoders extract features that are normalized to align their scales with the diffusion noise~\citep{rit2026}:
\begin{equation}
  h_m = E_m(x_m), \qquad z_m = N_m(h_m), \qquad m\in\{\mathrm i,\mathrm t\}.
  \label{eq:encoding}
\end{equation}
SigLIP2~\citep{siglip2} and Flan-T5-Small~\citep{flant5} are our default image and text encoders, and Section~\ref{sec:representation} looks at other choices. We do not finetune the encoders. Their spaces already encode visual and linguistic knowledge learned at scale, and if they stay fixed, the same features can be used both as targets to generate and as context to condition on.
 
\textbf{Multimodal Representation Diffusion Transformer (MM-RiT).}
MM-RiT implements $f_\theta$ in Equation~\ref{eq:joint-prediction}: it receives both streams at independently specified noise levels $(t_{\mathrm i},t_{\mathrm t})$~\citep{unidiffuser} and predicts both denoised clean states, which we convert to velocities for training (velocity matching loss) and inference (ODE solver).
To accommodate the distinct image and text representation spaces, each MM-RiT block uses separate Q/K/V projections for each modality, concatenates the resulting sequences for one bidirectional joint attention operation, and then applies separate output projections and feedforward networks~\citep{scalingrae,blm,minit2i2026}. Time tokens in the sequence and additive time embeddings for each token expose the noise level of each modality and each token, so the network can distinguish clean conditioning positions ($t{=}1$) from generation targets ($t{<}1$).

\textbf{Representation Decoders.}
Fixed decoders map the generated representations back to pixels and tokens at inference:
\begin{equation}
  \widehat x_{\mathrm i} = D_{\mathrm i}(z_{\mathrm i}), \qquad
  \widehat x_{\mathrm t} = D_{\mathrm t}(z_{\mathrm t}).
  \label{eq:decoding}
\end{equation}
For images we use the Scale-RAE decoder that belongs to the image encoder~\citep{scalingrae}. For text, a lightweight parallel decoder maps all generated states to tokens in one forward pass, which is much faster than autoregressive decoding; the pretraining study and the final models each use one decoder of this design. All decoders are trained separately and stay fixed during MM-RiT training. Appendix~\ref{app:decoders} describes them and compares parallel with autoregressive decoding.

\subsection{Training}
\label{sec:method-training}
\label{sec:training}
 
\textbf{Noise and Schedules for Each Modality.}
Image and text representations differ in dimensionality, scale, and geometry, and noise settings that work for one may not work for the other, so each modality gets its own.
First, the Gaussian noise scale is set separately ($\sigma_{\mathrm i}{=}1$, $\sigma_{\mathrm t}{=}2$ by default) and matched to the statistics of each space after normalization.
Second, the noise levels follow a separate distribution for each modality: for modality~$m$, we draw a base time~$u_m$ from a logit normal distribution and transform it by
\begin{equation}
  t_m = S_{\alpha_m}(u_m), \qquad
  S_\alpha(u) = \frac{u}{\alpha + (1-\alpha)\,u}.
  \label{eq:time-shift}
\end{equation}
For images, we set the shift to $\alpha_{\mathrm i}=\sqrt{n_{\mathrm i}\,d_{\mathrm i}/4096}$, where $n_{\mathrm i}$ is the number of image tokens and $d_{\mathrm i}$ their dimension. This moves probability mass toward noisier states. Text is not shifted ($\alpha_{\mathrm t}{=}1$). Features of high dimension can keep much of their information at moderate noise levels, so the shift makes the model see states dominated by noise more often~\citep{sd3,zheng2026diffusion}.
 
\textbf{Joint and Conditional Training.}
With these schedules, MM-RiT is trained on a mixture of three denoising tasks (Algorithm~\ref{alg:training}).
For each pair of image and text from a training set~$\mathcal D$, we draw a task with the mixture weights $(\pi_{\mathrm J},\pi_{\mathrm{T2I}},\pi_{\mathrm{I2T}})$:
\emph{joint} denoising draws the noise levels of both modalities independently, $t_m\sim p_m(t)$ with $p_m$ the distribution of Equation~\ref{eq:time-shift}, and corrupts both streams;
conditional \emph{T2I} keeps the text clean at $t_{\mathrm t}{=}1$ and corrupts only the image;
and conditional \emph{I2T} does the opposite, keeping the image clean at $t_{\mathrm i}{=}1$.
In the time domain of Section~\ref{sec:formulation}, joint denoising samples the interior ($t_{\mathrm i}{<}1,\,t_{\mathrm t}{<}1$) and the conditional tasks lie on its edges~\citep{unidiffuser}.
The conditional tasks match the clean conditions used at inference. Joint denoising is what teaches the model the joint distribution, which coupled sampling and the guidance reference for image conditions need (Section~\ref{sec:inference}). In pretraining, the velocity matching loss covers only corrupted target positions, so clean conditioning positions and text padding are left out; $\lambda_{\mathrm i}$ and $\lambda_{\mathrm t}$ balance the two modalities.
 
\begin{algorithm}[t]
\caption{\method{} training (one gradient step)}
\label{alg:training}
\begin{algorithmic}[1]
\REQUIRE Training set $\mathcal D$ of paired images and texts;\;
         task mixture weights $(\pi_{\mathrm J},\pi_{\mathrm{T2I}},\pi_{\mathrm{I2T}})$;\;
         MM-RiT $f_\theta$;\;
         modality loss coefficients $\lambda_{\mathrm i},\lambda_{\mathrm t}$
\STATE \textbf{Sample data:}\;
       $(x_{\mathrm i},x_{\mathrm t})\sim\mathcal D$
\STATE \textbf{Encode representations:}\;
       $z_m \leftarrow N_m\!\bigl(E_m(x_m)\bigr)$
       for each modality $m\in\{\mathrm i,\mathrm t\}$
\STATE \textbf{Sample denoising task:}\;
       $r\sim\mathrm{Categorical}(\pi_{\mathrm J},\pi_{\mathrm{T2I}},\pi_{\mathrm{I2T}})$
\STATE \quad\textbf{Joint denoising:}\;\;
       $t_{\mathrm i}\!\sim\!p_{\mathrm i}(t)$,\;
       $t_{\mathrm t}\!\sim\!p_{\mathrm t}(t)$;\;
       targets $\mathcal T\!=\!\{\mathrm i,\mathrm t\}$
\STATE \quad\textbf{Conditional T2I:}\;
       $t_{\mathrm i}\!\sim\!p_{\mathrm i}(t)$,\;
       $t_{\mathrm t}\!=\!1$;\;
       targets $\mathcal T\!=\!\{\mathrm i\}$
\STATE \quad\textbf{Conditional I2T:}\;\,
       $t_{\mathrm i}\!=\!1$,\;
       $t_{\mathrm t}\!\sim\!p_{\mathrm t}(t)$;\;
       targets $\mathcal T\!=\!\{\mathrm t\}$
\STATE \textbf{Corrupt target representations:}\;
       for each $m\!\in\!\mathcal T$,\;
       sample $\epsilon_m\!\sim\!\mathcal N(0,\sigma_m^2\mathbf I)$;\;
       $z_m^{t_m}\!\leftarrow\!(1\!-\!t_m)\,\epsilon_m+t_m\,z_m$
\STATE \textbf{Predict clean states:}\;
       $(\widehat z_{\mathrm i},\widehat z_{\mathrm t})
       \leftarrow f_\theta(z_{\mathrm i}^{t_{\mathrm i}},
                           z_{\mathrm t}^{t_{\mathrm t}};\,
                           t_{\mathrm i},t_{\mathrm t})$
\STATE \textbf{Velocity matching loss:}\;
       $\displaystyle
       \mathcal L \leftarrow \sum_{m\in\mathcal T}\lambda_m
       \operatorname*{mean}_{p,\,c}
       \frac{(\widehat z_{m,pc}-z_{m,pc})^2}
            {d(t_m)^2}$,\;
       over target positions $p$ and channels $c$
\STATE \textbf{Update MM-RiT:}\;
       $\theta \leftarrow \theta - \eta\,\nabla_\theta\mathcal L$
\end{algorithmic}
\end{algorithm}

\subsection{Inference}
\label{sec:method-inference}
\label{sec:inference}
 
\textbf{Conditional and Joint Generation.}
At inference, conditioning inputs stay at $t{=}1$, and generation targets are integrated from Gaussian noise up to $t{=}1$.
Text-to-image, image-to-text, and joint generation differ only in the path they take through the 2D time domain~\citep{unidiffuser} (Figure~\ref{fig:time-domain}):
\begin{equation}
  (t_{\mathrm i}(s),\,t_{\mathrm t}(s))=
  \begin{cases}
    (S_{\alpha_{\mathrm i}}(s),\;1), & \text{T2I},\\[2pt]
    (1,\;S_{\alpha_{\mathrm t}}(s)), & \text{I2T},\\[2pt]
    (S_{\alpha_{\mathrm i}}(s),\;S_{\alpha_{\mathrm t}}(s)),
      & \text{joint},
  \end{cases}
  \quad s\in[0,1].
  \label{eq:inference-paths}
\end{equation}

\textbf{Classifier-Free Guidance.}
For conditional tasks we apply classifier-free guidance~\citep{cfg} in velocity space: the prediction with the condition is pushed away from a reference prediction without it, which strengthens what the condition adds. No training stage drops the condition, so \method{} has no learned null token. The reference instead replaces the condition with an uninformative input that the model has learned to handle. For an image condition, the image is placed at its noise end, a state the joint task trains the model to see next to text; for a text condition, we use the clean states of the generic caption ``a photo.'', which says almost nothing about the image. We use $w{=}5$ for text-to-image generation, $w{=}3$ for captioning and question answering, and no guidance for joint generation and planning. Appendix~\ref{app:conditioning} gives the formulas, and Table~\ref{tab:guidance-grid} compares the text reference with placing the caption at its noise end.

\textbf{Coupled Sampling.}
In joint generation, both velocity fields are computed from the current joint state $(z_{\mathrm i}^{t_{\mathrm i}}, z_{\mathrm t}^{t_{\mathrm t}})$ before either modality is updated. We call this \emph{coupled sampling}. Each stream is thus conditioned, step by step, on the other stream as it evolves. \emph{Independent sampling} works differently. It integrates each stream separately and leaves the other one at its initial noise for the whole trajectory, which amounts to sampling each modality from its marginal. \emph{Sequential sampling} first generates one modality completely and then the other conditioned on it, with either the image or the text going first.

\begingroup
\section{Pretraining Study}
\label{sec:pretraining-study}
\label{sec:design-studies}

Putting both modalities in continuous encoder spaces under one denoiser raises the four questions of Section~\ref{sec:introduction}, and this section takes them in turn. Should text be generated as continuous states or as tokens (Section~\ref{sec:paradigm})? Which denoising tasks should training cover (Section~\ref{sec:mixture})? Which encoder spaces should the model read and write (Section~\ref{sec:representation})? And do the answers change with model size (Section~\ref{sec:scaling})? All runs share the setup below, and each study ends with a takeaway.

\FloatBarrier
\subsection{Setup}
\label{sec:setup}

All studies share one pretraining recipe and vary one design choice at a time. The reference model has an MM-RiT backbone with 668M trainable parameters over frozen SigLIP2-SO400M image features and Flan-T5-Small text states. It is trained for 50k updates at a global batch size of 4,096 with AdamW ($\beta_1{=}0.9$, $\beta_2{=}0.95$, weight decay 0.01) and a constant learning rate of $4{\times}10^{-4}$ after 5,000 warmup steps, on 38M pairs of images and captions~\citep{blip3o,scalingrae} and Cambrian-7M~\citep{cambrian}, using the default task ratios (55.6\% joint, 22.2\% T2I, 22.2\% I2T; Appendix~\ref{app:noise-loss}).
For image generation we report GenEval (GE)~\citep{geneval} and DPG-Bench (DPG)~\citep{ella,dpgbench}, with 25 Euler steps and guidance $w{=}5$ against the generic caption reference (a fully masked caption for the base D-DiT-style framework; Appendix~\ref{app:conditioning}). For captioning we report CIDEr~\citep{cider} and CLIPScore (CLIP, $\times100$)~\citep{clipscore} on 5,000 pairs held out from training, with 16 steps and $w{=}3$ (Section~\ref{sec:inference}).
Each configuration is trained once. The framework and representation comparisons change complete pipelines, since a representation comes with its own decoder, normalization, and noise settings; they therefore compare design choices as they would be deployed rather than isolated components. Appendices~\ref{app:details} and~\ref{app:design-studies} give the full configurations.

\FloatBarrier
\subsection{Generative Frameworks}
\label{sec:paradigm}

The least conventional choice in \method{} is that it generates text as continuous encoder states, not as tokens. We compare it with three frameworks that keep text discrete. All four use the same 668M trunk with two streams and the same data, task ratios, and schedule. The three baselines are Transfusion-style autoregressive (AR) text with flow matching (FM) for images~\citep{transfusion}, D-DiT-style masked diffusion (MD) for text with flow matching for images~\citep{dualdiffusion}, and MMaDA-style masked diffusion for both modalities over vector quantized (VQ) image codes~\citep{mmada} (Table~\ref{tab:paradigm}, Figure~\ref{fig:paradigm-scaling}; Appendix~\ref{app:architecture-implementations}).

\begin{table}[!htbp]
\caption{\textbf{Generative frameworks at 668M.} All configurations share the trunk with two streams, the data, and the schedule. FM: flow matching; AR: autoregressive; MD: masked diffusion. The middle block gives the discrete text frameworks the same frozen T5 condition as JUICE, and the last row is a model trained only for text-to-image generation.}
\label{tab:paradigm}
\label{tab:framework}
\centering\small
\setlength{\tabcolsep}{4pt}
\begin{tabularx}{\linewidth}{@{}lccl*{4}{R}@{}}
\toprule
& \multicolumn{2}{c}{Generation} & & \multicolumn{2}{c}{Image generation} & \multicolumn{2}{c}{Captioning}\\
\cmidrule(lr){2-3}\cmidrule(lr){5-6}\cmidrule(l){7-8}
Configuration & Text & Image & Text condition & GE (\%) & DPG (\%) & CIDEr & CLIP\\
\midrule
\rowcolor{paperblue!6}JUICE & FM & FM & frozen T5 states & \textbf{48.41} & \textbf{78.85} & 88.80 & 80.20\\
Transfusion-style~\citep{transfusion} & AR & FM & token embeddings & 44.23 & 76.84 & \textbf{160.52} & \textbf{80.37}\\
D-DiT-style~\citep{dualdiffusion} & MD & FM & token embeddings & 34.50 & 74.35 & 115.22 & 79.82\\
MMaDA-style~\citep{mmada} & MD & MD (VQ) & token embeddings & 41.61 & 66.50 & 89.53 & 71.61\\
\midrule
Transfusion-style & AR & FM & frozen T5 states & 46.07 & 78.24 & 158.63 & 80.36\\
D-DiT-style & MD & FM & frozen T5 states & 37.58 & 76.31 & 114.47 & 79.72\\
MMaDA-style & MD & MD (VQ) & frozen T5 states & 43.35 & 68.46 & 89.52 & 71.59\\
\midrule
T2I only reference & -- & FM & frozen T5 states & 47.86 & 77.50 & 2.83 & 25.15\\
\bottomrule
\end{tabularx}
\end{table}

\paragraph{Image generation.}
\method{} reaches 48.41 GenEval and 78.85 DPG, the highest of the four frameworks. Part of this lead comes from the condition encoder: the discrete text frameworks condition image generation on learned token embeddings, whereas \method{} conditions on contextual T5 states. With the same frozen T5 condition, all three improve (Transfusion-style goes from 44.23 to 46.07 GenEval), but none of them reaches \method{}. A model trained only for text-to-image generation gets 47.86 at the same guidance scale. Adding a continuous text stream thus leaves image generation intact at this scale (48.41), whereas every token text stream lowers GenEval, with or without the T5 condition (for MMaDA-style, the image process also changes). With the condition matched, what remains different is the text stream: \method{} trains both streams with the same flow matching objective, while the discrete text frameworks train them with different objectives and attention patterns. With the guidance scale tuned per model, the T2I only model peaks at $w{=}3$ (51.23) and leads \method{} by 2.8 points (Appendix~\ref{app:guidance-ablation}).

\begin{figure}[!htbp]
\centering
\includegraphics[width=\linewidth]{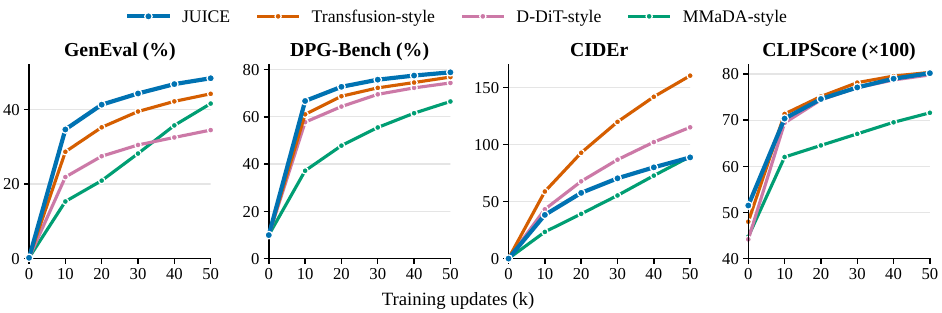}
\caption{\textbf{Continuous text generation gives the strongest image generation throughout training.} Generative framework comparison at the 668M scale (Table~\ref{tab:paradigm}).}
\label{fig:paradigm-scaling}
\end{figure}

\paragraph{Captioning.}
On the alignment between captions and images, \method{} matches autoregressive text (80.20 vs.\ 80.37 CLIPScore). On n-gram agreement with the reference captions, it is the weakest of the four: its CIDEr (88.80) is just over half that of autoregressive text (160.52) and below D-DiT-style masked text diffusion (115.22). The gap lies mainly in the generated states, not in the parallel caption decoder of this study, which recovers 99.2\% of tokens from clean T5 states (Table~\ref{tab:encoder-text-readout}; Appendix~\ref{app:text-decoder}). The captions thus describe the image content as well as autoregressive ones but word it differently from the references. This is the pattern expected when errors in a generated continuous state move it toward nearby states with a similar meaning: the wording changes more than the content.

\begin{takeawaybox}
\textbf{Takeaway: Continuous text spares images but costs wording.} Of the four frameworks, only the one with continuous text keeps up with a model trained only for text-to-image generation on GenEval at the shared guidance scale (48.41 vs.\ 47.86); the three token text frameworks all fall below it, even with the T5 condition. Continuous text pays in CIDEr instead (88.80 vs.\ 160.52 for autoregressive text), while its CLIPScore is on par.
\end{takeawaybox}

\FloatBarrier
\subsection{Training Task Mixtures}
\label{sec:mixture}

Because both modalities are continuous and each has its own noise level, \method{} can be trained not only on conditional tasks, where one modality is clean, but also on joint denoising, where both are noisy and each is predicted next to a noisy partner. In the 2D time domain $(t_{\mathrm i},t_{\mathrm t})$, the conditional T2I and I2T tasks lie on the edges and joint denoising covers the interior (Section~\ref{sec:training}; Figure~\ref{fig:time-domain}). We explore what joint denoising brings and which regions training should cover: Table~\ref{tab:mixture} compares five recipes, and Figure~\ref{fig:taskmix-training} follows two of them, Mixing and Multi-task, over training.

\begin{table}[!htbp]
\caption{\textbf{Training task mixtures at 668M.} Task ratios give the share of examples assigned to each denoising task. The controls hold the number of supervised targets or the cross-modal exchange fixed (Appendix~\ref{app:sim-supervision}): matched DC (dual conditional) replaces each joint example with two conditional ones, blocked joint removes cross-modal attention on joint examples, single target joint supervises only one stream of each joint example, and Multi-task trained for 77.8k updates matches the target count of Mixing.}
\label{tab:mixture}
\centering\small
\setlength{\tabcolsep}{4pt}
\begin{tabularx}{\linewidth}{@{}lRRRRRRR@{}}
\toprule
& \multicolumn{3}{c}{Task ratio (\%)} & \multicolumn{2}{c}{Image generation} & \multicolumn{2}{c}{Captioning}\\
\cmidrule(lr){2-4}\cmidrule(lr){5-6}\cmidrule(l){7-8}
Recipe & I2T & T2I & Joint & GE (\%) & DPG (\%) & CIDEr & CLIP\\
\midrule
T2I only & 0 & 100 & 0 & 47.86 & 77.50 & 2.83 & 25.15\\
I2T only & 100 & 0 & 0 & 0.21 & 9.95 & \textbf{107.22} & 77.81\\
Multi-task & 50 & 50 & 0 & 36.30 & 77.71 & 79.41 & 75.07\\
Joint only & 0 & 0 & 100 & 31.33 & 56.98 & 75.66 & 73.70\\
\rowcolor{paperblue!6}Mixing (default) & 22.2 & 22.2 & 55.6 & \textbf{48.41} & \textbf{78.85} & 88.80 & \textbf{80.20}\\
\midrule
\multicolumn{8}{@{}l}{\textit{Controls}}\\
Matched DC & \multicolumn{3}{c}{joint $\to$ 2 cond.} & 45.90 & 78.05 & 87.72 & 79.63\\
Blocked joint & \multicolumn{3}{c}{no cross attention} & 45.49 & 78.14 & 86.58 & 78.66\\
Single target joint & \multicolumn{3}{c}{1 target per joint} & 40.08 & 77.98 & 82.94 & 77.33\\
Multi-task, 77.8k updates & \multicolumn{3}{c}{matched targets} & 43.16 & 78.51 & 89.31 & 79.14\\
\bottomrule
\end{tabularx}
\end{table}

\begin{figure}[!htbp]
\centering
\includegraphics[width=\linewidth]{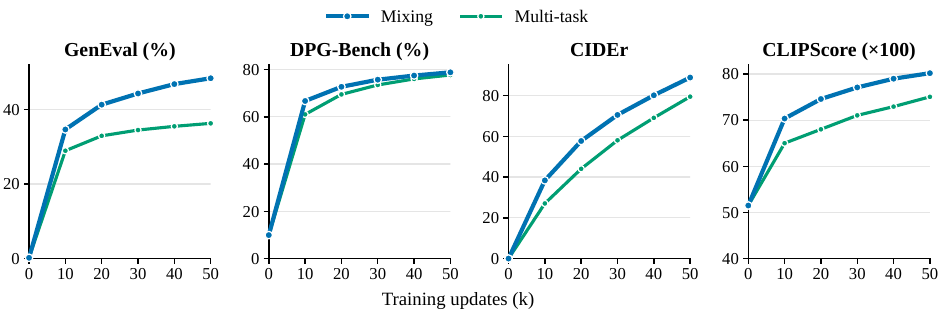}
\caption{\textbf{Joint plus conditional training leads conditional training alone throughout pretraining.} Mixing combines joint, T2I, and I2T denoising; Multi-task uses T2I and I2T only.}
\label{fig:taskmix-training}
\end{figure}

\paragraph{Edges and interior.}
Training in a single direction specializes: the T2I only and I2T only models do well in the direction they are trained for and fail in the other. Training only the edges (Multi-task) supports both directions, but training only the interior (Joint only) is worse on all four metrics. Joint only never sees the boundary. At conditional inference the condition is exactly clean, and training on the interior never produces that state. Mixing trains on both and is the best bidirectional recipe on every metric; it also stays ahead of Multi-task throughout training. The rule is simple: train where sampling goes. T2I and I2T sampling follow the edges, while joint sampling goes through the interior. At the shared guidance scale, Mixing is as good as the T2I only model at image generation, but it is still behind the I2T only model on CIDEr (88.80 vs.\ 107.22). The ordering of the bidirectional recipes also holds with the guidance scale tuned per recipe: with each recipe at its best scale in $w\in[1,5]$, Mixing (48.41) still leads Multi-task (44.29) and Joint only (31.33) on GenEval, while T2I only rises to 51.23 (Appendix~\ref{app:guidance-ablation}).

\paragraph{More targets or cross-modal exchange?}
Extra supervision explains only part of the gain. Mixing does supervise each modality more often than Multi-task (on 77.8\% vs.\ 50\% of examples), so we control for target count (lower block of Table~\ref{tab:mixture}; Appendix~\ref{app:sim-supervision}). One control replaces each joint example with two conditional ones at the same target count (matched dual conditional, DC), at 1.56 times the backbone work; another blocks cross-modal attention on joint examples. Both lower every metric relative to Mixing, by 2.51 and 2.92 GenEval points, respectively. In the other direction, a joint branch that keeps both streams noisy but supervises only one of them (single target joint) beats Multi-task on all four metrics with the same number of targets. Training Multi-task for longer, until it sees as many targets as Mixing, closes the CIDEr gap (89.31) but not the GenEval gap (43.16). The noisy partner and the exchange with it each contribute.

\begin{takeawaybox}
\textbf{Takeaway: Train where you sample.} Sampling uses both the edges of the square and its interior, and training has to cover both. Joint only never sees a clean condition and reaches just 31.33 GenEval at its best guidance scale; Multi-task never trains on the interior and reaches 44.29; Mixing gets 48.41. Matching the target count does not close the gap, so the joint examples give more than extra supervision.
\end{takeawaybox}

\FloatBarrier
\subsection{Representation Spaces}
\label{sec:representation}

\method{} reads and writes in the spaces of its frozen encoders, so the encoder determines both what the model sees and what it must generate. Encoders differ in what their spaces capture: some are trained to reconstruct their input, others to capture semantics, and some are aligned with the other modality. We explore these tradeoffs on both sides. With T5 fixed for text, we compare SigLIP2~\citep{siglip2} and WebSSL-DINO~\citep{webssl} (DINO) image features with FLUX VAE latents~\citep{flux} and raw pixels~\citep{jit,tuna2}. With SigLIP2 fixed for images, we compare T5 with Qwen3-Embedding (Qwen3-E)~\citep{qwen3embedding}, Qwen3-VL-Embedding (Qwen3-VL-E)~\citep{qwen3vlemb}, and SigLIP2 text features (Figure~\ref{fig:encoder}; full metrics in Table~\ref{tab:representation}).

\begin{figure}[!htbp]
\centering
\includegraphics[width=\linewidth]{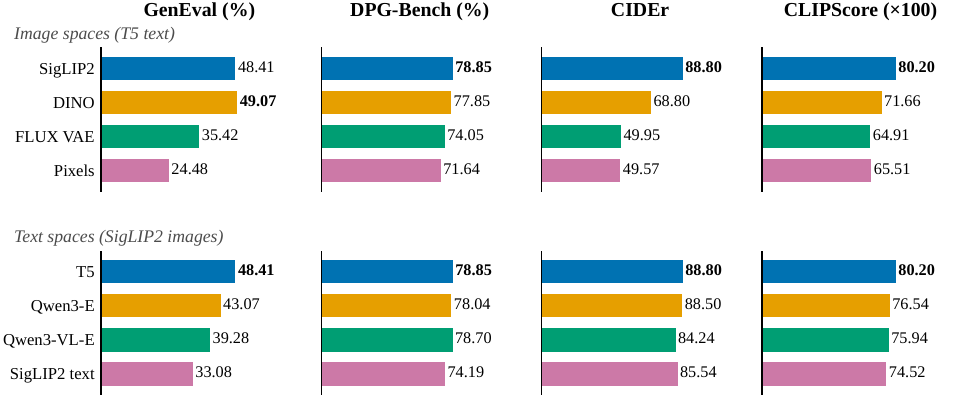}
\caption{\textbf{Semantic encoder spaces outperform VAE latents and pixels, and T5 leads among text spaces.} Top: image spaces with T5 text. Bottom: text spaces with SigLIP2 images.}
\label{fig:encoder}
\end{figure}

\paragraph{Image spaces.}
Semantic encoders outperform spaces built for reconstruction on all four metrics: SigLIP2 and DINO reach 48.41 and 49.07 GenEval, against 35.42 for the VAE and 24.48 for pixels. This ordering is broadly the reverse of reconstruction quality, as the FLUX VAE reconstructs best among the learned spaces (ImageNet rFID 0.288, against 0.435 for SigLIP2; Table~\ref{tab:encoder-image-codec}). The two semantic encoders are close on image generation, but captioning separates them: SigLIP2, which is aligned with language, leads DINO by 20.0 CIDEr and 8.5 CLIPScore. For a model that both generates and conditions on the same features, semantic structure thus matters more than reconstruction fidelity. The order of the two semantic encoders on image generation changes with model size (Section~\ref{sec:scaling}).

\paragraph{Text spaces.}
Flan-T5 states lead on all four metrics, ahead of larger embedding models and of SigLIP2 text features, which are aligned with the image space. Each text space is decoded with its own parallel decoder, which recovers 97.5\% to 99.2\% of tokens from clean states, and even a linear readout recovers 96.5\% to 98.5\% (Table~\ref{tab:encoder-text-readout}). All four spaces thus retain most of the token information. The decoder affects only the captioning metrics, and the GenEval differences, which do not involve it, indicate that the ordering reflects how well each space serves as a generation target and condition. Qwen3-E comes close to T5 on CIDEr (88.50 vs.\ 88.80) but trails it by 5.3 GenEval points, and the SigLIP2 text features give the lowest GenEval (33.08). Alignment with the image space therefore does not make a text space a better target or condition here.

\begin{takeawaybox}
\textbf{Takeaway: Semantics over reconstruction.} Good reconstruction does not make a good generation space. The FLUX VAE has the best rFID among the learned spaces (0.288 vs.\ 0.435 for SigLIP2), but GenEval in it drops to 35.42, from 48.41 with SigLIP2. SigLIP2 and DINO are close on generation, and SigLIP2's alignment with language shows up in captioning (20.0 CIDEr ahead). On the text side, Flan-T5 states beat larger embedding models and SigLIP2 text features.
\end{takeawaybox}

\FloatBarrier
\subsection{Scaling}
\label{sec:scaling}

Do these preferences survive as the model grows? Keeping the recipe above, we train MM-RiT backbones from 179M to 2.1B trainable parameters (d12 to d28; Appendix~\ref{app:total-parameters}) on both the SigLIP2 and the DINO image space (Figure~\ref{fig:scaling}). All four metrics improve with size in both spaces, but scale treats the two roles of the image space differently. For image generation the order flips. DINO is well ahead at 179M (32.90 vs.\ 18.99 GenEval), the gap shrinks as the backbone grows, and by 2.1B SigLIP2 is ahead (56.81 vs.\ 52.78). Captioning shows no such flip: SigLIP2 leads at every size and reaches 86.64 CLIPScore at 2.1B, against 72.32 for DINO. The two trends separate the roles of the image space: DINO's advantage as a generation target at small scale fades as capacity grows, whereas the benefit of SigLIP2's language alignment for producing text persists. After finetuning, the two 2.1B models of Section~\ref{sec:experiments} differ by less than 0.02 GenEval (DINO 0.88, SigLIP2 0.87), so neither image space keeps a clear GenEval advantage, while SigLIP2 leads on almost every understanding benchmark.

\begin{figure}[!htbp]
\centering
\includegraphics[width=\linewidth]{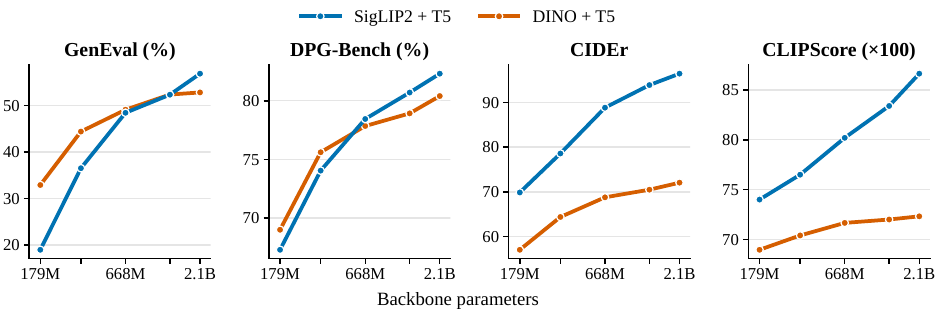}
\caption{\textbf{Scaling the backbone from 179M to 2.1B parameters.} DINO's early lead in image generation disappears with scale, while SigLIP2 leads captioning at every size. DPG-Bench in this figure uses the time swap text reference (Appendix~\ref{app:guidance-ablation}).}
\label{fig:scaling}
\end{figure}

\begin{takeawaybox}
\textbf{Takeaway: Scale reorders image spaces for generation, not captioning.} DINO's lead in image generation fades as the model grows, while SigLIP2's language alignment helps captioning at every size.
\end{takeawaybox}

\section{Experiments}
\label{sec:experiments}
\label{sec:scaling-unified}
\label{sec:unified}

A single model of the joint distribution of images and text should provide three abilities: generating images from text, answering questions about images, and producing image and text outputs that are individually correct and mutually consistent. We now evaluate \method{} on each of them at scale.

\subsection{Setup}
\label{sec:exp-setup}

We train two 2.1B models, one per image space. Each is pretrained with the recipe of Section~\ref{sec:setup}, but for 60k instead of 50k updates, and then finetuned for 15k updates at a global batch size of 1,024 on 20.6M text-to-image and visual understanding examples (Appendix~\ref{app:training-stages}). Text-to-image generation uses 25 Euler steps with guidance $w{=}5$; visual question answering uses 16 steps with $w{=}3$, with the parallel decoder reading out all answer positions at once. For joint reasoning, we adapt the finetuned models to each task separately (both models to mazes, the SigLIP2 model to visual spatial planning), read the moves with a linear head trained for each task, and sample without guidance (Section~\ref{sec:joint-reasoning}). All input and output images are $224{\times}224$. We compare with unified models of 1B to 3B parameters; Appendix~\ref{app:extended-eval} includes comparisons with larger models.

\subsection{Image Generation}
\label{sec:generation}

\paragraph{Compositional generation.}
On compositional prompts, both variants are close to the best unified models of 1B to 3B parameters (Table~\ref{tab:generation-comparison}). On GenEval, the SigLIP2 and DINO variants reach 0.87 and 0.88. The DINO variant thus ties TUNA (0.88) for the best result among models of 1B to 3B parameters, and both variants have the highest position score (0.85); TUNA is still ahead on counting and attribute binding. On DPG-Bench, the SigLIP2 variant scores 85.22, which places it above Show-o2 (85.02) and below TUNA (86.03). The number of sampling steps hardly matters: from 25 to 100 steps, GenEval changes by at most 0.003 for either variant (Table~\ref{tab:geneval-steps}).

\begin{table}[!htbp]
\caption{\textbf{Text-to-image generation against unified models of 1B to 3B parameters.} GenEval on the [0,1] scale; DPG-Bench with the official script. JUICE samples $224{\times}224$ images with 25 steps; its size counts the MM-RiT backbone without the frozen encoders and decoders, and baseline sizes are as reported.}
\label{tab:generation-comparison}\label{tab:scale-comparison}
\centering\small\setlength{\tabcolsep}{4pt}\renewcommand{\arraystretch}{1.0}
\begin{tabularx}{\linewidth}{@{}lRRRRRRRR@{}}
\toprule
\multicolumn{9}{@{}l}{\textit{(a) GenEval}$\uparrow$}\\
Model & Size & Single & Two & Count & Color & Position & Attribute & Overall\\
\midrule
Muddit~\citep{muddit} & 1B & 0.98 & 0.72 & 0.54 & 0.82 & 0.19 & 0.41 & 0.61\\
Show-o~\citep{showo} & 1.3B & 0.98 & 0.80 & 0.66 & 0.84 & 0.31 & 0.50 & 0.68\\
UniDisc~\citep{unidisc} & 1.4B & 0.92 & 0.47 & 0.15 & 0.67 & 0.13 & 0.19 & 0.42\\
FUDOKI~\citep{fudoki} & 1.5B & 0.96 & 0.85 & 0.56 & 0.88 & 0.68 & 0.67 & 0.77\\
Janus-Pro-1B~\citep{januspro} & 1.5B & 0.98 & 0.82 & 0.51 & 0.89 & 0.65 & 0.56 & 0.73\\
Show-o2~\citep{showo2} & 1.5B & 0.99 & 0.86 & 0.55 & 0.86 & 0.46 & 0.63 & 0.73\\
Harmon~\citep{harmon} & 1.5B & 0.99 & 0.86 & 0.66 & 0.85 & 0.74 & 0.48 & 0.76\\
TUNA~\citep{tuna} & 1.5B & 1.00 & 0.94 & 0.83 & 0.91 & 0.81 & 0.79 & 0.88\\
D-DiT~\citep{dualdiffusion} & 2.1B & 0.97 & 0.80 & 0.54 & 0.76 & 0.32 & 0.50 & 0.65\\
\midrule
\rowcolor{paperblue!6}JUICE (SigLIP2) & 2.1B & 0.99 & 0.93 & 0.78 & 0.91 & 0.85 & 0.73 & 0.87\\
\rowcolor{paperblue!6}JUICE (DINO) & 2.1B & 0.99 & 0.95 & 0.81 & 0.93 & 0.85 & 0.78 & 0.88\\
\bottomrule
\end{tabularx}

\vspace{6pt}
\begin{tabularx}{\linewidth}{@{}lRRRRRRR@{}}
\toprule
\multicolumn{8}{@{}l}{\textit{(b) DPG-Bench (\%)}$\uparrow$}\\
Model & Size & Global & Entity & Attribute & Relation & Other & Overall\\
\midrule
OneFlow Mixed~\citep{oneflow} & 1B & -- & -- & -- & -- & -- & 80.30\\
FUDOKI~\citep{fudoki} & 1.5B & 80.55 & 89.73 & 88.05 & 93.66 & 78.00 & 83.63\\
Janus-Pro-1B~\citep{januspro} & 1.5B & 87.58 & 88.63 & 88.17 & 88.98 & 88.30 & 82.63\\
Show-o2~\citep{showo2} & 1.5B & 87.53 & 90.38 & 91.34 & 90.30 & 91.21 & 85.02\\
TUNA~\citep{tuna} & 1.5B & 88.87 & 90.32 & 91.71 & 91.79 & 90.14 & 86.03\\
\midrule
\rowcolor{paperblue!6}JUICE (SigLIP2) & 2.1B & 82.67 & 91.41 & 90.44 & 94.58 & 86.80 & 85.22\\
\rowcolor{paperblue!6}JUICE (DINO) & 2.1B & 83.28 & 90.24 & 88.78 & 95.05 & 81.60 & 84.38\\
\bottomrule
\end{tabularx}
\end{table}

Figure~\ref{fig:qualitative-gen} shows selected samples for the same prompts from both variants. Both count small numbers of objects, place objects on the requested side, combine unrelated objects in one scene, and follow a range of artistic styles. The two image spaces often render the same prompt with a different composition and palette.

\begin{figure}[t]
\centering
\setlength{\caseimg}{0.152\linewidth}
\setlength{\tabcolsep}{1.3pt}\renewcommand{\arraystretch}{0.5}
{\small\textbf{(a) Counting, spatial relations, and unusual combinations}}\\[3pt]
\begin{tabular}{@{}c*{6}{c}@{}}
\rotatebox{90}{\makebox[\caseimg][c]{\scriptsize\textbf{\color{sigcolor}SigLIP2}}} & \includegraphics[width=\caseimg]{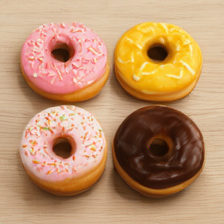} & \includegraphics[width=\caseimg]{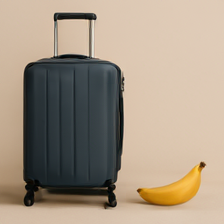} & \includegraphics[width=\caseimg]{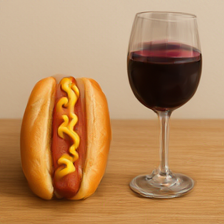} & \includegraphics[width=\caseimg]{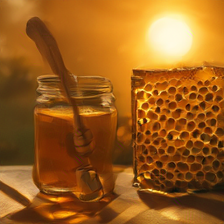} & \includegraphics[width=\caseimg]{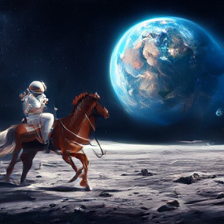} & \includegraphics[width=\caseimg]{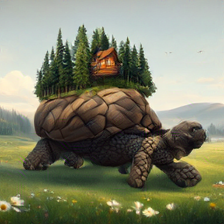}\\
\rotatebox{90}{\makebox[\caseimg][c]{\scriptsize\textbf{\color{dinocolor}DINO}}} & \includegraphics[width=\caseimg]{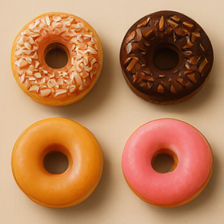} & \includegraphics[width=\caseimg]{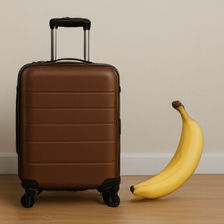} & \includegraphics[width=\caseimg]{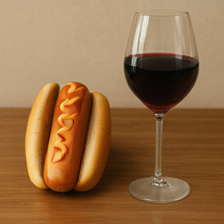} & \includegraphics[width=\caseimg]{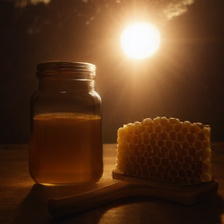} & \includegraphics[width=\caseimg]{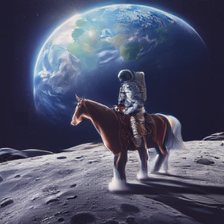} & \includegraphics[width=\caseimg]{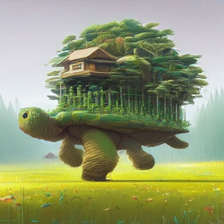}\\
 & \parbox[t]{\caseimg}{\centering\fontsize{6.5}{7.6}\selectfont a photo of four donuts} & \parbox[t]{\caseimg}{\centering\fontsize{6.5}{7.6}\selectfont a photo of a suitcase left of a banana} & \parbox[t]{\caseimg}{\centering\fontsize{6.5}{7.6}\selectfont a photo of a wine glass right of a hot dog} & \parbox[t]{\caseimg}{\centering\fontsize{6.5}{7.6}\selectfont a glass jar of honey with a wooden dipper beside a honeycomb, backlit by the sun} & \parbox[t]{\caseimg}{\centering\fontsize{6.5}{7.6}\selectfont an astronaut riding a horse on the surface of the moon, Earth rising in the background} & \parbox[t]{\caseimg}{\centering\fontsize{6.5}{7.6}\selectfont a giant tortoise carrying a tiny forest with a cabin on its shell, walking across a meadow}\\
\end{tabular}

\vspace{8pt}
{\small\textbf{(b) Artistic styles}}\\[3pt]
\begin{tabular}{@{}c*{6}{c}@{}}
\rotatebox{90}{\makebox[\caseimg][c]{\scriptsize\textbf{\color{sigcolor}SigLIP2}}} & \includegraphics[width=\caseimg]{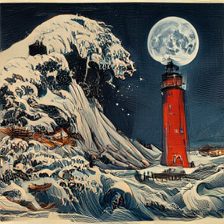} & \includegraphics[width=\caseimg]{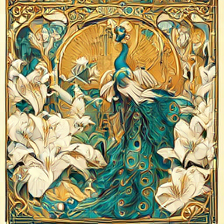} & \includegraphics[width=\caseimg]{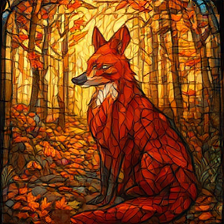} & \includegraphics[width=\caseimg]{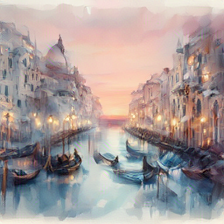} & \includegraphics[width=\caseimg]{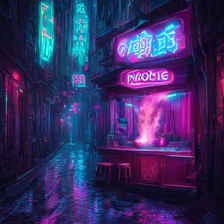} & \includegraphics[width=\caseimg]{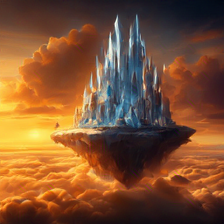}\\
\rotatebox{90}{\makebox[\caseimg][c]{\scriptsize\textbf{\color{dinocolor}DINO}}} & \includegraphics[width=\caseimg]{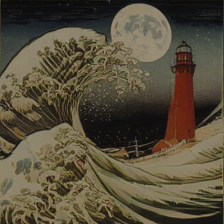} & \includegraphics[width=\caseimg]{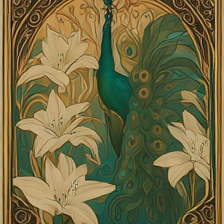} & \includegraphics[width=\caseimg]{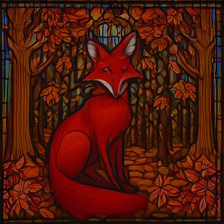} & \includegraphics[width=\caseimg]{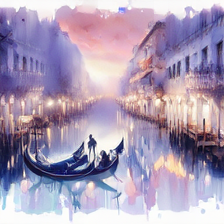} & \includegraphics[width=\caseimg]{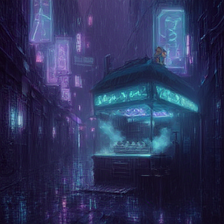} & \includegraphics[width=\caseimg]{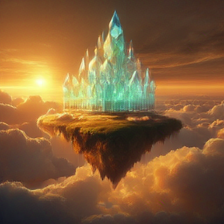}\\
 & \parbox[t]{\caseimg}{\centering\fontsize{6.5}{7.6}\selectfont a ukiyo-e woodblock print of a great wave crashing beside a red lighthouse under a full moon} & \parbox[t]{\caseimg}{\centering\fontsize{6.5}{7.6}\selectfont an art nouveau poster of a peacock among white lilies, gold and teal palette, ornate border} & \parbox[t]{\caseimg}{\centering\fontsize{6.5}{7.6}\selectfont a stained-glass window depicting a red fox in an autumn forest} & \parbox[t]{\caseimg}{\centering\fontsize{6.5}{7.6}\selectfont a loose watercolor of a Venetian canal at dusk with gondolas, soft pastel wash} & \parbox[t]{\caseimg}{\centering\fontsize{6.5}{7.6}\selectfont a rain-soaked cyberpunk alley with holographic signs and a steaming noodle stand, teal and magenta light} & \parbox[t]{\caseimg}{\centering\fontsize{6.5}{7.6}\selectfont a crystal castle on a floating island above the clouds at golden hour, matte painting}\\
\end{tabular}
\caption{\textbf{Text-to-image samples from both variants for the same prompts.} (a) Counting, spatial relations, and unusual combinations of objects. (b) Prompts that ask for a specific artistic style. All samples use guidance $w{=}5$ with the generic caption reference; Appendix~\ref{app:qualitative-generation} gives more samples.}
\label{fig:qualitative-gen}
\end{figure}

\begin{table}[!htbp]
\caption{\textbf{Text-to-image generation with world knowledge on WISE.} WiScore under the original protocol (GPT-4o-2024-05-13 judge); Overall weights categories by prompt count. JUICE is scored on two images per prompt, and its 95\% confidence interval on Overall is about $\pm0.02$. JUICE uses the same guidance as on GenEval ($w{=}5$, generic caption reference). Baseline values are from the WISE paper~\citep{wise} or the models' own papers; OneFlow reports only the cultural category.}
\label{tab:wise}
\centering\small\setlength{\tabcolsep}{4pt}\renewcommand{\arraystretch}{1.0}
\begin{tabularx}{\linewidth}{@{}lRRRRRRRR@{}}
\toprule
Model & Size & Cult. & Time & Space & Bio. & Phys. & Chem. & Overall\\
\midrule
\rowcolor{black!6}\multicolumn{9}{c}{\textit{Unified models of 1B to 3B parameters}}\\
OneFlow~\citep{oneflow} & 1B & 0.63 & -- & -- & -- & -- & -- & --\\
JanusFlow~\citep{janusflow} & 1.3B & 0.13 & 0.26 & 0.28 & 0.20 & 0.19 & 0.11 & 0.18\\
Show-o ($512^2$)~\citep{showo} & 1.3B & 0.28 & 0.40 & 0.48 & 0.30 & 0.46 & 0.30 & 0.35\\
Janus-Pro-1B~\citep{januspro} & 1.5B & 0.20 & 0.28 & 0.45 & 0.24 & 0.32 & 0.16 & 0.26\\
Show-o2~\citep{showo2} & 1.5B & 0.30 & 0.38 & 0.47 & 0.32 & 0.42 & 0.28 & 0.35\\
Harmon~\citep{harmon} & 1.5B & 0.38 & 0.48 & 0.52 & 0.37 & 0.44 & 0.29 & 0.41\\
\rowcolor{black!6}\multicolumn{9}{c}{\textit{Larger references}}\\
Janus-Pro-7B~\citep{januspro} & 7B & 0.30 & 0.37 & 0.49 & 0.36 & 0.42 & 0.26 & 0.35\\
BAGEL~\citep{bagel} & 7B/14B & 0.44 & 0.55 & 0.68 & 0.44 & 0.60 & 0.39 & 0.52\\
FLUX.1-dev (T2I)~\citep{flux} & 12B & 0.48 & 0.58 & 0.62 & 0.42 & 0.51 & 0.35 & 0.50\\
\midrule
\rowcolor{paperblue!6}JUICE (SigLIP2) & 2.1B & 0.41 & 0.48 & 0.62 & 0.35 & 0.51 & 0.33 & 0.45\\
\rowcolor{paperblue!6}JUICE (DINO) & 2.1B & 0.31 & 0.43 & 0.59 & 0.29 & 0.42 & 0.27 & 0.37\\
\bottomrule
\end{tabularx}
\end{table}

\begin{figure}[t]
\centering
\setlength{\caseimg}{0.152\linewidth}
\setlength{\tabcolsep}{1.3pt}\renewcommand{\arraystretch}{0.5}
\begin{tabular}{@{}c*{6}{c}@{}}
 & {\scriptsize\textsc{Culture}} & {\scriptsize\textsc{Time}} & {\scriptsize\textsc{Space}} & {\scriptsize\textsc{Biology}} & {\scriptsize\textsc{Physics}} & {\scriptsize\textsc{Chemistry}} \\[3pt]
\rotatebox{90}{\makebox[\caseimg][c]{\scriptsize\textbf{\color{sigcolor}SigLIP2}}} & \includegraphics[width=\caseimg]{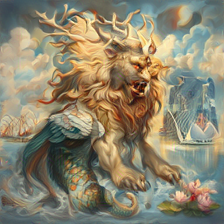} & \includegraphics[width=\caseimg]{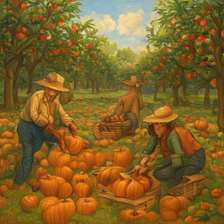} & \includegraphics[width=\caseimg]{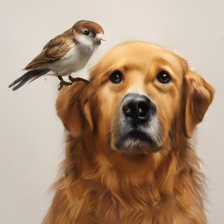} & \includegraphics[width=\caseimg]{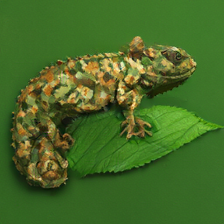} & \includegraphics[width=\caseimg]{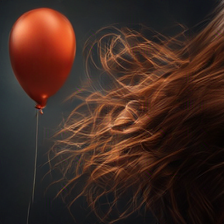} & \includegraphics[width=\caseimg]{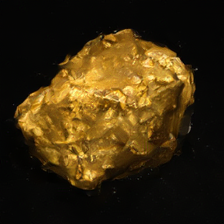} \\
\rotatebox{90}{\makebox[\caseimg][c]{\scriptsize\textbf{\color{dinocolor}DINO}}} & \includegraphics[width=\caseimg]{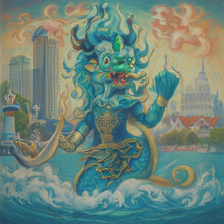} & \includegraphics[width=\caseimg]{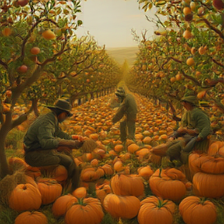} & \includegraphics[width=\caseimg]{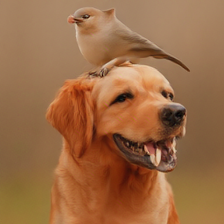} & \includegraphics[width=\caseimg]{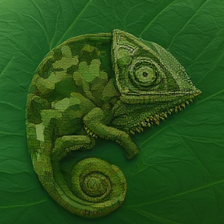} & \includegraphics[width=\caseimg]{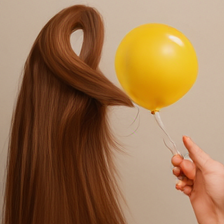} & \includegraphics[width=\caseimg]{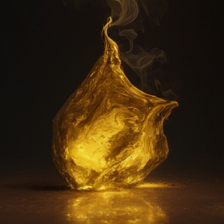} \\
 & \parbox[t]{\caseimg}{\centering\fontsize{6.5}{7.6}\selectfont A mythical creature embodying Singapore's origins} & \parbox[t]{\caseimg}{\centering\fontsize{6.5}{7.6}\selectfont The apple orchard when the farmers are harvesting pumpkins.} & \parbox[t]{\caseimg}{\centering\fontsize{6.5}{7.6}\selectfont Generate an image of a bird and a dog, with the smaller animal on top and the larger below} & \parbox[t]{\caseimg}{\centering\fontsize{6.5}{7.6}\selectfont A chameleon perfectly camouflaged against a green leaf} & \parbox[t]{\caseimg}{\centering\fontsize{6.5}{7.6}\selectfont The balloon that has been rubbed comes into contact with long hair} & \parbox[t]{\caseimg}{\centering\fontsize{6.5}{7.6}\selectfont A piece of gold that has been exposed to oxygen for decades} \\[2pt]
 & \parbox[t]{\caseimg}{\centering\fontsize{6.0}{7.0}\selectfont\color{black!60} Singapore's founding legend $\rightarrow$ the Merlion, a lion with a fish body.} & \parbox[t]{\caseimg}{\centering\fontsize{6.0}{7.0}\selectfont\color{black!60} Pumpkin harvest $\rightarrow$ autumn; apples still on the trees.} & \parbox[t]{\caseimg}{\centering\fontsize{6.0}{7.0}\selectfont\color{black!60} The bird is the smaller animal $\rightarrow$ the bird on top, the dog below.} & \parbox[t]{\caseimg}{\centering\fontsize{6.0}{7.0}\selectfont\color{black!60} Camouflage $\rightarrow$ skin color and pattern match the leaf.} & \parbox[t]{\caseimg}{\centering\fontsize{6.0}{7.0}\selectfont\color{black!60} Rubbing charges the balloon $\rightarrow$ the hair is pulled toward it.} & \parbox[t]{\caseimg}{\centering\fontsize{6.0}{7.0}\selectfont\color{black!60} Gold does not oxidize $\rightarrow$ still bright, no corrosion.} \\
\end{tabular}
\caption{\textbf{WISE prompts that require world knowledge.} Each prompt names what to draw only indirectly; the gray line gives the implied knowledge (abridged from the WISE explanation). The prompt alone is the condition, with guidance $w{=}5$.}
\label{fig:wise-cases}
\end{figure}

\paragraph{Generation with world knowledge.}
WISE~\citep{wise} checks whether generated images reflect the world knowledge a prompt implies, with cultural, temporal, spatial, biological, physical, and chemical categories (Table~\ref{tab:wise}). At the guidance scale used for GenEval, the SigLIP2 variant has a WiScore of 0.45. This is the highest overall score among unified models of 1B to 3B parameters that report one (which reach at most 0.41) and above Janus-Pro-7B (0.35), though below FLUX.1-dev (0.50) and BAGEL (0.52); OneFlow reports only the cultural category, where it scores higher. WISE, unlike GenEval, separates the two image spaces: SigLIP2 scores 0.45 and DINO 0.37. The advantage that Section~\ref{sec:representation} found for SigLIP2 in captioning thus carries over to generation that needs world knowledge. The difference is largest for cultural prompts (0.41 vs.\ 0.31) and smallest for spatial ones (0.62 vs.\ 0.59), where the SigLIP2 variant is level with FLUX.1-dev. Its lead over Harmon, the strongest baseline of 1B to 3B parameters, comes mostly from spatial and physical prompts, and it is slightly lower on biology (0.35 vs.\ 0.37). Since \method{} has no language model backbone, this knowledge comes from its frozen encoders and its training data. A lower guidance scale with the time swap reference raises both scores, to 0.49 and 0.42, respectively (Appendix~\ref{app:guidance-ablation}). Figure~\ref{fig:wise-cases} shows one example per category.

\subsection{Visual Understanding}
\label{sec:understanding}

\begin{table}[!htbp]
\caption{\textbf{Visual understanding against unified models of 1B to 3B parameters.} Init.\ gives the backbone initialization and Res.\ the input resolution (LLM: pretrained language model; T2I: pretrained text-to-image model). Baseline values and sizes are from each model's paper; the size of JUICE counts the MM-RiT backbone without the frozen encoders and decoders (Appendix~\ref{app:total-parameters}). $^\dagger$The paper reports an unlabeled MME score, which may be the sum of perception and cognition. MME-P: MME perception score; MMB: MMBench; RWQA: RealWorldQA. JUICE uses VQAv2 test-dev and MMBench dev; Appendix~\ref{app:extended-eval} gives all 10 benchmarks.}
\label{tab:understanding-comparison}
\centering\small\setlength{\tabcolsep}{3.4pt}
\begin{tabularx}{\linewidth}{@{}>{\raggedright\arraybackslash}Xccc*{8}{r}@{}}
\toprule
Model & Size & Init. & Res. & VQAv2 & GQA & POPE & \mbox{MME-P} & MMB & AI2D & MMMU & RWQA\\
\midrule
Muddit~\citep{muddit} & 1B & T2I & $512^2$ & 68.2 & 57.5 & -- & 1107.4$^\dagger$ & -- & -- & 27.6 & --\\
OneFlow~\citep{oneflow} & 1B & LLM & $512^2$ & 67.7 & 57.8 & 85.7 & 1497.1$^\dagger$ & 69.0 & 58.5 & 29.8 & 50.6\\
Show-o~\citep{showo} & 1.3B & LLM & $256^2$ & 64.7 & 54.2 & 76.2 & 1014.9$^\dagger$ & -- & -- & 25.0 & --\\
Show-o~\citep{showo} & 1.3B & LLM & $512^2$ & 69.4 & 58.0 & 80.0 & 1097.2$^\dagger$ & -- & -- & 26.7 & --\\
JanusFlow~\citep{janusflow} & 1.3B & LLM & $384^2$ & 79.8 & 60.3 & 88.0 & 1333.1 & 74.9 & -- & 29.3 & --\\
Harmon~\citep{harmon} & 1.5B & LLM & $512^2$ & -- & 58.9 & 87.6 & 1155.0 & 65.5 & -- & 38.9 & --\\
FUDOKI~\citep{fudoki} & 1.5B & LLM & $384^2$ & -- & 57.6 & 86.1 & 1485.4 & 73.9 & -- & 34.3 & --\\
Janus-Pro-1B~\citep{januspro} & 1.5B & LLM & $384^2$ & -- & 59.3 & 86.2 & 1444.0 & 75.5 & -- & 36.3 & --\\
Show-o2~\citep{showo2} & 1.5B & LLM & $432^2$ & -- & 60.0 & -- & 1450.9 & 67.4 & 69.0 & 37.1 & --\\
TUNA~\citep{tuna} & 1.5B & LLM & $512^2$ & -- & 61.4 & -- & 1461.5 & -- & 71.4 & 39.1 & 62.5\\
D-DiT~\citep{dualdiffusion} & 2.1B & T2I & $256^2$ & 59.5 & 55.1 & 79.2 & 897.5$^\dagger$ & -- & -- & -- & --\\
D-DiT~\citep{dualdiffusion} & 2.1B & T2I & $512^2$ & 60.1 & 59.2 & 84.0 & 1124.7$^\dagger$ & -- & -- & -- & --\\
\midrule
\rowcolor{paperblue!6}\textbf{JUICE (SigLIP2)} & 2.1B & None & $224^2$ & 73.1 & 58.6 & 84.0 & 1115.7 & 66.0 & 34.4 & 29.6 & 53.6\\
\rowcolor{paperblue!6}\textbf{JUICE (DINO)} & 2.1B & None & $224^2$ & 68.8 & 55.7 & 82.3 & 960.2 & 55.8 & 31.9 & 31.5 & 40.5\\
\bottomrule
\end{tabularx}
\end{table}

On visual question answering, \method{} is competitive on several perception benchmarks and weak on those that need domain knowledge (Table~\ref{tab:understanding-comparison}). All other models in the table start from a pretrained language or text-to-image model, whereas \method{} trains MM-RiT from scratch and sees images at $224{\times}224$.
Its POPE (84.0) and MME-P (1115.7) scores are still below those of most models initialized from LLMs. On VQAv2, the SigLIP2 variant reaches 73.1. That is above Show-o at $512^2$ (69.4), Muddit (68.2), and OneFlow (67.7), and below JanusFlow (79.8). Its GQA (58.6) and MMBench (66.0) scores are close to those of Harmon and Show-o2.
The weak side is domain knowledge. On AI2D, \method{} scores 34.4, far below models initialized from LLMs (58.5 to 71.4). On MMMU (29.6), it is above Show-o and Muddit, comparable to JanusFlow and OneFlow, and below Harmon, FUDOKI, Janus-Pro, Show-o2, and TUNA (34.3 to 39.1).
This is what one would expect without a language model backbone: visual recognition transfers through the pretrained image space, but the knowledge that models initialized from LLMs inherit from text pretraining is missing.

\begin{figure}[t]
\centering
\qacell{}{Attribute}{GQA}{Which type of material is used to make the trash bin below the pepper shaker?}{plastic}{plastic}{plastic}\hfill
\qacell{}{Counting}{VQAv2}{How many cows are there?}{4}{4}{4}\hfill
\qacell{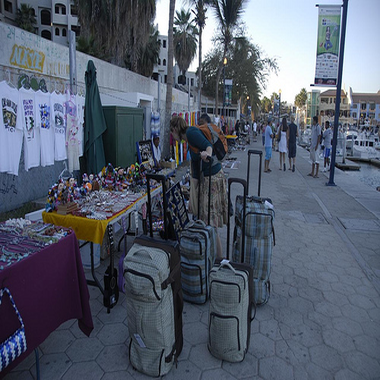}{Spatial relation}{GQA}{Is the green umbrella to the right or to the left of the person who is to the left of the palm tree?}{left}{left}{left}\hfill
\qacell{}{Spatial relation}{GQA}{Are the thick books to the left or to the right of the computer that is to the right of the lamps?}{right}{right}{right}\\[8pt]
\qacell{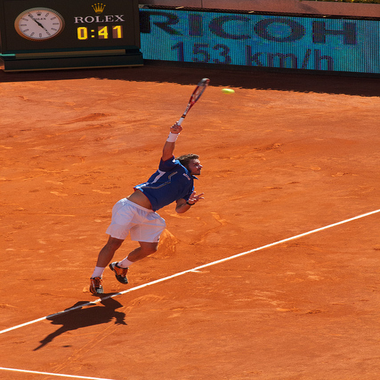}{Action}{GQA}{What's the player doing?}{jumping}{jumping}{jumping}\hfill
\qacell{}{Multiple entities}{GQA}{Which material is the clock that is to the left of the tennis ball made of, stainless steel or brick?}{stainless steel}{stainless steel}{stainless steel}\hfill
\qacell{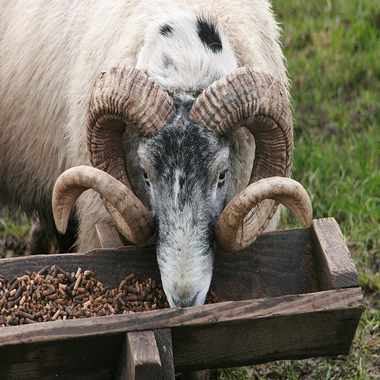}{Outside knowledge}{OK-VQA}{What is a male of this animal called?}{ram}{ram}{ram}\hfill
\qacell{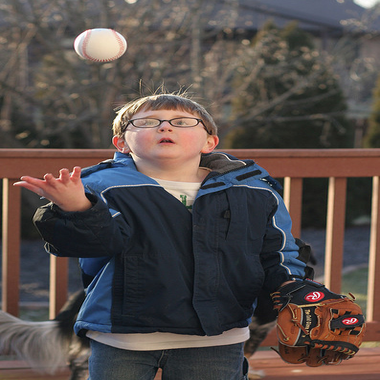}{Object existence}{POPE}{Is there a baseball bat in the image?}{no}{no}{no}
\caption{\textbf{Selected questions of different types, answered by both variants.} Each panel gives the question type and benchmark, the question, both answers, and the reference answer. Appendix~\ref{app:qualitative} gives more examples, including captions and long answers.}
\label{fig:qual-understanding-main}
\end{figure}

The choice of image space matters: SigLIP2 leads DINO on every benchmark except MMMU (Table~\ref{tab:understanding-full}), by 4.4 points on VQAv2 and 10.2 points on MMBench, again in line with Section~\ref{sec:representation}. Appendix~\ref{app:extended-eval} reports the remaining benchmarks (OK-VQA and ScienceQA-IMG) and comparisons with larger models.
Figure~\ref{fig:qual-understanding-main} shows eight selected questions of seven types, from attributes and counting to chained spatial relations, outside knowledge, and object existence, which both variants answer correctly; both also produce fluent, concise captions (Appendix~\ref{app:qualitative-captioning}). Longer outputs are less reliable. The detailed descriptions in Appendix~\ref{app:qualitative-long-captioning} get the main objects, colors, and layout right but contain grammatical errors and the occasional invented word, in line with the lexical cost seen in Section~\ref{sec:paradigm}.

Reading out the answer states in parallel loses little: on VQA with open answers, an autoregressive decoder on the same states is within 1.5 points but 2.1 to 6.4 times slower on these short answers. It is much worse on most multiple choice benchmarks (45.3 vs.\ 66.0 on MMBench), but only the parallel decoder saw multiple choice answers in training (Appendix~\ref{app:text-decoder}).

\subsection{Joint Reasoning}
\label{sec:joint-reasoning}

Some tasks need an image and a text output that are each correct and that agree, such as a route drawn on a map and the moves that describe it. When several answers are valid, being correct is not enough: the two outputs must also commit to the same answer. Sampling each modality from its own marginal can pick a different valid answer in each, which we call a cross-modal mode mismatch. A model of the joint distribution can instead produce both outputs together, with each modality conditioning on the other's evolving state (coupled sampling; Section~\ref{sec:inference}). We first look at unconditional joint generation and then turn to two planning tasks, in which the input is a map image and a text question and the model generates the map with the route drawn on it together with the move sequence.

\paragraph{Joint generation.}
Without any condition, both streams start from noise and the model generates an image and a caption together (selected samples in Figure~\ref{fig:qual-joint-main}). In these samples, the caption describes the image it is generated with, so the two streams agree even though neither is given. Planning tasks test the harder case, in which the outputs must also be correct.

\begin{figure}[t]
\centering
\jointcell{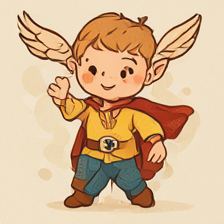}{\textcolor{sigcolor}{SigLIP2 + T5}}{The image depicts a charming, cartoon-style character that appears to be a young boy. The boy has light brown hair, adorned with wing-like wings and stands out backward with a cheerful expression. He is dressed in a yellow tunic over a yellow shirt, paired with blue pants that have dotted patterns, a red cape and brown boots. He has a brown belt with a design on it. The character is also showing a peace sign on his forehead and is depicted in a playful, mischievous pose.}\hfill
\jointcell{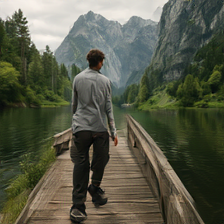}{\textcolor{sigcolor}{SigLIP2 + T5}}{The image depicts a person standing on the end of a wooden bridge, facing away from the camera, towards a stunning mountainous landscape. The individual is dressed in outdoor attire, including a light gray long-sleeved shirt, dark pants, and black shoes. The bridge appears to be made of well-maintained, with lush green trees and grass on both sides. The landscape reflects the surrounding scenery, including the rugged mountains and overcast sky above.}\hfill
\jointcell{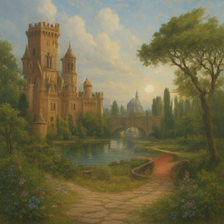}{\textcolor{dinocolor}{DINO + T5}}{The image depicts a tranquil and picturesque park scene with an old European castle in the background, characterized by its multiple levels, turrets, and golden domes. The sky is partly cloudy, suggesting it might be late afternoon or early evening. A stone bridge arches over a body of water adorned with greenery. Surrounding the water are lush green trees and flowers, and a gentle red path winds through the center. The overall atmosphere is serene and idyllic, evoking a sense of history and isolation.}\hfill
\jointcell{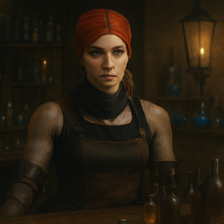}{\textcolor{dinocolor}{DINO + T5}}{The image showcases a character that appears to be from a fantasy setting, characterized by a striking use of leather arm bands and a red headscarf. The character has neatly braided hair and is seated at a bar, surrounded by wooden shelves and various bottles filled with beer. The individual is also wearing a black sleeveless top, which adds to the sturdy and rugged aesthetic. The background reveals more shelves and other patrons with blue bottles, suggesting a typical setting in a beer shop.}
\caption{\textbf{Joint samples of images and text.} With no condition, both streams start from noise and are sampled together (64 steps, no guidance). Text is read out with the autoregressive caption decoder (Appendix~\ref{app:text-decoder}) and shown up to the point where it starts to repeat. Appendix~\ref{app:qualitative-joint} gives more samples, and Figure~\ref{fig:qual-joint-trajectory} shows how both streams form during sampling.}
\label{fig:qual-joint-main}
\end{figure}

\paragraph{Mazes with multiple solutions.}
Our grid mazes have two to sixteen shortest routes each~\citep{ivanitskiy2023configurable}, so both outputs can be valid and still disagree. For each training pair, one of these routes is picked at random. At evaluation, any shortest route counts, and a joint success (J) needs both outputs to be valid and to show the same route (Appendix~\ref{app:joint-reasoning}). We adapt the finetuned 2.1B models for 5k updates and compare samplers at an equal number of backbone calls (Table~\ref{tab:maze-specialists}).

\begin{table}[!htbp]
\caption{\textbf{Joint success on mazes with multiple solutions.} J: both outputs are shortest routes and describe the same route. Samplers are compared at 16 backbone calls; mean of two sampling seeds. Lower rows give the route validity of models finetuned on a single modality.}
\label{tab:maze-specialists}
\centering
\small
\setlength{\tabcolsep}{3pt}\renewcommand{\arraystretch}{1.0}
\begin{tabular*}{0.6\linewidth}{@{\extracolsep{\fill}}llrr@{}}
\toprule
Sampler & Order & SigLIP2 & DINO\\
\midrule
Coupled & none & 93.7 & 81.7\\
Image first & image $\to$ text & \textbf{95.2} & \textbf{82.7}\\
Text first & text $\to$ image & 85.3 & 59.9\\
Independent & none & 54.6 & 42.7\\
\midrule
\multicolumn{4}{@{}l}{\textit{Validity of one modality alone}}\\
Image FT only & -- & 95.1 & 83.3\\
Text FT only & -- & 86.5 & 64.6\\
\bottomrule
\end{tabular*}
\end{table}

\begin{figure}[!htbp]
\centering
\includegraphics[width=\linewidth]{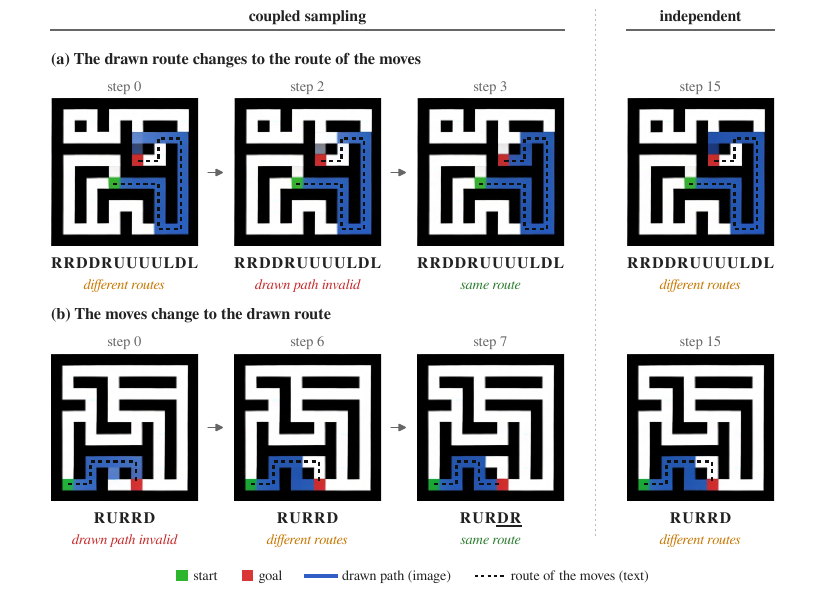}
\caption{\textbf{Coupled sampling brings the two outputs onto one route; independent sampling does not.} Two mazes under coupled sampling at three of its 16 steps per stream, and under independent sampling with the same 16 steps per stream (32 backbone calls, twice the budget of Table~\ref{tab:maze-specialists}) at its last step; both samplers start from the same state at step~0, and the coupled outputs do not change after the last step shown. The blue path is the decoded image, and the dashed line traces the decoded moves from the start. Underlined moves differ from the previous column.}
\label{fig:maze-cases}
\end{figure}

Coupled sampling comes close to the better sequential order without having to know which order that is. If each modality is sampled without seeing the other (independent sampling), joint success is only 54.6\% for SigLIP2 and 42.7\% for DINO, since the two streams often settle on different routes. Sequential sampling avoids this disagreement, but then the order matters, because mistakes in the first modality are passed on to the second. Image first and then text reaches 95.2\% and 82.7\%; the reverse order reaches 85.3\% and 59.9\%. The better order starts with the more reliable modality, here the image, whose validity alone is higher than that of the text. For a new task, though, which modality is more reliable is not known in advance. Coupled sampling needs no order. It reaches 93.7\% and 81.7\%, within 1.5 and 1.0 points of the better order and 8.4 and 21.8 points above the worse one, and it loses at most 0.3 points when the budget is halved to 8 calls (Appendix~\ref{app:maze-sampling}). These numbers are means over two seeds. For SigLIP2, the gap to image first is 0.6 points for one seed and 2.4 for the other, and a paired test still finds image first better ($p{=}0.017$; Appendix~\ref{app:maze-sampling}).

Coupled sampling does not reach agreement by collapsing onto one canonical route. On the mazes it solves, coupled sampling returns the stored reference route in 43.5\% and 47.0\% of cases, close to the 44.1\% and 44.4\% expected if it chose uniformly among the shortest routes, so the model picks among valid routes. Figure~\ref{fig:maze-cases} follows two mazes, with both samplers starting from the same state. Under coupled sampling, whichever output settles later changes to match the other: the drawn route in one maze, the moves in the other. Independent sampling ends with two valid but different routes. Because either modality can be the one that changes, coupled sampling is not simply copying one modality into the other, which fits with the fact that it does not need to know in advance which modality is more reliable.

\begin{table}[!t]
\caption{\textbf{Visual spatial planning on VSP.} Success rate under the official rule; levels 3 to 6 match the training map sizes, and levels 7 and 8 are larger maps unseen in training. Baseline values are from LatentUM~\citep{latentum}, which reports InternVL3.5 and ThinkMorph only as averages; ThinkMorph does not state which levels its average covers. Task FT marks finetuning on maps in the style of VSP: JUICE uses 55.9k maps, ThinkMorph and coarse LatentUM 6k, and fine LatentUM 18k problems solved step by step.}
\label{tab:vsp}
\centering
\small
\setlength{\tabcolsep}{2.5pt}\renewcommand{\arraystretch}{1.0}
\begin{tabularx}{\linewidth}{@{}lRCRRRRRRR@{}}
\toprule
Model & Size & Task FT & Lv3 & Lv4 & Lv5 & Lv6 & Avg. & Lv7 & Lv8\\
\midrule
InternVL3.5~\citep{internvl35} & 38B & \ding{55} & -- & -- & -- & -- & 0.20 & -- & --\\
ThinkMorph~\citep{thinkmorph} & 7B & \ding{51} & -- & -- & -- & -- & 0.76 & -- & --\\
LatentUM, coarse~\citep{latentum} & 4B & \ding{51} & 1.00 & 0.85 & 0.83 & 0.71 & 0.85 & -- & --\\
LatentUM, fine~\citep{latentum} & 4B & \ding{51} & 1.00 & 1.00 & 1.00 & 0.97 & 0.99 & -- & --\\
\midrule
\rowcolor{paperblue!6}JUICE (SigLIP2) & 2.1B & \ding{51} & 1.00 & 1.00 & 1.00 & 0.99 & 0.998 & 0.94 & 0.69\\
\bottomrule
\end{tabularx}
\end{table}

\begin{figure}[!htbp]
\centering
\includegraphics[width=\linewidth]{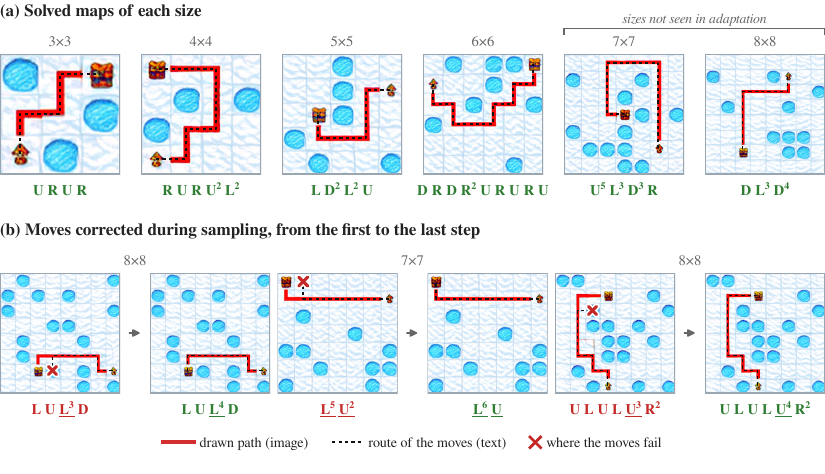}
\caption{\textbf{VSP cases.} Adaptation uses sizes $3{\times}3$ to $6{\times}6$. (a)~One solved map of each size. (b)~Maps whose moves fail at the first of 16 sampling steps and solve the map at the last, with the drawn path fixed throughout. The red path is the decoded image, and the dashed line traces the decoded moves from the start; a cross marks where the moves fail, in a hole or short of the goal. A repeated move is written with its count ($\mathrm{L}^4$ for LLLL), and underlined runs change between the two steps. Decoded images are matched in color to the input rendering.}
\label{fig:vsp-cases}
\end{figure}

\paragraph{Visual spatial planning.}
VSP~\citep{vsp} asks for a path through a FrozenLake grid from the start to the goal without stepping into a hole; any such path counts as a success. We adapt the finetuned SigLIP2 model for 5k updates on 55.9k maps of sizes $3{\times}3$ to $6{\times}6$, synthesized by us or drawn from ThinkMorph~\citep{thinkmorph}, with all benchmark maps removed, and evaluate on the official 100 maps per level (Table~\ref{tab:vsp}). \method{} solves 99.8\% of maps at levels 3 to 6, on par with the fine variant of LatentUM (99\%), which plans visually step by step, and above its coarse variant (85\%), ThinkMorph (76\%), and InternVL3.5-38B (20\%). On larger map sizes not seen in training, it solves 94\% at level 7 and 69\% at level 8. Its image and text outputs agree on 99.6\% of our own test maps. Appendix~\ref{app:vsp} describes the setup and the baselines.

On the unseen sizes, the drawn path is right more often than the moves, although the image test is the stricter one. At levels 7 and 8, the drawn path is a shortest route on 99\% and 96\% of maps, while the moves reach the goal on 94\% and 69\% (Figure~\ref{fig:vsp-cases}a shows a solved map of each size). Of the 37 failures there, 29 lose one move from the reference route while the drawn path is exactly that route. The rate of lost moves grows with the length of the run of repeated moves, from under 2\% of runs of one or two moves to 7 of the 9 runs of six, a length that a shortest route cannot have on the sizes used in adaptation. The outputs mostly settle early: on 574 of the 600 maps, both are fixed from the first of 16 steps. Coupled sampling can still restore a lost move (Figure~\ref{fig:vsp-cases}b). On nine maps the moves fail at the first step and solve the map at the end, seven of them with the drawn path fixed throughout, and no map solved at the first step ends unsolved. On the unseen sizes, as on mazes, the image is the more reliable modality, and these text errors match the cost in exact wording that Section~\ref{sec:paradigm} finds for continuous text.

\section{Conclusion}
\label{sec:discussion}

\method{} models the joint distribution of images and text within frozen pretrained representation spaces, using a single flow matching backbone. Because each modality has its own noise level, image generation, visual understanding, and joint generation become paths through one square of noise levels, all handled by the same model. Our pretraining studies point to four principles. Text can be generated as continuous states at little cost to image generation, which stays close to that of a model trained only for text-to-image generation; what suffers is mainly exact wording. Training needs both the edges of the square, where sampling is conditioned on clean context, and the interior, which joint generation passes through. Semantic encoder spaces are better generation targets than spaces built for reconstruction. And a preference between image spaces found at one scale carries over across 179M to 2.1B parameters for captioning but not for generation. At 2.1B parameters and without language model initialization, the backbone is competitive in compositional image generation, in generation that needs world knowledge, and, on several perception benchmarks, in visual understanding. Since the model learns the joint distribution, it can also draw a route and write the matching moves together. On mazes with multiple solutions, coupled sampling makes the two agree without a fixed generation order, and on visual spatial planning it solves nearly all maps of the training sizes. In such a model, the order in which modalities are generated is no longer fixed by the architecture: it becomes a choice at sampling time, and coupled sampling can leave it open. Pretrained continuous spaces thus let one model understand, generate, and coordinate images and text without putting either one first.

\textbf{Limitations and Future Work.}
The models run at a native $224{\times}224$ resolution, which limits fine visual detail. Exact wording lags behind autoregressive text, and without a language model backbone the models trail most of those initialized from LLMs on benchmarks that need domain knowledge, such as AI2D and MMMU. Each pretraining configuration was run once, and the joint reasoning tasks are synthetic planning problems. The next steps follow from these limits: higher resolution, more accurate continuous text, and joint generation on real tasks in which several pairs of image and text outputs are valid.

\FloatBarrier
\endgroup
\section*{Acknowledgments}
\addcontentsline{toc}{section}{Acknowledgments}
This work was supported by the Institute of Information \& Communications Technology Planning \& Evaluation (IITP) with a grant funded by the Ministry of Science and ICT (MSIT) of the Republic of Korea in connection with the Global AI Frontier Lab International Collaborative Research Agreement. (No.~RS-2024-00469482 \& RS-2024-00509279) This work was partly supported in part by the NYUAD Center for Interdisciplinary Data Science \& AI (CIDSAI), funded by Tamkeen under the NYUAD Research Institute Award CG016.
\raggedbottom
\begingroup
\interlinepenalty=10000
\phantomsection\addcontentsline{toc}{section}{References}
\bibliography{references}
\endgroup
\bibliographystyle{unsrtnat}
\clearpage
\appendix
\section{Network details}
\label{app:details}

\subsection{Representation encoders}

The pretrained encoders are never updated. Table~\ref{tab:encoder-inputs} lists the four image inputs. SigLIP2~\citep{siglip2} and WebSSL-DINO~\citep{webssl} give us their last patch states (we drop DINO's CLS token), which we pass through LayerNorm per token without learned parameters. The VAE input is the posterior mode after the VAE's own shift and scale, with each $2{\times}2$ latent neighborhood packed into one token. In all four cases the result is a $16{\times}16$ grid of image tokens. A learned linear projection feeds frozen latent features into MM-RiT. Pixels go through a trainable patch stem instead: a convolution with patch stride, then a $1{\times}1$ projection to the backbone width.

\begin{table}[!htbp]
\centering\small
\setlength{\tabcolsep}{4pt}
\renewcommand{\arraystretch}{1.13}
\caption{\textbf{Image encoder sources and token shapes.}}
\label{tab:encoder-inputs}
\begin{tabularx}{\linewidth}{@{}lXc@{}}
\toprule
Input & Frozen source and extraction & Tokens $\times$ channels \\
\midrule
SigLIP2 & \href{https://huggingface.co/google/siglip2-so400m-patch14-224}{SigLIP2-SO400M/14}, $224^2$, final patch states & $256\times1152$ \\
DINO & \href{https://huggingface.co/facebook/webssl-dino300m-full2b-224}{WebSSL-DINO300M-full2b}, $224^2$, final patch states & $256\times1024$ \\
VAE & \href{https://huggingface.co/black-forest-labs/FLUX.1-schnell}{FLUX.1-schnell VAE}, $256^2$, packed latent mode & $256\times64$ \\
Pixel & None; $256^2$ RGB in $[-1,1]$, $16\times16$ patches & $256\times768$ \\
\bottomrule
\end{tabularx}
\end{table}

For text, we use contextual states at the layers in Table~\ref{tab:text-encoder-layers}, then apply the representation normalization before the shared backbone. Indices refer to the returned hidden state list, with index 0 denoting the embedding input. Qwen3-VL-Embedding uses its text tower; we remove the fixed chat prefix of 14 tokens from the returned states. The SigLIP2 text tower uses lowercased input and a length of 64 tokens with unmasked encoder attention (its max length is 64); the other arms use a length of 128 tokens.

\begin{table}[!htbp]
\centering\small
\setlength{\tabcolsep}{5pt}
\caption{\textbf{Selected text encoder states.}}
\label{tab:text-encoder-layers}
\begin{tabularx}{\linewidth}{@{}Xccc@{}}
\toprule
Frozen encoder & Layer & Positions & Channels \\
\midrule
Flan-T5-Small~\citep{flant5} & Final (8) & 128 & 512 \\
Qwen3-Embedding-0.6B~\citep{qwen3embedding} & 14 & 128 & 1,024 \\
Qwen3-VL-Embedding-2B~\citep{qwen3vlemb} & 14 & 128 & 2,048 \\
SigLIP2-SO400M/14 text~\citep{siglip2} & 13 & 64 & 1,152 \\
\bottomrule
\end{tabularx}
\end{table}
 
\subsection{Representation normalization}
\label{app:normalization}
The transformation $N_m$ in Section~\ref{sec:method-representations} defines the spaces in which diffusion noise is added.
Its form depends on the encoder (Table~\ref{tab:normalization}).

For the RAE representations, we use LayerNorm per token without affine parameters~\citep{scalingrae,tuna2}. The VAE uses its supplied shift and scale, whereas pixels are scaled to $[-1,1]$ before patchification~\citep{blm,tuna,showo}.

We divide Flan-T5 states by a fixed scalar, $z=h/0.2$, with no centering and no scaling per channel. The other text encoders are standardized per channel. For each of them, we compute a mean and a standard deviation for every feature channel over valid text positions in the training data, keep these statistics fixed, and use the same transformation in training and at inference. All token positions share the statistics, and no extra scalar divisor is applied.
 
\begin{table}[!htbp]
\centering\small
\setlength{\tabcolsep}{5pt}
\caption{\textbf{Representation normalization before diffusion.} $h$ denotes an encoder output, $c$ a feature channel, and $(b_{\mathrm{VAE}},a_{\mathrm{VAE}})$ the shift and scale supplied by the pretrained FLUX VAE.}
\label{tab:normalization}
\begin{tabularx}{\linewidth}{@{}lX@{}}
\toprule
Representation & Transformation \\
\midrule
SigLIP2 / DINO image & Per token LayerNorm, without affine parameters \\
FLUX VAE image & $z=(h-b_{\mathrm{VAE}})a_{\mathrm{VAE}}$ using predefined factors \\
Pixel image & RGB values scaled to $[-1,1]$, then patchified \\
Flan-T5 text & $z=h/0.2$, a fixed scalar for all channels \\
Qwen3-E / Qwen3-VL-E text & $z_c=(h_c-\mu_c)/s_c$ using fixed corpus statistics \\
SigLIP2 text & $z_c=(h_c-\mu_c)/s_c$ using fixed corpus statistics \\
\bottomrule
\end{tabularx}
\end{table}
 
\subsection{MM-RiT architecture}
The default encoders produce 256 image tokens with 1,152 channels and 128 text tokens with 512 channels.
The default 668M MM-RiT has two input layers for text only (which we empirically add) followed by 17 joint attention blocks, with four time tokens per modality.
\paragraph{MM-RiT block computation.}
\label{app:mmrit-block-details}

Image and text representations are first projected to a shared hidden width. Given hidden states $H_{\mathrm i}^{\ell}$ and $H_{\mathrm t}^{\ell}$, an MM-RiT block computes
\begin{equation}
\begin{aligned}
(Q_{\mathrm i},K_{\mathrm i},V_{\mathrm i})&=W_{\mathrm i}^{\mathrm{QKV}}R_{\mathrm i,1}(H_{\mathrm i}^{\ell}),\quad
(Q_{\mathrm t},K_{\mathrm t},V_{\mathrm t})=W_{\mathrm t}^{\mathrm{QKV}}R_{\mathrm t,1}(H_{\mathrm t}^{\ell}),\\
(A_{\mathrm i},A_{\mathrm t})&=\operatorname{Split}\!\left[\operatorname{Attn}
\bigl([Q_{\mathrm i};Q_{\mathrm t}],[K_{\mathrm i};K_{\mathrm t}],[V_{\mathrm i};V_{\mathrm t}]\bigr)\right],\\
U_{\mathrm i}&=H_{\mathrm i}^{\ell}+W_{\mathrm i}^{\mathrm O}A_{\mathrm i},\qquad
U_{\mathrm t}=H_{\mathrm t}^{\ell}+W_{\mathrm t}^{\mathrm O}A_{\mathrm t},\\
H_{\mathrm i}^{\ell+1}&=U_{\mathrm i}+F_{\mathrm i}\!\left(R_{\mathrm i,2}(U_{\mathrm i})\right),\qquad
H_{\mathrm t}^{\ell+1}=U_{\mathrm t}+F_{\mathrm t}\!\left(R_{\mathrm t,2}(U_{\mathrm t})\right),
\end{aligned}
\label{eq:mmrit-block}
\end{equation}
where $[\cdot;\cdot]$ joins tokens, $R$ is RMSNorm, and $F$ is a SwiGLU feedforward network. Queries and keys are normalized with RMSNorm, and positions are encoded with rotary embeddings, in 2D for image tokens and in 1D for text tokens. We use RMSNorm instead of the popular AdaLN (MM-RiT does not modulate scale or shift with time). Each token~$j$ of modality~$m$ receives its own time embedding, $h_j\leftarrow h_j+W_m\,\mathrm{MLP}_m(\mathrm{emb}(t_j))$, where $\mathrm{emb}$ is a 256 dimensional sinusoidal embedding, $\mathrm{MLP}_m$ maps it to the hidden width with SiLU, $W_m$ is a linear layer initialized to zero, and $t_j{=}1$ at conditioning positions. The time tokens are four learned vectors per modality, each plus the embedding of the mean time of its stream through the same MLP; all of them are appended to the end of the text stream and take part in joint attention. Separate output projections for each modality return predictions to the original feature dimensions.

\begin{table}[ht]
\caption{\textbf{Backbone scaling varies width and depth (trainable parameters).} Parameter counts exclude encoders and decoders. Each backbone also has two input layers for text only; d12 to d28 denote head counts.}
\label{tab:shapes}
\centering\smalltable
\begin{tabular}{@{}lrrrrrr@{}}
\toprule
Configuration & Parameters & Width & Heads & Head dim. & Joint blocks & Total blocks\\
\midrule
d12 & 178.6M & 768   & 12 & 64 & 10 & 12\\
d15 & 335.0M & 960   & 15 & 64 & 13 & 15\\
d19 & 668.3M & 1,216 & 19 & 64 & 17 & 19\\
d24 & 1.33B  & 1,536 & 24 & 64 & 22 & 24\\
d28 & 2.12B  & 1,792 & 28 & 64 & 26 & 28\\
\bottomrule
\end{tabular}
\end{table}

In the names d12 to d28, the number gives the number of attention heads, which also equals the total number of blocks.
Each head has 64 channels.
The 668M point is the same reference configuration used in the framework, task mixture, and representation studies. 

\subsection{Representation decoders}
\label{app:decoders}
\subsubsection{Image decoders}
Each image representation uses its corresponding reconstruction decoder.
Following the representation autoencoder construction of \citet{zheng2026diffusion}, our SigLIP2-SO400M and WebSSL-DINO encoders use their corresponding pretrained Scale-RAE decoders~\citep{scalingrae}. The VAE encoder uses its pretrained VAE decoder, and the pixel arm uses an unpatchification head.

\subsubsection{Text decoders}
\label{app:text-decoder}
Each text representation has its own decoder, trained on text alone and kept fixed during MM-RiT training, and all quantitative text results outside the decoder comparison of Table~\ref{tab:decoder-qa} are read out in one pass: captions and answers to questions by parallel decoders, and the moves of the planning tasks by linear task heads. Flan-T5 states have two parallel decoders. The parallel caption decoder is trained on the pretraining captions alone for 3,000 steps; it reads out the captioning results of the pretraining study and gives the Flan-T5 entry of the parallel decoder row in Table~\ref{tab:encoder-text-readout}. The final parallel decoder, described next, is trained on captions and question answering text and reads out the question answering results of the final models. The other text representations use their own parallel decoders (below). In the mazes and in VSP, the moves are read from the generated text states by a linear head trained for each task (end of this section), not by a parallel decoder. An autoregressive decoder over the same states, described below, serves as a comparison and reads out the qualitative joint samples (Figures~\ref{fig:qual-joint-main}, \ref{fig:qual-joint-1}, \ref{fig:qual-joint-2}, and~\ref{fig:qual-joint-trajectory}).

\paragraph{Final parallel decoder.}
The decoder maps the 128 generated text states to 128 tokens in one forward pass. It is a copy of the Flan-T5-Small encoder stack (8 layers) followed by a vocabulary head. Every position attends to every other position, so the decoder needs no answer length as input; the decoded text is the token sequence before the first predicted EOS.

We train the decoder alone, on text only, without MM-RiT: the inputs are the frozen Flan-T5-Small states of the captions and ``Question: \ldots{} Answer: \ldots'' rows in the pretraining data, and no images are used. The decoder reads unnormalized encoder states: generated states are multiplied by 0.2 (Appendix~\ref{app:normalization}) before decoding. To make the decoder robust to imperfectly generated states, the first half of each batch is perturbed per position as $\widetilde z_j=\lambda_j z_j+(1-\lambda_j)\epsilon_j$ with $\lambda_j=\operatorname{sigmoid}(g_j)$, $g_j\sim\mathcal N(0.8,0.8^2)$, and $\epsilon_j\sim\mathcal N(0,0.2^2\mathbf I)$. The loss is cross entropy at all 128 positions; content and EOS positions have weight 1, and padding positions are labeled EOS with weight 0.2. There are two training stages of 8,000 steps each, with 64 texts per step. Stage one starts from the pretrained Flan-T5-Small encoder stack, and stage two picks up where it ended. The optimizer is AdamW with a constant learning rate of $10^{-4}$ and no weight decay; we clip gradients at 1.0, run bfloat16 autocast over float32 weights, and keep the T5 dropout of 0.1. One stage takes about 1.3 hours on a single A100 GPU.

\paragraph{Decoders for other text representations.}
The decoders of Qwen3-E, Qwen3-VL-E, and SigLIP2 text (Section~\ref{sec:representation}) share one design. They read the normalized text states, the space in which the model generates, and for a state $x$ with $D$ channels they output $Wx+b+H(T(Px))$. The linear branch $Wx+b$ is copied from the linear readout head of the same representation (end of this section), so each decoder starts out identical to that head. $P$ projects the $D$ channels to 512, $T$ is a randomly initialized trunk of 8 bidirectional layers with the shape of the Flan-T5-Small encoder, and the vocabulary head $H$ starts at zero. Each decoder is trained on the pretraining captions of its representation for 3,000 steps with 64 texts per step; padding positions are labeled EOS with weight 0.2, and the first half of each batch is perturbed as for the Flan-T5 decoder, with noise of unit scale in the normalized space ($\epsilon_j\sim\mathcal N(0,\mathbf I)$, the same scale relative to the states as $\mathcal N(0,0.2^2\mathbf I)$ for raw T5 states). Table~\ref{tab:alt-decoders} lists their sizes; on held out captions, each decoder recovers more tokens than the linear head it starts from (Table~\ref{tab:encoder-text-readout}).

\begin{table}[!htbp]
\centering\small
\caption{\textbf{Parallel decoders of the other text representations.} $D$ is the state width, $V$ the vocabulary size, and length the number of text positions. Token accuracy is measured from clean states on the 5,000 held out captions of Table~\ref{tab:encoder-text-readout}.}
\label{tab:alt-decoders}
\begin{tabular*}{\linewidth}{@{\extracolsep{\fill}}lrrrrr@{}}
\toprule
Representation & $D$ & $V$ & Length & Parameters & Token accuracy\\
\midrule
Qwen3-E & 1,024 & 151,680 & 128 & 253M & 0.9797\\
Qwen3-VL-E & 2,048 & 151,936 & 128 & 409M & 0.9748\\
SigLIP2 text & 1,152 & 256,000 & 64 & 446M & 0.9870\\
\bottomrule
\end{tabular*}
\end{table}

\paragraph{Comparison with an autoregressive decoder.}
We compare the parallel decoder with an autoregressive decoder that reads the same states and with a linear readout head (described at the end of this section). The autoregressive decoder is the Flan-T5-Small decoder (8 layers, width 512, 6 heads, feedforward width 1,024, gated GELU) with the frozen Flan-T5-Small vocabulary head (58.1M parameters in total). It attends to the 128 unnormalized states through cross attention and decodes greedily with a KV cache. We train only its blocks and final layer norm (25.2M parameters) with teacher forcing and cross entropy on the content tokens and the final EOS, perturbing half of each batch with the same noise per position as above (AdamW, learning rate $10^{-4}$, no weight decay, 64 texts per step). For captions, we take a version of this decoder that was already finetuned on perturbed states and train it for 1,500 more steps on caption states; we call the result the autoregressive caption decoder. For question answering, we train it for 3,000 more steps on the VQAv2 training split. Under teacher forcing, the decoder input starts with the ``Question: \ldots{} Answer:'' prefix, and the loss covers only the answer and EOS. Answer positions are perturbed, and the positions after the answer are kept, set to zero, or replaced by noise, each with equal probability. On 300 VQAv2 validation questions, this raises its VQA accuracy on clean answer states from 61.2\% to 95.9\%. On captions held out from training, the parallel decoder recovers almost every token and 98.4\% of captions exactly from clean states. It keeps most of this accuracy (93.7\%) when the clean fraction of each state drops to 0.7, whereas the autoregressive decoder falls to 58.5\% (Table~\ref{tab:decoder-fidelity}). The linear head recovers 98.7\% of tokens but only 50.8\% of captions exactly, and its EOS error of 10.5 positions indicates that it often misplaces the end of the text. On answer states generated by \method{}, the two decoders are within 1.5 points on VQA benchmarks with open answers (VQAv2, OK-VQA, GQA), with the autoregressive decoder slightly ahead on most of them and on MME-P. On most multiple choice benchmarks the parallel decoder is far ahead, for example 66.0 vs.\ 45.3 on MMBench, while on MMMU the two are within 1.8 points (Table~\ref{tab:decoder-qa}). The training data of the parallel decoder include multiple choice answers given as option letters, whereas the question answering stage of the autoregressive decoder uses only VQAv2 answers, so the gap on multiple choice benchmarks reflects training data at least in part and cannot be attributed to the decoding scheme alone. The autoregressive decoder needs one forward pass per generated token, so its cost rises with the length of the output. For timing, we use one A100 80GB PCIe GPU (PyTorch 2.5.1, TF32 disabled), do two warmup runs, and synchronize the GPU before and after every timed call. On 2,000 captions of 84.6 tokens on average, with both decoders in bfloat16, the autoregressive decoder is 116, 163, 98, and 54 times slower than the parallel decoder at batch sizes 1, 16, 64, and 128. On VQAv2 answers of 1.6 tokens on average, again with both decoders in bfloat16, it is 3.9 times slower at batch size 1 (30.89 ms against 7.86 ms per answer) and 6.4 times slower at batch size 128, where it groups answers by prefix length (1.630 ms against 0.254 ms per answer). With both decoders in float32, the ratios are 4.1 and 2.1: at batch size 128, bfloat16 makes the parallel decoder 2.8 times faster but the autoregressive decoder slightly slower (1.501 ms per answer in float32).

\begin{table}[!htbp]
\centering\small
\caption{\textbf{Text decoder fidelity on 2,000 captions held out from training.} $\tau$ is the clean fraction of each input state ($\tau=1$: clean). String match compares the lowercased text up to the first EOS. EOS error is the mean distance, in positions, between the predicted and true end of the text. The row of the linear head differs slightly from Table~\ref{tab:encoder-text-readout}, which uses a different set of captions.}
\label{tab:decoder-fidelity}
\setlength{\tabcolsep}{4pt}
\begin{tabular*}{\linewidth}{@{\extracolsep{\fill}}lcccc@{}}
\toprule
Decoder & $\tau$ & Token accuracy & String match & EOS error\\
\midrule
Parallel (final) & 1.0 & 0.9997 & 0.984 & 0.00\\
Autoregressive & 1.0 & 0.9759 & 0.966 & 0.45\\
Parallel, first stage only & 1.0 & -- & 0.941 & --\\
Linear head & 1.0 & 0.9872 & 0.508 & 10.5\\
\midrule
Parallel (final) & 0.9 & 0.9997 & 0.983 & 0.00\\
Autoregressive & 0.9 & 0.9689 & 0.953 & 0.45\\
\midrule
Parallel (final) & 0.7 & 0.9989 & 0.937 & 0.00\\
Autoregressive & 0.7 & 0.7505 & 0.585 & 2.28\\
\bottomrule
\end{tabular*}
\end{table}

\begin{table}[!htbp]
\centering\small
\caption{\textbf{Parallel and autoregressive decoding of the same generated answer states} (16 steps, $w=3$).}
\label{tab:decoder-qa}
\setlength{\tabcolsep}{4pt}
\begin{tabular*}{\linewidth}{@{\extracolsep{\fill}}lrrrr@{}}
\toprule
& \multicolumn{2}{c}{SigLIP2} & \multicolumn{2}{c}{DINO}\\
\cmidrule(lr){2-3}\cmidrule(l){4-5}
Benchmark & Parallel & Autoregressive & Parallel & Autoregressive\\
\midrule
VQAv2 (val, 1,000) & 71.21 & 71.71 & 66.32 & 66.68\\
OK-VQA & 33.87 & 35.32 & 28.65 & 29.69\\
POPE & 83.97 & 83.97 & 82.29 & 82.29\\
GQA & 58.58 & 58.42 & 55.68 & 55.76\\
MME-P & 1115.7 & 1133.1 & 960.2 & 983.6\\
MMBench & 66.04 & 45.30 & 55.82 & 41.39\\
ScienceQA-IMG & 80.57 & 51.23 & 73.75 & 50.99\\
RealWorldQA & 53.63 & 42.86 & 40.52 & 35.83\\
AI2D & 34.40 & 29.51 & 31.87 & 27.40\\
MMMU & 29.58 & 27.81 & 31.51 & 31.51\\
\bottomrule
\end{tabular*}
\end{table}

\paragraph{Linear readout diagnostic.}
To measure how much token information each text representation retains (Table~\ref{tab:encoder-text-readout}), we also fit one linear vocabulary head per representation. These heads also initialize the linear branch of the parallel decoders for Qwen3-E, Qwen3-VL-E, and SigLIP2 text. Given a $D$ channel state $z_j$, the head predicts
\begin{equation}
\ell_j=Wz_j+b,\qquad
\widehat y_j=\arg\max_{v\in\{1,\ldots,V\}}\ell_{j,v},
\end{equation}
where $W\in\mathbb R^{V\times D}$ and $b\in\mathbb R^V$ use the encoder's token vocabulary. All positions are evaluated in parallel. The head adds no Transformer blocks and has $V(D+1)$ parameters (Table~\ref{tab:parallel-head-sizes}).

\begin{table}[!htbp]
\centering\small
\setlength{\tabcolsep}{5pt}
\caption{\textbf{Linear readout head sizes.}}
\label{tab:parallel-head-sizes}
\begin{tabularx}{\linewidth}{@{}Xrrr@{}}
\toprule
Text representation & Input $D$ & Vocabulary $V$ & Parameters (M) \\
\midrule
Flan-T5-Small & 512 & 32,128 & 16.48 \\
Qwen3-Embedding-0.6B & 1,024 & 151,680 & 155.47 \\
Qwen3-VL-Embedding-2B & 2,048 & 151,936 & 311.32 \\
SigLIP2 text & 1,152 & 256,000 & 295.17 \\
\bottomrule
\end{tabularx}
\end{table}

We train each linear head on our pretraining dataset using the corresponding text tokenizer and the frozen encoder for the corresponding representation. We fit the head separately from MM-RiT training using cross entropy loss. For each position in an augmented example (the first half of each batch),
\begin{equation}
\widetilde z_j=\lambda_j z_j+(1-\lambda_j)\epsilon_j,
\qquad \epsilon_j\sim\mathcal N(0,\mathbf I),
\qquad \lambda_j=\operatorname{sigmoid}(g_j),\quad
 g_j\sim\mathcal N(0.8,0.8^2).
\end{equation}
We add the noise in the normalized coordinates read by the linear head; the remaining examples stay clean. Padding is excluded from the cross entropy, while content and EOS positions are supervised. Decoding takes the argmax at every position at once and cuts the text at the first EOS. Only the vocabulary head is trained. We use AdamW (learning rate $10^{-3}$, batch size 64, 15{,}000 updates, weight decay $0.01$, $(\beta_1,\beta_2)=(0.9,0.999)$).

\paragraph{Linear task heads.}
The planning tasks read their moves with linear heads of the same form over Flan-T5 states, one for the mazes and one for VSP. Each is trained with the recipe above, on the texts of its task's training data (questions and answers, without images and without MM-RiT), for 1{,}500 updates instead of 15{,}000. Like the readout heads, a task head reads each position on its own.

\begingroup

\subsection{Frozen components and total parameters}
\label{app:total-parameters}
Table~\ref{tab:parameter-components} lists the fixed modules used by the principal and alternative representation configurations. Table~\ref{tab:parameter-stacks} then distinguishes three quantities for the principal SigLIP2/DINO+T5 configurations: trainable backbone parameters, modules executed for a particular direction, and the union of modules that must be loaded to support both directions.

\begingroup

\begin{table}[tbp]
\centering\smalltable
\caption{\textbf{Fixed component sizes.} Parameters were counted from the recorded module state dictionaries.  All text decoders are fixed during MM-RiT training.}
\label{tab:parameter-components}
\begin{tabularx}{\linewidth}{@{}XrX@{}}
\toprule
Component & Parameters & Used in\\
\midrule
\multicolumn{3}{@{}l}{\textit{Principal image/text pipeline components}}\\
SigLIP2-SO400M image encoder & 427.7M & I2T and VQA (SigLIP2 arm)\\
WebSSL-DINO300M image encoder & 303.7M & I2T and VQA (DINO arm)\\
Flan-T5-Small encoder & 35.3M & T2I, I2T, and VQA\\
SigLIP2 Scale-RAE image decoder & 415.6M & T2I (SigLIP2 arm)\\
DINO Scale-RAE image decoder & 415.4M & T2I (DINO arm)\\
FLUX.1-schnell VAE & 83.8M & VAE representation control\\
Parallel text decoder (stack of 8 layers, 18.88M + vocabulary head 16.48M) & 35.36M & I2T and VQA\\
Autoregressive text decoder (Flan-T5-Small decoder + vocabulary head) & 58.1M & Decoder comparison and qualitative joint samples\\
Flan-T5 linear vocabulary head & 16.48M & Readout diagnostic only\\
\midrule
\multicolumn{3}{@{}l}{\textit{Alternative text representations with their parallel decoders and linear readout heads}}\\
Qwen3-Embedding-0.6B text encoder & 595.8M & Qwen3-E representation arm\\
Qwen3-VL-Embedding-2B text tower & 1,720.6M & Qwen3-VL-E representation arm\\
SigLIP2 text tower & 707.8M & SigLIP2 text representation arm\\
Qwen3-E parallel decoder & 253M & Qwen3-E representation arm\\
Qwen3-VL-E parallel decoder & 409M & Qwen3-VL-E representation arm\\
SigLIP2 text parallel decoder & 446M & SigLIP2 text representation arm\\
Qwen3-E linear vocabulary head & 155.47M & Readout diagnostic; initializes the decoder\\
Qwen3-VL-E linear vocabulary head & 311.32M & Readout diagnostic; initializes the decoder\\
SigLIP2 text linear vocabulary head & 295.17M & Readout diagnostic; initializes the decoder\\
\bottomrule
\end{tabularx}
\arrayrulecolor{black}
\end{table}
\color{black}
\endgroup

\begingroup
\begin{table}[tbp]
\centering\scriptsize
\setlength{\tabcolsep}{3.4pt}
\renewcommand{\arraystretch}{1.05}
\caption{\textbf{Task active and fully loaded parameter counts.} T2I is backbone $+$ T5 encoder $+$ matched image decoder.  I2T/VQA is backbone $+$ image encoder $+$ T5 encoder $+$ parallel text decoder (35.36M); the d24 and d28 entries are computed from the rounded backbone sizes, so their last digit is approximate.  ``All loaded'' is the union.}
\label{tab:parameter-stacks}
\begin{tabular}{@{}llrrrr@{}}
\toprule
Image arm & Scale & Trainable & T2I forward & I2T/VQA forward & All loaded\\
\midrule
SigLIP2 & d12 & 178.6M & 629.5M & 677.0M & $\approx$1.09B\\
        & d15 & 335.0M & 785.9M & 833.4M & $\approx$1.25B\\
        & d19 & 668.3M & $\approx$1.12B & $\approx$1.17B & $\approx$1.58B\\
        & d24 & 1.33B  & $\approx$1.78B & $\approx$1.83B & $\approx$2.24B\\
        & d28 & 2.12B  & $\approx$2.57B & $\approx$2.62B & $\approx$3.03B\\
\midrule
DINO    & d12 & 178.6M & 629.3M & 553.0M & 968.4M\\
        & d15 & 335.0M & 785.7M & 709.4M & $\approx$1.12B\\
        & d19 & 668.3M & $\approx$1.12B & $\approx$1.04B & $\approx$1.46B\\
        & d24 & 1.33B  & $\approx$1.78B & $\approx$1.70B & $\approx$2.12B\\
        & d28 & 2.12B  & $\approx$2.57B & $\approx$2.49B & $\approx$2.91B\\
\bottomrule
\end{tabular}
\arrayrulecolor{black}
\end{table}
\color{black}
\endgroup

\arrayrulecolor{black}
\color{black}
\endgroup

\subsection{Training and inference details}
\label{app:setup}

\subsubsection{Training details}
\label{app:noise-loss}
Section~\ref{sec:method-training} defines joint, T2I, and I2T training. Here we give their noise distributions and the microbatch loss reduction; optimization settings are in Appendix~\ref{app:training-stages}.
 
The default pretraining mixture assigns $55.6\%$ of examples to joint denoising and $22.2\%$ each to T2I and I2T conditional denoising.
No training stage drops or replaces the condition. Each modality is a target with probability $0.556+0.222=0.778$ in Mixing and $0.5$ in Multi-task, which is only conditional. The modality weights are $\lambda_{\mathrm i}=1$ and $\lambda_{\mathrm t}=0.25$; their relation to the realized task weights is given in Equation~\ref{eq:task-weights}.
For Equation~\ref{eq:time-shift}, the base time is
\begin{equation}
u_m=\operatorname{sigmoid}(g_m),\qquad g_m\sim\mathcal N(\mu_m,s_m^2).
\end{equation}
Pretrained representation and VAE image streams use $(\mu_{\mathrm i},s_{\mathrm i})=(0,1)$, while the pixel control uses $(-0.8,0.8)$; these values follow the works cited in Table~\ref{tab:image-noise}.
Text streams use $(\mu_{\mathrm t},s_{\mathrm t})=(-1.5,0.8)$.
All continuous text configurations use $\sigma_{\mathrm t}=2$ and $\alpha_{\mathrm t}=1$.
The image settings are representation specific (Table~\ref{tab:image-noise}); the three text encoder substitutions retain the SigLIP2 image settings.
For representation and VAE image streams, $\alpha_{\mathrm i}=\sqrt{n_{\mathrm i}d_{\mathrm i}/4096}$, where $n_{\mathrm i}$ is the number of tokens and $d_{\mathrm i}$ is their feature dimension; the pixel control uses an unshifted schedule.
The FLUX VAE control encodes $256^2$ inputs to $32\times32$ latents with 16 channels, then packs $2\times2$ latent patches into 256 tokens of 64 channels; hence $\alpha_{\mathrm i}=\sqrt{256\cdot64/4096}=2$.
We also apply the corresponding time shift to the sampling grid~\citep{blm}.
 
\begin{table}[!htbp]
\centering\small
\caption{\textbf{Image noise settings in the representation configurations.} The base logit normal parameters, noise scales, and time shifts depend on the representation.}
\label{tab:image-noise}
\begin{tabular}{@{}lccc@{}}
\toprule
Image representation & $(\mu_{\mathrm i},s_{\mathrm i})$ & Noise scale $\sigma_{\mathrm i}$ & Time shift $\alpha_{\mathrm i}$ \\
\midrule
SigLIP2~\citep{zheng2026diffusion} & $(0,1)$ & 1 & 8.485 \\
DINO~\citep{zheng2026diffusion} & $(0,1)$ & 1 & 8 \\
FLUX VAE~\citep{sd3} & $(0,1)$ & 1 & 2 \\
Pixel~\citep{jit} & $(-0.8,0.8)$ & 2 & 1 \\
\bottomrule
\end{tabular}
\end{table}
 
The network predicts the clean state, and we train it with a stabilized velocity loss in which the predicted and the target velocity are divided by the same $d(t)=\max(1-t,0.05)$:
\begin{equation}
\widehat v_{\theta,m}=\frac{\widehat z_m-z_m^{t_m}}{d(t_m)},\qquad
v_m^{\mathrm{target}}=\frac{z_m-z_m^{t_m}}{d(t_m)},\qquad
d(t)=\max(1-t,0.05).
\label{eq:bounded-velocity}
\end{equation}
For the linear path $z_m^{t_m}=(1-t_m)\epsilon_m+t_m z_m$, this gives
\begin{equation}
\left\|\widehat v_{\theta,m}-v_m^{\mathrm{target}}\right\|_2^2
=\frac{\left\|\widehat z_m-z_m\right\|_2^2}{{d(t_m)}^2}.
\end{equation}
For $t_m\leq 0.95$ we have $d(t_m)=1-t_m$ and $v_m^{\mathrm{target}}=z_m-\epsilon_m$, which is exactly the usual velocity loss of flow matching on the linear path. Above 0.95, the floor in the denominator turns the loss into a stabilized surrogate near the clean endpoint instead of exact velocity supervision. Each stream averages the error over its valid corrupted positions and over feature dimensions; given positions and text padding are left out, and a stream with no targets contributes zero.
 
\paragraph{Task weights and active positions.}
For brevity, $J$, $T$, and $I$ denote the joint, T2I, and I2T tasks.
Let $a_{m,bp}$ indicate whether position $p$ of example $b$ is a generated, valid target of modality $m$, and let $e_{m,bp}$ be its squared velocity error averaged over feature channels.
For task $r\in\{J,T,I\}$, define the active count and empirical task loss in a microbatch:
\begin{equation}
C_{m,r}=\sum_{b:r_b=r}\sum_p a_{m,bp},\qquad
\bar\ell_m^r=\frac{\sum_{b:r_b=r}\sum_p a_{m,bp}e_{m,bp}}{\max(C_{m,r},1)},\qquad
C_m=\sum_r C_{m,r}.
\label{eq:task-reduction}
\end{equation}
An empty task target set has zero empirical loss. The denominator guard also makes a modality with no active targets contribute zero.
The corresponding microbatch averages are $\widehat{\mathcal L}_{\mathrm{joint}}^{\mathrm i}=\bar\ell_{\mathrm i}^J$, $\widehat{\mathcal L}_{\mathrm{joint}}^{\mathrm t}=\bar\ell_{\mathrm t}^J$, $\widehat{\mathcal L}_{\mathrm{T2I}}=\bar\ell_{\mathrm i}^T$, and $\widehat{\mathcal L}_{\mathrm{I2T}}=\bar\ell_{\mathrm t}^I$.
The realized count weights are
\begin{equation}
\begin{aligned}
w_{J,\mathrm i}&=\lambda_{\mathrm i}\frac{C_{\mathrm i,J}}{\max(C_{\mathrm i},1)},\qquad
w_{J,\mathrm t}=\lambda_{\mathrm t}\frac{C_{\mathrm t,J}}{\max(C_{\mathrm t},1)},\\
w_{\mathrm{T2I}}&=\lambda_{\mathrm i}\frac{C_{\mathrm i,T}}{\max(C_{\mathrm i},1)},\qquad
w_{\mathrm{I2T}}=\lambda_{\mathrm t}\frac{C_{\mathrm t,I}}{\max(C_{\mathrm t},1)}.
\end{aligned}
\label{eq:task-weights}
\end{equation}
With these weights, the weighted sum of the four task losses reduces exactly to
\begin{equation}
\widehat{\mathcal L}=\lambda_{\mathrm i}\frac{\sum_{b,p}a_{\mathrm i,bp}e_{\mathrm i,bp}}{\max(C_{\mathrm i},1)}
+\lambda_{\mathrm t}\frac{\sum_{b,p}a_{\mathrm t,bp}e_{\mathrm t,bp}}{\max(C_{\mathrm t},1)}.
\end{equation}
We minimize $\mathbb E_{\mathcal B}[\widehat{\mathcal L}(\mathcal B)]$, where the weights and the averages come from the same sampled microbatch $\mathcal B$.
Joint and T2I examples have image targets, and joint and I2T examples have text targets.
The task probabilities decide which task each example gets, but the loss weights follow the number of active positions actually drawn, counted without text padding or clean context.
Without times per token, the image branch first averages each active example over its positions and channels and then averages over examples, which is the same as averaging over tokens when every image has the same number of tokens.
 
\paragraph{Training times per token.}
We extend these task choices to partial inputs by varying time across positions, following the noising per token of Diffusion Forcing~\citep{diffusionforcing}. We first complete encoding and normalization (Appendix~\ref{app:normalization}), and then apply span masks and times per token only to the normalized target latents. For each target stream, with probability $15\%$, only a contiguous target span is corrupted and the remaining target positions stay clean. With probability $10\%$, target positions independently draw their times; otherwise they share the stream time. The same interpolation is applied at each selected latent position. These operations do not construct the clean VQA question condition, which uses the independent encoding of the question alone described below.
 
\paragraph{Parameter updates.}
For each microbatch, the frozen encoders produce the representations, each example gets a denoising task, and the targets are corrupted at the sampled times. MM-RiT predicts the clean representations, and the loss above updates its parameters, including the input and output projections. Encoders, image decoders, and text decoders do not change in this stage. We also keep two exponential moving averages (EMA) of the MM-RiT weights, with the decay rates given in Table~\ref{tab:optimizer}.
 
\subsubsection{Inference details}
\label{app:conditioning}
Sampling follows the T2I, I2T, and joint paths in Equation~\ref{eq:inference-paths}.
For a grid $0=s_0<\cdots<s_K=1$, a generated stream uses $t_{m,k}=S_{\alpha_m}(s_k)$ and the Euler update
\begin{equation}
z_{m,k+1}=z_{m,k}+(t_{m,k+1}-t_{m,k})v_{m,k}^{\mathrm{CFG}}.
\end{equation}
Both velocities in a joint step are computed from the same current pair before advancing either stream.

\paragraph{Guidance reference.}
The reference for a text condition uses the normalized states of a generic caption, and the reference for an image condition replaces the image with Gaussian noise at time $0$:
\begin{equation}
\emptyset_{\mathrm t}=N_{\mathrm t}(E_{\mathrm t}(\text{``a photo.''})),\qquad
v_{\mathrm i}^{\mathrm{ref}}=\widehat v_{\theta,\mathrm i}
(z_{\mathrm i}^{t_{\mathrm i}},\emptyset_{\mathrm t};t_{\mathrm i},1),\qquad
v_{\mathrm t}^{\mathrm{ref}}=\widehat v_{\theta,\mathrm t}
(\epsilon_{\mathrm i},z_{\mathrm t}^{t_{\mathrm t}};0,t_{\mathrm t}).
\label{eq:null-reference}
\end{equation}
Each clean prediction is converted to velocity before applying
\begin{equation}
v_m^{\mathrm{CFG}}=v_m^{\mathrm{ref}}+
w\bigl(\widehat v_{\theta,m}-v_m^{\mathrm{ref}}\bigr).
\label{eq:time-swap-cfg}
\end{equation}
For text-to-image generation we use 25 Euler steps and $w=5$, or 50 backbone evaluations per sample. In each text encoder arm, the generic caption goes through that arm's own encoder and normalization. The discrete text frameworks of Section~\ref{sec:paradigm} use no T5 encoder for it: they tokenize it with the Flan-T5 tokenizer and look up their own token embeddings. For the base D-DiT-style framework with token embeddings, Table~\ref{tab:paradigm} instead uses a fully masked caption as the reference. The controls conditioned on T5 swap only the clean condition for frozen T5 states. Captioning and question answering take 16 steps with $w=3$ and the image reference above; the question tokens are clean in both branches. Joint reasoning takes 16 backbone calls (8 in the comparison with a halved budget; Table~\ref{tab:maze-call-budgets} lists the steps per stream), and unconditional joint generation takes 64 steps. Neither uses guidance, so no reference branch is evaluated.

\paragraph{Guidance reference for text conditions.}
\label{app:guidance-ablation}
There are two ways to remove a text condition: use the clean generic caption ``a photo.'' (our default), or move the caption to its noise end, $t_{\mathrm t}=0$ (time swap). Table~\ref{tab:guidance-grid} gives GenEval after 50k updates for the four pretraining task mixtures that generate images. The generic caption scores higher for every mixture and every guidance scale from 2 to 5. With time swap, the Multi-task and Joint only models produce images that look like mosaics at most scales, which never happens with the generic caption. The default mixture depends least on the choice: at $w=5$ it reaches 48.41 GenEval with the generic caption and 44.64 with time swap, and 78.85 and 78.45 on DPG-Bench.

The mixtures also differ in their best scale. T2I only and Multi-task peak at $w=3$ and decline at larger scales, whereas the default mixture and Joint only improve up to $w=5$, the largest scale we evaluated. The pretraining studies use $w=5$ for every model. With the scale chosen per model in this range, T2I only reaches 51.23 and leads the default mixture by 2.8 points, while the default mixture still leads Multi-task (44.29) and Joint only (31.33).

\begin{table}[!htbp]
\centering\small
\caption{\textbf{GenEval of the pretraining task mixtures across guidance scales and text references} (50k updates, 25 steps). At $w=1$ the reference is not used. Bold marks each model's best score with the generic caption; $^\dagger$marks collapse into images that look like mosaics. I2T only cannot generate images (0.21 at every scale) and is omitted.}
\label{tab:guidance-grid}
\setlength{\tabcolsep}{4pt}
\begin{tabular*}{\linewidth}{@{\extracolsep{\fill}}llrrrrr@{}}
\toprule
Recipe & Text reference & $w=1$ & $w=2$ & $w=3$ & $w=4$ & $w=5$\\
\midrule
T2I only & generic caption & 36.12 & 47.26 & \textbf{51.23} & 51.03 & 47.86\\
 & time swap & 36.12 & 43.38 & 42.90 & 38.67 & 31.34\\
\midrule
Multi-task & generic caption & 32.36 & 43.44 & \textbf{44.29} & 40.85 & 36.30\\
 & time swap & 32.36 & 24.67 & 5.75$^\dagger$ & 2.16$^\dagger$ & 0.92$^\dagger$\\
\midrule
Joint only & generic caption & 10.09 & 21.54 & 27.43 & 31.06 & \textbf{31.33}\\
 & time swap & 10.09 & 3.30$^\dagger$ & 0.78$^\dagger$ & 0.31$^\dagger$ & 0.25$^\dagger$\\
\midrule
Mixing (default) & generic caption & 31.57 & 41.79 & 45.65 & 47.51 & \textbf{48.41}\\
 & time swap & 31.57 & 40.92 & 43.53 & 45.06 & 44.64\\
\bottomrule
\end{tabular*}
\end{table}

On WISE, the finetuned SigLIP2 model at $w=5$ scores 0.447 with the generic caption and 0.392 with time swap. Time swap at the weaker scale $w=3$ gives the highest WISE scores we measured (Table~\ref{tab:wise-timeswap}), but we keep one reference and one scale for all text-to-image benchmarks.

\begin{table}[!htbp]
\centering\small
\caption{\textbf{WISE with the time swap text reference at $w=3$.} 95\% confidence intervals in brackets. The main text uses the generic caption reference at $w=5$ (Table~\ref{tab:wise}).}
\label{tab:wise-timeswap}
\setlength{\tabcolsep}{3pt}
\begin{tabular*}{\linewidth}{@{\extracolsep{\fill}}lrrrrrrl@{}}
\toprule
Model & Cult. & Time & Space & Bio. & Phys. & Chem. & Overall\\
\midrule
JUICE (SigLIP2) & 0.466 & 0.533 & 0.642 & 0.396 & 0.531 & 0.346 & 0.488 [0.468, 0.508]\\
JUICE (DINO) & 0.379 & 0.481 & 0.607 & 0.344 & 0.456 & 0.313 & 0.424 [0.405, 0.443]\\
\bottomrule
\end{tabular*}
\end{table}

\paragraph{Given and generated positions.}
\label{app:partial}
For question answering and the puzzle tasks, the given context occupies only part of the image and text inputs. The conditional distribution is
\begin{equation}
(\widehat z_{\mathrm i}^{\mathrm{tar}},\widehat z_{\mathrm t}^{\mathrm{tar}})
\sim p_\theta\!\left(z_{\mathrm i}^{\mathrm{tar}},z_{\mathrm t}^{\mathrm{tar}}\mid
z_{\mathrm i}^{\mathrm{ctx}},z_{\mathrm t}^{\mathrm{ctx}}\right).
\label{eq:partial-conditioning}
\end{equation}
A modality with no target positions stays fixed. For VQA training, the frozen text encoder is run twice. The pass over the question alone, $N_{\mathrm t}(E_{\mathrm t}(Q))$, supplies the clean question prefix condition. A separate complete sequence pass $N_{\mathrm t}(E_{\mathrm t}(Q\Vert A))$ supplies target states only at answer positions. The question condition is never obtained by slicing question positions from the $Q\Vert A$ encoding; neither the ground truth answer content nor its length enters the conditioning representation. After encoding and normalization, we corrupt and optimize only the answer target latents. At inference, the image and the separately encoded question are held at time~$1$. The answer canvas, whose length is fixed in advance, starts from noise and follows the text sampling grid, and only its positions are updated.

\subsection{Data, optimization, and training stages}
\label{app:training-stages}
The pretraining studies in Section~\ref{sec:pretraining-study} train each configuration for 50k updates with the settings in Table~\ref{tab:optimizer}, except the budget controls of Table~\ref{tab:sim-supervision} (32.1k and 77.8k updates). The two 2.1B models in Section~\ref{sec:experiments} follow the stages in Table~\ref{tab:training-stages}: pretraining for 60k updates, supervised finetuning (SFT), and a separate adaptation to each planning task (mazes for both models, visual spatial planning for the SigLIP2 model).
\begin{table}[ht]
\caption{\textbf{Controlled comparisons share the same pretraining settings.} The budget controls of Table~\ref{tab:sim-supervision} differ only in the number of updates.}
\label{tab:optimizer}
\centering\smalltable
\begin{tabular}{@{}ll@{}}
\toprule
Setting & Value\\
\midrule
Updates / global batch & 50,000 / 4,096\\
Optimizer & AdamW, $(\beta_1,\beta_2)=(0.9,0.95)$, weight decay 0.01\\
Learning rate & $4\times10^{-4}$; warmup over 5,000 steps, then constant\\
Gradient norm cap / precision & 1.0 / bfloat16\\
EMA decay (slow / fast) & 0.99995 / 0.9996\\
Image / text loss coefficient & 1.0 / 0.25\\
Default image/text Gaussian noise std. & 1.0 / 2.0\\
Joint / T2I / I2T task ratio & 55.6\% / 22.2\% / 22.2\%\\
Probability of span / independent token times & 15\% / 10\%\\
Condition dropout & none\\
\bottomrule
\end{tabular}
\end{table}

\begin{table}[!htbp]
\centering\small
\caption{\textbf{Training stages of the final models.} Each later stage starts from the raw weights (without EMA) of the stage before it.}
\label{tab:training-stages}
\setlength{\tabcolsep}{4pt}
\begin{tabular*}{\linewidth}{@{\extracolsep{\fill}}llrrrl@{}}
\toprule
Stage & Starts from & Updates & Global batch & GPUs & Evaluated weights\\
\midrule
1. Pretraining & random initialization & 60,000 & 4,096 & 8 & fast EMA\\
2. SFT & stage 1, step 60,000 & 15,000 & 1,024 & 8 & fast EMA\\
3a. Maze adaptation & stage 2, step 15,000 & 5,000 & 1,024 & 8 & raw\\
3b. VSP adaptation & stage 2, step 15,000 & 5,000 & 1,024 & 4 & raw\\
\bottomrule
\end{tabular*}
\end{table}

\paragraph{Unified multimodal pretraining.} Following \citet{scalingrae} and \citet{blip3o}, we use 38M image caption pairs with recaptioned text, including images from CC12M~\citep{cc12m}, SA-1B~\citep{sam}, and JourneyDB~\citep{journeydb}. We also include Cambrian-7M for visual understanding supervision~\citep{cambrian}.

\paragraph{Supervised finetuning.}
SFT mixes text-to-image and understanding data (Table~\ref{tab:sft-data}). Text-to-image rows train text-to-image generation only; question answering, caption prompt, and long answer rows train image-to-text generation only, so SFT has no joint denoising. Each understanding row generates a fixed answer window whose length $K$ depends on the prompt type (12 for short answers and multiple choice, 48 for concise captions, 101 for detailed captions and long answers) and never on the answer length. Padding inside the window has loss weight 0.1, and positions after the window stay clean and are masked. Both image encoder variants use the same data. Table~\ref{tab:sft-recipe} gives the optimization settings.

\begin{table}[!htbp]
\centering
\caption{\textbf{SFT data.} Row counts include repetition. $K$ is the length of the generated answer window. Text-to-image rows train text-to-image generation; all other rows train image-to-text generation.}
\label{tab:sft-data}
\resizebox{\linewidth}{!}{%
\begin{tabular}{@{}>{\raggedright\arraybackslash}p{2.1cm}>{\raggedright\arraybackslash}p{6.6cm}>{\raggedright\arraybackslash}p{4.2cm}rrr@{}}
\toprule
Block & Sources (rows) & Task & $K$ & Rows & Share\\
\midrule
Text-to-image & BLIP3o-60k~\citep{blip3o} (3,776,625), DALL-E 3 synthetic pairs (989,248), ShareGPT-4o-Image~\citep{sharegpt4oimage} (2,377,284) & caption $\to$ image & -- & 7,143,157 & 34.7\%\\
Short QA & VQAv2 train~\citep{vqav2} (3,010,000); LLaVA-1.5 mix~\citep{llava15}, twice: GQA~\citep{gqa} (1,887,666), COCO~\citep{lin2014microsoft} (1,269,778), Visual Genome~\citep{krishna2017visual} (672,448), OCR-VQA~\citep{mishra2019ocr} (635,350), VizWiz~\citep{vizwiz} (40,256) & image + question $\to$ short answer & 12 & 7,515,498 & 36.5\%\\
Multiple choice & Cambrian~\citep{cambrian} multiple choice questions (433,570), ScienceQA train~\citep{lu2022scienceqa} (53,130) & image + question + options $\to$ letter & 12 & 486,700 & 2.4\%\\
Concise captions & COCO train2014 captions, twice (827,606), BLIP3-o short captions (1,250,000) & image + prompt (11 templates) $\to$ 1 to 40 tokens & 48 & 2,077,606 & 10.1\%\\
Detailed captions & BLIP3-o long captions (1,500,000), BLIP3o-60k and DALL-E 3 captions, four times (116,516) & image + prompt (16 templates) $\to$ 30 to 100 tokens & 101 & 1,616,516 & 7.9\%\\
Long answers & ALLaVA~\citep{allava} via Cambrian (671,552), ShareGPT4V~\citep{sharegpt4v} (1,042,950), LLaVA (3,719) & image + question $\to$ 9 to 100 tokens & 101 & 1,718,221 & 8.3\%\\
COCO val questions & Cambrian questions on COCO val2014/val2017 images (32,010) & image + question $\to$ short answer & 12 & 32,010 & 0.2\%\\
\midrule
Total & & & & 20,589,708 & 100\%\\
\bottomrule
\end{tabular}}
\end{table}

\begin{table}[!htbp]
\centering\small\setlength{\tabcolsep}{4pt}
\caption{\textbf{SFT recipe.}}
\label{tab:sft-recipe}
\begin{tabularx}{\linewidth}{@{}lX@{}}
\toprule
Initialization & Raw weights of the 2.1B pretraining run at step 60,000\\
Updates / global batch & 15,000 / 1,024 (8 rows $\times$ 16 accumulation steps $\times$ 8 GPUs); 15.36M samples, about 0.75 epoch\\
Optimizer / learning rate & AdamW, $(\beta_1,\beta_2)=(0.9,0.95)$; $10^{-4}$ after 250 linear warmup updates, then constant\\
Weight decay / gradient cap & 0.01 / 1.0\\
Precision / seed & bfloat16 / 4242\\
EMA decay (slow / fast) & 0.99995 / 0.9996; we evaluate the fast EMA\\
Task assignment & text-to-image rows: text-to-image only; question, caption prompt, and long answer rows: image-to-text only\\
Text loss weight & 0.25\\
Answer window & $K\in\{12,48,101\}$ set by prompt type, never by answer length; padding inside the window has weight 0.1; positions after the window stay clean and masked\\
Condition dropout & none\\
Image time shift & 8.485 (SigLIP2), 8.0 (DINO)\\
\bottomrule
\end{tabularx}
\end{table}

\FloatBarrier
\section{Pretraining study details}
\label{app:design-studies}\label{app:supplemental-studies}
\subsection{Generative frameworks}
\label{app:paradigm}
Tables~\ref{tab:paradigm} and~\ref{tab:framework-protocols} give the results and configurations for the controlled comparison in Section~\ref{sec:paradigm}. AR denotes autoregressive generation, FM denotes flow matching, and MD denotes masked diffusion.
\begin{table}[!htbp]
\centering\scriptsize\setlength{\tabcolsep}{3pt}
\caption{\textbf{Configurations in the framework comparison.} All use the 668M d19 backbone family and a schedule of 50k updates. FM denotes flow matching; AR denotes autoregressive; MD denotes masked diffusion.}
\label{tab:framework-protocols}
\begin{tabularx}{\linewidth}{@{}l*{3}{>{\raggedright\arraybackslash}X}@{}}
\toprule
Configuration & Image space / process & Text space / process & Image / text decoder \\
\midrule
JUICE & SigLIP2 features / FM & T5 features / FM & RAE / parallel decoder of T5 states \\
Transfusion-style & SigLIP2 features / FM & Discrete tokens / AR & RAE / token vocabulary head  \\
D-DiT-style & SigLIP2 features / FM & Discrete tokens / MD & RAE / token vocabulary head  \\
MMaDA-style & aMUSEd VQ codes~\citep{amused} / MD & Discrete tokens / MD & VQ decoder / token vocabulary head\\
\bottomrule
\end{tabularx}
\end{table}

\begingroup
\subsubsection{Framework implementations}
\label{app:architecture-implementations}

Figure~\ref{fig:four-architectures} summarizes the four generative configurations compared in Table~\ref{tab:paradigm}.
All use an MM-RiT trunk of width 1,216 with two streams, 17 joint blocks, two input layers for text only, 19 attention heads of dimension 64, and four time tokens per modality, trained under the same schedule of 50k updates.
The configurations differ in their state representations, generation objectives, attention masks, and output interfaces.

\paragraph{Text representations and conditioning.}
In the base framework comparison, only JUICE uses the frozen Flan-T5-Small encoder for both clean conditioning states and continuous generation targets. Appendix~\ref{app:matched-t5} additionally supplies frozen T5 states to the clean text condition branch of the discrete text configurations, while retaining their original generated targets and objectives.
The other three configurations use discrete text tokens with learned embeddings and token prediction heads.
The base token configurations reuse the T5 tokenizer and its vocabulary of 32,128 entries, but do not run the T5 encoder for either clean conditioning text or generated text targets.
Their clean text conditions enter the backbone directly through token embeddings.
Thus, sharing a tokenizer does not imply sharing pretrained contextual text representations.

\begin{figure}[!htbp]
\centering
\includegraphics[width=\linewidth]{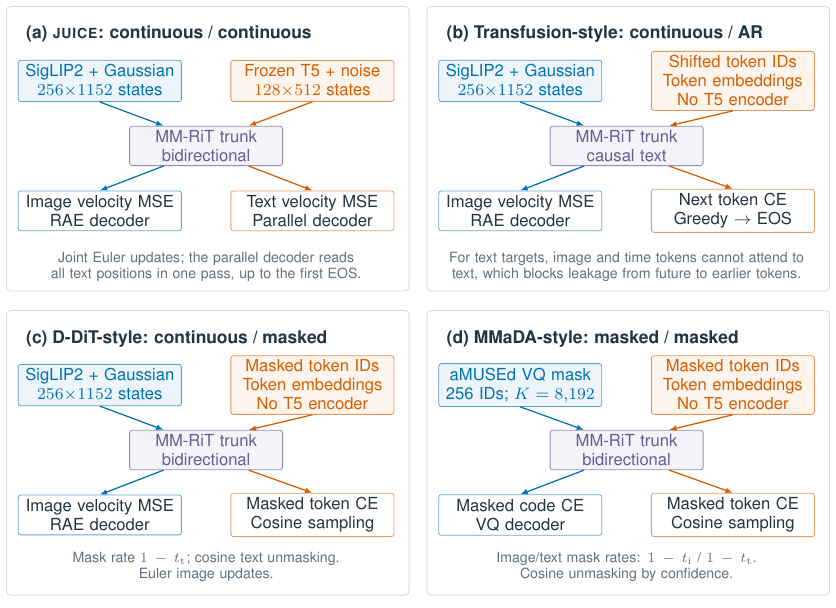}
\caption{\textbf{Four generative configurations at a shared MM-RiT trunk scale.}
JUICE uses frozen SigLIP2 and T5 features for continuous image and text generation.
The token controls use learned text embeddings without a T5 encoder; only the tokenizer and vocabulary are shared.
The first three configurations use continuous SigLIP2 image states, whereas the fully discrete control uses VQ image codes.
MSE and CE denote mean squared error and cross entropy, respectively.}
\label{fig:four-architectures}
\end{figure}

\paragraph{Continuous image and text (JUICE).}
At $224{\times}224$, the image stream holds 256 frozen SigLIP2 states with 1,152 channels and the text stream 128 frozen T5 states with 512 channels.
Both streams are generated with continuous flow matching, each with its own sampled time.
The bidirectional trunk predicts clean states; we turn them into velocities for the stabilized velocity loss on generated image tokens and valid text positions.
During Euler sampling, both streams are advanced from the same current pair of image and text.
The matched RAE decoder gives the image, and the parallel text decoder (Appendix~\ref{app:text-decoder}) reads all final text states in one pass, cutting the text at the first predicted EOS.

\paragraph{Continuous image and autoregressive text (Transfusion-style).}
This control keeps the continuous SigLIP2 image stream and predicts text tokens autoregressively instead of generating text states with flow matching.
At text target positions, the inputs are the token embeddings shifted by one, with the mask token as the start symbol, and the input embedding matrix is shared with the output head.
Valid text positions are trained with cross entropy on the next token.
In examples with text targets, text queries attend causally, and image and time queries cannot attend to text target keys, so future tokens cannot leak to earlier text positions through the other streams.
A clean text condition for image generation stays fully visible through its token embeddings.
Images are generated by Euler integration and decoded with the RAE decoder, and text is decoded greedily up to EOS.

\paragraph{Continuous image and masked text (D-DiT-style).}
Here the image stream stays continuous (SigLIP2), but text becomes a canvas of discrete tokens~\citep{dualdiffusion}.
Each text target token is masked with probability $1-t_{\mathrm t}$; the text time therefore controls how many tokens are visible.
Visible tokens enter the bidirectional trunk through their embeddings, and masked ones through a learned mask embedding.
We train masked valid text positions with cross entropy and keep the continuous velocity loss for the image.
For sampling, the text canvas starts fully masked, and a cosine schedule reveals the most confident positions step by step. The image is sampled with Euler updates.

\paragraph{Masked image and text (MMaDA-style).}
Here both modalities are discrete~\citep{mmada}. Images become a $16\times16$ canvas of aMUSEd VQ codes (codebook of 8,192 entries, $256\times256$ input).
Image and text targets are masked independently according to their modality times and trained with cross entropy on the masked tokens.
The embeddings of both canvases go into the bidirectional trunk with two streams, and at sampling time each canvas is unmasked in order of confidence with a cosine schedule.
The frozen VQ decoder turns the image codes into pixels, and the text tokens are read up to EOS.

\endgroup

\subsubsection{Matching the text-to-image conditioning encoder}
\label{app:matched-t5}
In their base form, the Transfusion-style, D-DiT-style, and MMaDA-style frameworks condition image generation on token embeddings, while JUICE conditions on contextual frozen T5 states. To see how much this matters, we give each token configuration the same frozen Flan-T5-Small states in its clean T2I condition branch. Nothing else changes: the generated text stays discrete, with the original autoregressive or masked token objective and readout, and the image representation and image objective are the same as before. We use the 668M backbone family, 50k pretraining updates, and the common guidance setting ($w=5$ with the generic caption reference).

Table~\ref{tab:paradigm} in the main text includes these rows.

Replacing token embeddings with T5 raises Transfusion-style GenEval from 44.23 to 46.07 (a gain of 1.84 points) and DPG from 76.84 to 78.24 (a gain of 1.40 points). This GenEval gain of 1.84 points accounts for part of the original gap of 4.18 points to JUICE; the remaining gap in the point estimate is 2.34 points. D-DiT-style gains 3.08 points in GenEval and 1.96 points in DPG, while MMaDA-style gains 1.74 points in GenEval and 1.96 points in DPG. Their remaining gaps may reflect other differences in the generation configuration, which this comparison does not isolate: D-DiT-style retains a masked token objective for text, and MMaDA-style also uses a VQ image interface with masked image generation.

\FloatBarrier

\subsection{Task mixtures}
\label{app:mixture}
Table~\ref{tab:mixture} gives the five training recipes in Section~\ref{sec:mixture}.

\subsubsection{Supervision and cross-modal information}
\label{app:sim-supervision}
This study separates how many targets the model gets from what information the joint branch makes available.

\begin{table}[!htbp]
\centering\small
\setlength{\tabcolsep}{3pt}
\caption{\textbf{Supervision and budget controls, 668M.} Work proxy: backbone forward passes summed over training, relative to Mixing at 50k updates; DC spends two passes on each joint example.}
\label{tab:sim-supervision}
\begin{tabular}{@{}lrrrrrrr@{}}
\toprule
Recipe & \shortstack{Updates\\(k)} & \shortstack{Targets\\(M/mod.)} & \shortstack{Work\\proxy} & GE & DPG & CIDEr & CLIP \\
\midrule
Mixing & 50 & 159.33 & 1.000 & 48.41 & 78.85 & 88.80 & 80.20 \\
DC (matched targets) & 50 & 159.33 & 1.556 & 45.90 & 78.05 & 87.72 & 79.63 \\
DC (matched work) & 32.1 & 102.40 & 1.000 & 43.18 & 75.24 & 71.55 & 76.81 \\
Blocked joint & 50 & 159.33 & 1.000 & 45.49 & 78.14 & 86.58 & 78.66 \\
Multi-task / 50k & 50 & 102.40 & 1.000 & 36.30 & 77.71 & 79.41 & 75.07 \\
Multi-task / 77.8k & 77.8 & 159.33 & 1.556 & 43.16 & 78.51 & 89.31 & 79.14 \\
Mixing / 32.1k & 32.1 & 102.40 & 0.643 & 44.90 & 75.56 & 72.30 & 77.37 \\
Single target joint & 50 & 102.40 & 1.000 & 40.08 & 77.98 & 82.94 & 77.33 \\
\bottomrule
\end{tabular}
\par\vspace{2pt}\raggedright\footnotesize{GenEval (GE), DPG-Bench (DPG), and CLIPScore (CLIP) use a scale from 0 to 100.}
\end{table}

\paragraph{Settings.}
Every model uses the default 668M d19 pretraining protocol with frozen SigLIP2 and Flan-T5-Small encoders and a global batch of 4,096; Table~\ref{tab:sim-supervision} lists update counts and work proxies. Mixing keeps the default $55.6/22.2/22.2$ task mixture. The controls change either the computation of the joint branch or the assignment of targets, as described below and in Table~\ref{tab:sim-supervision}. All arms are scored on the same evaluation inputs.

\paragraph{Dual conditional (DC) control.}
Each example assigned to the original joint branch gets two conditional forward passes instead of one joint pass: one predicts the image from clean text, and the other predicts the text from the clean image. Both reuse the sampled target masks, corruption times, and noises. We sum the two target loss numerators and divide by the same active target counts of each modality as in Mixing, followed by one optimizer update; averaging the two branches separately and adding them would double the loss scale. The original conditional branches are kept. 

\paragraph{Blocked joint control.}
This control keeps everything about the joint branch (both targets, their noise, the task probabilities, and the loss reduction) except that image and text cannot see each other on joint examples. We mask every path between the two modalities at every layer, including paths through shared time or auxiliary tokens. Shared tokens are duplicated per modality, and each copy carries only its own modality's information. The conditional branches stay as they are.

\paragraph{Single target joint control.}
Joint examples keep both streams noisy, but only one modality, picked with probability $1/2$, is supervised, and the other stream serves only as context. With the conditional branches unchanged, each modality is then a target with probability $0.222+0.556/2=0.5$, the same as in Multi-task at the same number of updates.

\paragraph{Interpretation.}
Mixing ends 2.51 GenEval and 1.08 CIDEr ahead of matched DC after 50k updates, and 2.92 GenEval and 2.22 CIDEr ahead of blocked joint. DC alters the clean or noisy context and the computation at the same time, so this comparison alone cannot tell how much semantic information the noisy partner provides. Still, the results fit a contribution from cross-modal exchange beyond the target count. Compared with Multi-task at 50k updates, DC with matched work (32.1k updates, the same work) gains 6.9 GenEval but loses 7.9 CIDEr and 2.5 DPG. Because this is a mixed result and not a uniform gain, we draw our conclusions about joint training from the matched DC and blocked joint controls.

\FloatBarrier

\subsection{Controls on representation spaces}
\label{app:representation}
Table~\ref{tab:representation} provides the full metrics underlying Figure~\ref{fig:encoder}.
\begin{table}[!htbp]
\caption{\textbf{Configurations of representation spaces exhibit different strengths across generation and captioning.}}
\label{tab:representation}
\centering\smalltable
\begin{tabular}{@{}llrrrr@{}}
\toprule
Image representation & Text representation & GE & DPG & CIDEr & CLIP\\
\midrule

SigLIP2 & T5 & 48.41 & \textbf{78.85} & \textbf{88.80} & \textbf{80.20}\\
DINO & T5 & \textbf{49.07} & 77.85 & 68.80 & 71.66\\
FLUX VAE & T5 & 35.42 & 74.05 & 49.95 & 64.91\\
Pixels & T5 & 24.48 & 71.64 & 49.57 & 65.51\\
SigLIP2 & Qwen3-E & 43.07 & 78.04 & 88.50 & 76.54\\
SigLIP2 & Qwen3-VL-E & 39.28 & 78.70 & 84.24 & 75.94\\
SigLIP2 & SigLIP2 & 33.08 & 74.19 & 85.54 & 74.52\\
\bottomrule
\end{tabular}
\end{table}

\begingroup
\subsubsection{Diagnostics of encoders and decoders}
\label{app:encoder-decoder-diagnostics}

\paragraph{Text readout across representations.}
To separate representation recovery from generation of continuous states, we encode reference captions held out from training and decode the resulting clean states without passing them through MM-RiT.  Table~\ref{tab:encoder-text-readout} reports the token accuracy of linear readout heads fitted to each representation, which measure how much token information each space retains, and of the parallel decoders of the pretraining study (Appendix~\ref{app:text-decoder}), which decode its captioning results.

\begingroup
\begin{table}[!htbp]
\centering\small
\caption{\textbf{Token accuracy of text readouts from clean states.} 5,000 reference captions held out from training are encoded and immediately decoded, without MM-RiT generation. The linear heads are fitted to each representation; the parallel decoders are those of the pretraining study (Appendix~\ref{app:text-decoder}), whose linear branch starts from the linear head for Qwen3-E, Qwen3-VL-E, and SigLIP2 text. SigLIP2 text uses 64 positions and the other representations 128.}
\label{tab:encoder-text-readout}
\begin{tabular*}{\linewidth}{@{\extracolsep{\fill}}lcccc@{}}
\toprule
Readout & Flan-T5 & Qwen3-E & Qwen3-VL-E & SigLIP2 text\\
\midrule
Linear head & .9849 & .9686 & .9652 & .9813\\
Parallel decoder & .9922 & .9797 & .9748 & .9870\\
\bottomrule
\end{tabular*}
\end{table}
\endgroup

\paragraph{Pairs of image encoders and decoders.}
Table~\ref{tab:encoder-image-codec} complements the downstream representation comparison with reconstruction fidelity. Lower rFID is better.  The FLUX VAE reconstructs most faithfully on all three reconstruction sets among the learned pairs of encoder and decoder, while DINO and SigLIP2 give substantially stronger downstream GenEval in the fixed 668M generation comparison. These results show that reconstruction fidelity alone does not determine the downstream generation ranking. 

\begingroup
\begin{table}[!htbp]
\centering\small
\setlength{\tabcolsep}{3.0pt}
\renewcommand{\arraystretch}{1.05}
\caption{\textbf{Reconstruction diagnostics of image encoders and decoders.} Lower rFID is better. ImageNet, YFCC, and RenderedText each contain 50k reconstruction samples, following the Scale-RAE comparison~\citep[Table~2]{scalingrae}.}
\label{tab:encoder-image-codec}
\begin{tabular}{@{}lrrr@{}}
\toprule
Image representation / decoder & rFID ImageNet & rFID YFCC & rFID RenderedText\\
\midrule
FLUX VAE & .288 & .410 & .638\\
WebSSL-DINO / matched Scale-RAE decoder & .388 & .558 & 1.372\\
SigLIP2 / matched Scale-RAE decoder & .435 & .702 & 1.621\\
SDXL VAE (reconstruction reference) & .930 & 1.168 & 2.057\\
Pixels (identity reference) & .000 & .000 & .000\\
\bottomrule
\end{tabular}
\arrayrulecolor{black}
\end{table}
\color{black}
\endgroup

\arrayrulecolor{black}
\color{black}
\endgroup

\FloatBarrier

\FloatBarrier
\section{Extended benchmark comparisons}
\label{app:main-comparisons}\label{app:extended-eval}
Tables~\ref{tab:generation-breakdown} and~\ref{tab:understanding-extended} extend the main comparisons to larger models, and Table~\ref{tab:geneval-steps} gives GenEval at 25, 50, and 100 sampling steps.
\begin{table}[!htbp]
\caption{\textbf{Image generation with larger models.}}
\label{tab:generation-breakdown}
\centering\footnotesize\setlength{\tabcolsep}{3pt}\renewcommand{\arraystretch}{1.0}
\begin{tabularx}{\linewidth}{@{}lRRRRRRRR@{}}
\toprule
\multicolumn{9}{l}{\textit{(a) GenEval}}\\
Model & Size & Single & Two & Count & Color & Position & Attr. & Overall\\
\midrule
BLIP3-o~\citep{blip3o} & 8B & -- & -- & -- & -- & -- & -- & 0.84\\
Janus-Pro-7B~\citep{januspro} & 7B & 0.99 & 0.89 & 0.59 & 0.90 & 0.79 & 0.66 & 0.80\\
Show-o2~\citep{showo2} & 7B & 1.00 & 0.87 & 0.58 & 0.92 & 0.52 & 0.62 & 0.76\\
TUNA~\citep{tuna} & 7B & 1.00 & 0.97 & 0.81 & 0.91 & 0.88 & 0.83 & 0.90\\
TUNA-R~\citep{tuna2} & 7B & 1.00 & 0.95 & 0.82 & 0.89 & 0.86 & 0.79 & 0.88\\
Tuna-2~\citep{tuna2} & 7B & 0.99 & 0.96 & 0.80 & 0.91 & 0.84 & 0.76 & 0.87\\
BAGEL~\citep{bagel} & 7B/14B & 0.98 & 0.95 & 0.84 & 0.95 & 0.78 & 0.77 & 0.88\\
MMaDA~\citep{mmada} & 8B & 0.99 & 0.76 & 0.61 & 0.84 & 0.20 & 0.37 & 0.63\\
\midrule
\rowcolor{paperblue!6}JUICE (SigLIP2) & 2.1B & 0.99 & 0.93 & 0.78 & 0.91 & 0.85 & 0.73 & 0.87\\
\rowcolor{paperblue!6}JUICE (DINO) & 2.1B & 0.99 & 0.95 & 0.81 & 0.93 & 0.85 & 0.78 & 0.88\\
\bottomrule
\end{tabularx}
\par\smallskip
\begin{tabularx}{\linewidth}{@{}lRRRRRRR@{}}
\toprule
\multicolumn{8}{l}{\textit{(b) DPG-Bench}}\\
Model & Size & Global & Entity & Attribute & Relation & Other & Overall\\
\midrule
BLIP3-o~\citep{blip3o} & 8B & -- & -- & -- & -- & -- & 81.60\\
Janus-Pro-7B~\citep{januspro} & 7B & 86.90 & 88.90 & 89.40 & 89.32 & 89.48 & 84.19\\
Show-o2~\citep{showo2} & 7B & 89.00 & 91.78 & 89.96 & 91.81 & 91.64 & 86.14\\
TUNA~\citep{tuna} & 7B & 90.42 & 91.68 & 90.94 & 91.87 & 90.73 & 86.76\\
TUNA-R~\citep{tuna2} & 7B & 86.00 & 91.80 & 91.03 & 93.48 & 84.89 & 86.35\\
Tuna-2~\citep{tuna2} & 7B & 89.50 & 91.40 & 92.07 & 91.91 & 88.81 & 86.54\\
\midrule
\rowcolor{paperblue!6}JUICE (SigLIP2) & 2.1B & 82.67 & 91.41 & 90.44 & 94.58 & 86.80 & 85.22\\
\rowcolor{paperblue!6}JUICE (DINO) & 2.1B & 83.28 & 90.24 & 88.78 & 95.05 & 81.60 & 84.38\\
\bottomrule
\end{tabularx}
\end{table}

\begin{table}[!htbp]
\centering\small
\caption{\textbf{GenEval of JUICE at different numbers of sampling steps} ($w=5$, $\times100$). The main text uses 25 steps.}
\label{tab:geneval-steps}
\setlength{\tabcolsep}{4pt}
\begin{tabular*}{\linewidth}{@{\extracolsep{\fill}}lrrrrrrrr@{}}
\toprule
Model & Steps & Single & Two & Count & Color & Position & Attr. & Overall\\
\midrule
JUICE (SigLIP2) & 25 & 99.4 & 93.4 & 77.8 & 91.0 & 85.0 & 73.0 & 86.60\\
 & 50 & 98.8 & 92.7 & 77.8 & 92.3 & 84.5 & 74.8 & 86.80\\
 & 100 & 98.8 & 93.2 & 78.1 & 91.8 & 84.3 & 74.5 & 86.76\\
\midrule
JUICE (DINO) & 25 & 99.4 & 94.7 & 80.6 & 93.1 & 84.8 & 77.5 & 88.34\\
 & 50 & 99.7 & 94.2 & 81.3 & 92.6 & 84.8 & 78.8 & 88.53\\
 & 100 & 99.7 & 94.2 & 81.9 & 92.8 & 84.3 & 79.0 & 88.64\\
\bottomrule
\end{tabular*}
\end{table}

\begin{table}[!htbp]
\caption{\textbf{Visual understanding with larger models.}}
\label{tab:understanding-extended}
\centering\footnotesize\setlength{\tabcolsep}{2.5pt}\renewcommand{\arraystretch}{1.0}
\begin{tabularx}{\linewidth}{@{}lRRRRRR@{}}
\toprule
Model & Size & Input & VQAv2$\uparrow$ & GQA$\uparrow$ & MME-P$\uparrow$ & POPE$\uparrow$\\
\midrule
\rowcolor{black!6}\multicolumn{7}{c}{\textit{Models for understanding only}}\\
LLaVA-1.5~\citep{llava15} & 7B & $336^2$ & 78.50 & 62.00 & 1510.7 & 85.90\\
Qwen-VL-Chat~\citep{qwenvl} & 7B & $448^2$ & 78.20 & 57.50 & -- & --\\
\rowcolor{black!6}\multicolumn{7}{c}{\textit{Composite unified models}}\\
BLIP3-o~\citep{blip3o} & 8B & Dyn. & 83.10 & -- & 1682.6 & --\\
\rowcolor{black!6}\multicolumn{7}{c}{\textit{Native unified models}}\\
Janus-Pro-7B~\citep{januspro} & 7B & $384^2$ & -- & 62.00 & 1567.1 & 87.40\\
Show-o2~\citep{showo2} & 7B & $432^2$ & -- & 63.10 & 1620.5 & --\\
TUNA~\citep{tuna} & 7B & $512^2$ & -- & 63.90 & 1641.5 & --\\
TUNA-R~\citep{tuna2} & 7B & $512^2$  & -- & 63.50 & -- & --\\
Tuna-2~\citep{tuna2} & 7B & $512^2$  & -- & 65.00 & -- & --\\
BAGEL~\citep{bagel} & 7B/14B & Dyn. & -- & -- & 1687.0 & --\\
OneFlow Mixed~\citep{oneflow} & 8B & $512^2$ & 73.70 & 61.90 & 1542.5 & 86.30\\
MMaDA~\citep{mmada} & 8B & $512^2$ & 76.70 & 61.30 & 1410.7 & 86.10\\
\midrule
\rowcolor{paperblue!6}\textbf{JUICE (SigLIP2)} & 2.1B & $224^2$ & 73.10 & 58.58 & 1115.7 & 83.97\\
\rowcolor{paperblue!6}\textbf{JUICE (DINO)} & 2.1B & $224^2$ & 68.75 & 55.68 & 960.2 & 82.29\\
\bottomrule
\end{tabularx}
\end{table}

Table~\ref{tab:understanding-full} lists all 10 understanding benchmarks for both JUICE variants, with VQAv2 on two splits, and Table~\ref{tab:understanding-other} compares the benchmarks that are not in the main table with models that report them.
\begin{table}[!htbp]
\caption{\textbf{JUICE on 10 understanding benchmarks.} VQAv2 is reported on two splits. Question answering with 16 sampling steps, guidance $w{=}3$, and the parallel text decoder. OK-VQA is scored against the annotator answers as written (the raw\_answer field of OK-VQA v1.1); the lemmatized answers of the same file give 29.15 and 24.63, since predictions are not lemmatized.}
\label{tab:understanding-full}
\centering\footnotesize\setlength{\tabcolsep}{4pt}
\begin{tabular*}{\linewidth}{@{\extracolsep{\fill}}lllrr@{}}
\toprule
Benchmark & Split & Metric & SigLIP2 & DINO\\
\midrule
VQAv2~\citep{vqav2} & test-dev & VQA accuracy & 73.10 & 68.75\\
\quad yes/no & & & 90.66 & 88.18\\
\quad number & & & 58.73 & 53.30\\
\quad other & & & 61.29 & 55.57\\
VQAv2 & val2014, fixed 1,000 & VQA accuracy & 71.21 & 66.32\\
OK-VQA~\citep{okvqa} & val2014 & VQA accuracy & 33.87 & 28.65\\
POPE~\citep{pope} & random, popular, adversarial & accuracy & 83.97 & 82.29\\
MME-P~\citep{mme} & test & perception score & 1115.7 & 960.2\\
GQA~\citep{gqa} & testdev-balanced & accuracy & 58.58 & 55.68\\
MMBench~\citep{mmbench} & dev (English) & accuracy & 66.04 & 55.82\\
AI2D~\citep{ai2d} & test & accuracy & 34.40 & 31.87\\
MMMU~\citep{mmmu} & validation & accuracy & 29.58 & 31.51\\
ScienceQA-IMG~\citep{lu2022scienceqa} & test, image questions & accuracy & 80.57 & 73.75\\
RealWorldQA~\citep{realworldqa} & test, multiple choice & accuracy & 53.63 & 40.52\\
\bottomrule
\end{tabular*}
\end{table}

\begin{table}[!htbp]
\centering
\caption{\textbf{Remaining understanding benchmarks.} Only values reported in each model's paper are listed. JUICE uses OK-VQA val and ScienceQA-IMG test, and its SFT data include the ScienceQA training split (Table~\ref{tab:sft-data}).}
\label{tab:understanding-other}
\footnotesize\setlength{\tabcolsep}{4pt}
\begin{tabularx}{\linewidth}{@{}llRR@{}}
\toprule
Model & Size & OK-VQA & SQA-IMG\\
\midrule
D-DiT ($256^2$)~\citep{dualdiffusion} & 2.1B & 28.5 & --\\
D-DiT ($512^2$)~\citep{dualdiffusion} & 2.1B & 25.3 & --\\
LLaVA-1.5~\citep{llava15} & 7B & -- & 66.8\\
Qwen-VL-Chat~\citep{qwenvl} & 7B & 56.6 & 68.2\\
\midrule
\rowcolor{paperblue!6}JUICE (SigLIP2) & 2.1B & 33.9 & 80.6\\
\rowcolor{paperblue!6}JUICE (DINO) & 2.1B & 28.7 & 73.7\\
\bottomrule
\end{tabularx}
\end{table}

\section{Joint reasoning}
\label{app:maze}
\begingroup

\subsection{Mazes: dataset, adaptation, and evaluation}
\label{app:joint-reasoning}

\paragraph{Dataset construction.}
We construct a variant of the maze task with multiple solutions~\citep{ivanitskiy2023configurable} to evaluate the correctness and consistency of jointly generated image and text outputs.
Starting from grid mazes generated by randomized depth first search, we remove additional walls at dead ends to introduce cycles.
We sample distinct start and goal cells and retain puzzles with between two and sixteen shortest routes.
For each training example, we uniformly select one shortest route and use it to construct both targets: an image with the route painted blue and its corresponding sequence of moves from $\{\mathrm{U},\mathrm{D},\mathrm{L},\mathrm{R}\}$.
Thus, each training pair is consistent, while a given puzzle admits multiple valid output pairs. The model receives an unsolved maze image together with a text question that gives the grid size and the start and goal coordinates. Images are $224{\times}224$, and each maze block lines up with one $14\times14$ image patch. Walls, background, and the start and goal markers are part of the input, and the model draws the route. There are 200{,}000 training examples with grids of $5\times5$, $6\times6$, and $7\times7$. We test on 500 puzzles (167, 167, and 166 of the three sizes) whose layouts do not appear in the training set, and at evaluation we accept any valid shortest route, not only the route stored as the target.

\paragraph{Training.}
We initialize the 2.1B models from the raw weights of the SFT stage (step 15{,}000; Table~\ref{tab:training-stages}) and adapt them for 5{,}000 updates.
We use AdamW with a global batch size of 1{,}024, a learning rate of $10^{-4}$, 250 warmup steps, and weight decay $0.01$.
The joint adaptation uses 50\% joint generation examples, 25\% image generation examples conditioned on text, and 25\% text generation examples conditioned on images (4, 2, and 2 of every 8 rows in a microbatch, from nominal weights 0.6, 0.2, and 0.2). The Image FT only and Text FT only models are finetuned (FT) on the image or the text target alone; Table~\ref{tab:maze-specialists} reports the validity of that modality.

\paragraph{Inference and evaluation.}
At inference, the given image patches and question tokens remain fixed, while the image target and answer positions are initialized from noise.
Coupled sampling evaluates both velocity fields from the same current state of image and text and advances both generated streams at each Euler step. For sequential sampling, one sampler first generates the solved image and then runs image-to-text generation to get the text solution, while the other generates text first and then runs text-to-image generation. In all samplers, the linear task head of the mazes reads the moves from the generated text states (Appendix~\ref{app:text-decoder}).

We use three measures. Image validity (I) checks that the decoded route is a legal shortest path from start to goal, and text validity (T) applies the same check to the move sequence. Joint success (J) further asks that both outputs show the same route. If the image and the moves each give a valid but different shortest route, the sample fails.

\endgroup

\FloatBarrier
\subsection{Maze sampling budgets}
\label{app:maze-sampling}

One backbone evaluation is one forward pass of MM-RiT; calls to the encoders and decoders are not counted. Without guidance, coupled sampling gets both velocity fields from the shared state in a single forward pass and then advances both streams. Independent sampling advances each stream on its own, with the other stream left at its initial noise. Sequential sampling completes the first stream, then uses its clean output as the condition for the second. Table~\ref{tab:maze-call-budgets} gives the steps per stream at 16 and at 8 calls, together with joint success at 8 calls. Halving the budget changes coupled sampling by at most 0.3 points and leaves the order of the samplers unchanged. Results are means over two sampling seeds; Table~\ref{tab:maze-seeds} lists each seed at 16 calls. On the same 500 mazes, paired by maze over both seeds, coupled sampling alone solves 10 mazes and image first alone 25 for SigLIP2 (exact McNemar test, $p{=}0.017$), and 19 against 29 for DINO ($p{=}0.19$). For SigLIP2, image first is therefore better than coupled sampling, while for DINO the two do not differ significantly.

\begin{table}[!htbp]
\centering\small
\caption{\textbf{Joint success of each sampling seed on mazes} (16 backbone calls, 500 mazes).}
\label{tab:maze-seeds}
\begin{tabular}{@{}lcccc@{}}
\toprule
& \multicolumn{2}{c}{SigLIP2} & \multicolumn{2}{c}{DINO}\\
\cmidrule(lr){2-3}\cmidrule(l){4-5}
Sampler & Seed 1 & Seed 2 & Seed 1 & Seed 2\\
\midrule
Coupled & 95.2 & 92.2 & 82.2 & 81.2\\
Image first & 95.8 & 94.6 & 83.4 & 82.0\\
Text first & 84.8 & 85.8 & 59.4 & 60.4\\
Independent (8 steps per stream) & 56.0 & 53.2 & 42.6 & 42.8\\
\bottomrule
\end{tabular}
\end{table}

\begin{table}[!htbp]
\centering\small
\caption{\textbf{Maze samplers at 16 and at 8 backbone calls.} A coupled step advances both streams with one call; independent and sequential sampling spend one call per stream per step. J at 16 calls is in Table~\ref{tab:maze-specialists}; J at 8 calls is the mean of two sampling seeds.}
\label{tab:maze-call-budgets}
\setlength{\tabcolsep}{4pt}\renewcommand{\arraystretch}{1.10}
\begin{tabular*}{\linewidth}{@{\extracolsep{\fill}}lccrr@{}}
\toprule
& & & \multicolumn{2}{c}{J at 8 calls}\\
\cmidrule(l){4-5}
Sampler & Steps at 16 calls & Steps at 8 calls & SigLIP2 & DINO\\
\midrule
Coupled & 16 shared & 8 shared & 93.7 & 81.4\\
Image first & 8 + 8 & 4 + 4 & \textbf{94.9} & \textbf{83.0}\\
Text first & 8 + 8 & 4 + 4 & 84.9 & 61.6\\
Independent & 8 per stream & 4 per stream & 55.9 & 45.7\\
\bottomrule
\end{tabular*}
\end{table}

\FloatBarrier

\FloatBarrier
\subsection{Maze trajectory case studies}
\label{app:maze-trajectories}

Figure~\ref{fig:maze-cases} shows mazes 37 and 61, in which coupled sampling reaches agreement while independent sampling ends with two valid but different routes. In maze 37 (Figure~\ref{fig:maze-cases}a), the moves are fixed from the first step and the drawn route changes to their route at step 3; in maze 61 (Figure~\ref{fig:maze-cases}b), the drawn route is fixed from step 2 and the moves change to it at step 7. Both directions also occur in other mazes of the same run: the drawn route changes in maze 208, and the moves change in maze 23. Their independent counterparts again end with two valid but different routes.

\FloatBarrier
\subsection{Visual spatial planning}
\label{app:vsp}
VSP~\citep{vsp} poses navigation problems on FrozenLake grids. The input is an image of the map with the start, the goal, and holes, and the answer is a sequence of moves (U, D, L, R) that reaches the goal without stepping into a hole; any such sequence counts as a success, and moves off the grid leave the player in place. Levels 3 to 8 correspond to $3{\times}3$ to $8{\times}8$ grids, with 100 test maps per level. As for the mazes, \method{} generates the map with the path drawn on it and the move sequence together, with coupled sampling, 16 steps, and no guidance, and a linear task head reads the moves (Appendix~\ref{app:text-decoder}). We adapt the finetuned SigLIP2 model for 5,000 updates (Table~\ref{tab:training-stages}) on maps of sizes $3{\times}3$ to $6{\times}6$, so levels 7 and 8 use map sizes that do not appear in training.

The training set has 55,939 maps. We generate FrozenLake maps with random start and goal positions and a hole probability drawn from $\mathcal U(0,0.35)$, keep only solvable maps, and use the shortest path found by breadth first search as the target route, breaking ties in the order L, U, D, R. After removing duplicates and every map that appears in the VSP benchmark, 50,000 generated maps remain. The other 5,939 maps are the 6,000 FrozenLake problems of ThinkMorph~\citep{thinkmorph} without the 61 that coincide with benchmark maps. All maps are rendered again on a $224{\times}224$ canvas in which each cell covers $2\times2$ SigLIP2 patches, and the target image draws the route in red. Adaptation uses the optimizer settings and task ratios of the maze recipe (Appendix~\ref{app:joint-reasoning}).

We score the move sequences with the official evaluation script on the 100 benchmark maps of each level. The main metric averages levels 3 to 6, the sizes seen in training, and levels 7 and 8 measure extrapolation to larger maps. The official rule checks only the move sequence; on 500 maps of our own held out from training, we also check that the drawn path and the moves agree (joint consistency, 99.6\%).

On the benchmark maps of levels 3 to 8, the drawn path is a shortest route on 98, 100, 100, 99, 99, and 96 of the 100 maps, and the moves reach the goal on 100, 100, 100, 99, 94, and 69. We split each reference route into runs of the same repeated move. On levels 7 and 8, the generated moves drop exactly one move from 2 of the 238 runs of length one, from 2 of 117 runs of length two, 4 of 81 of length three, 7 of 45 of length four, 7 of 23 of length five, and 7 of 9 of length six. On levels 3 to 6, no move is dropped from any of the 810 runs.

The baseline values in Table~\ref{tab:vsp} are taken from LatentUM~\citep{latentum}. InternVL3.5-38B~\citep{internvl35} is evaluated without training for the task, using the official VSP prompt. ThinkMorph~\citep{thinkmorph} is finetuned from BAGEL-7B on interleaved reasoning data that include 6,000 FrozenLake problems with grids of size 3 to 6. LatentUM is finetuned on data in the style of VSP: its coarse variant on the 6,000 ThinkMorph problems, and its fine variant on 18,000 problems solved step by step that interleave single moves with updated maps. LatentUM reports InternVL3.5 and ThinkMorph only as averages, and ThinkMorph does not state which levels its evaluation covers.

\clearpage
\section{Qualitative examples}
\label{app:qualitative}
\label{app:more-qualitative}

\begin{figure}[!htbp]
\centering
\includegraphics[width=\linewidth]{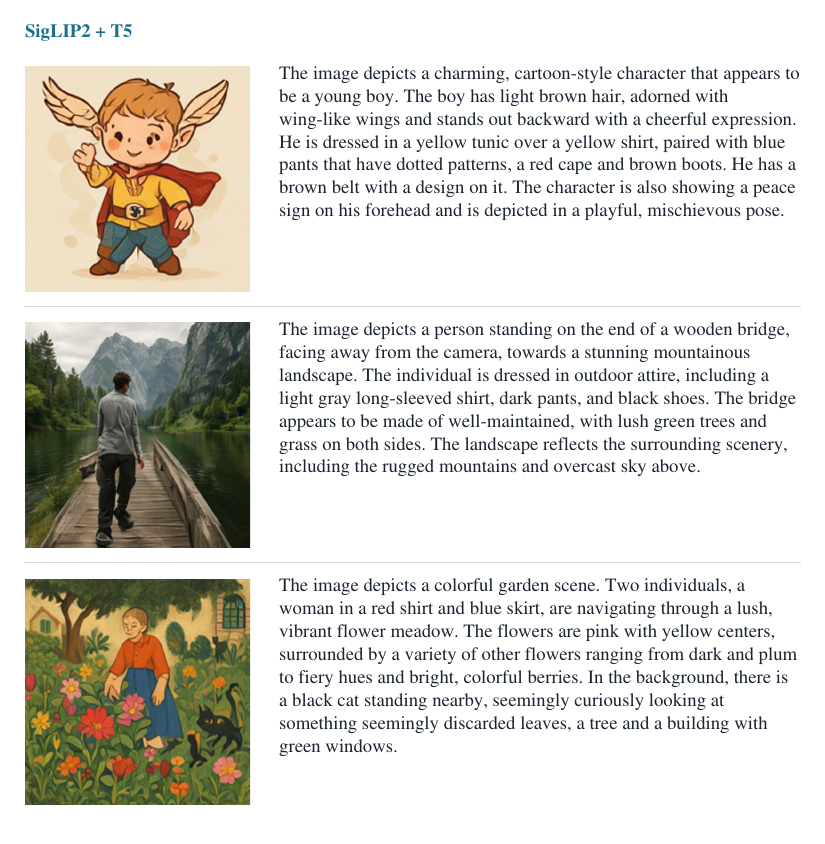}
\caption{\textbf{Joint generation of images and text with SigLIP2+T5.} Three generated pairs depict a winged cartoon boy, a person on a lakeside pier, and a woman in a flower garden.}
\label{fig:qual-joint-1}
\end{figure}

\begin{figure}[!htbp]
\centering
\includegraphics[width=\linewidth]{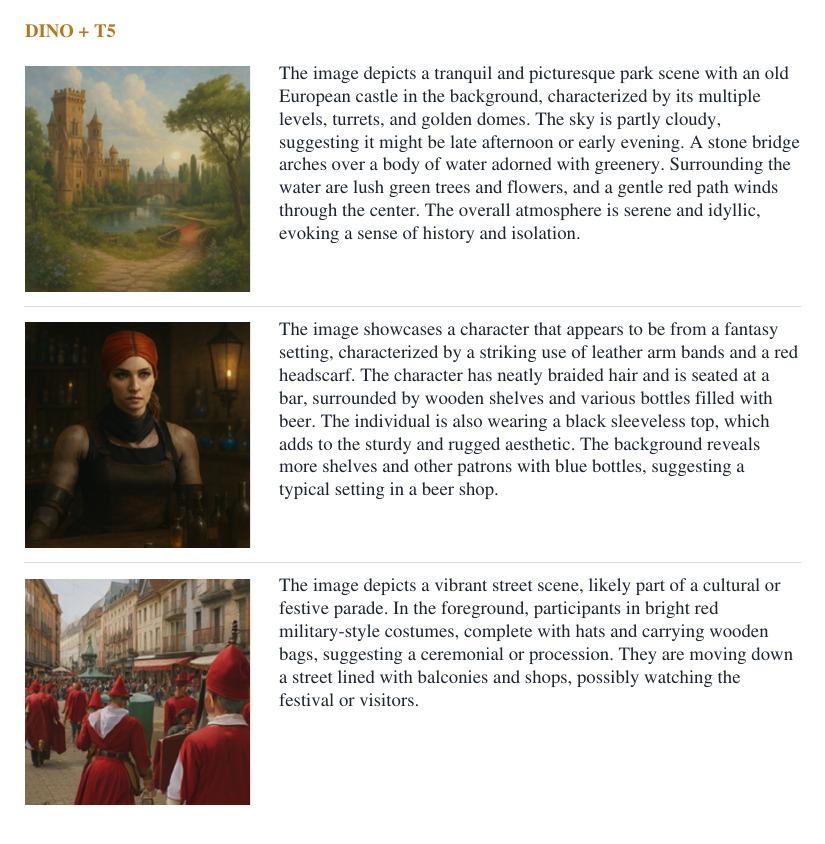}
\caption{\textbf{Joint generation of images and text with DINO+T5.} Three generated pairs depict a castle in a park, a woman at a bar, and a street parade.}
\label{fig:qual-joint-2}
\end{figure}

\begin{figure}[!htbp]
\centering
\includegraphics[width=\linewidth,height=.70\textheight,keepaspectratio]{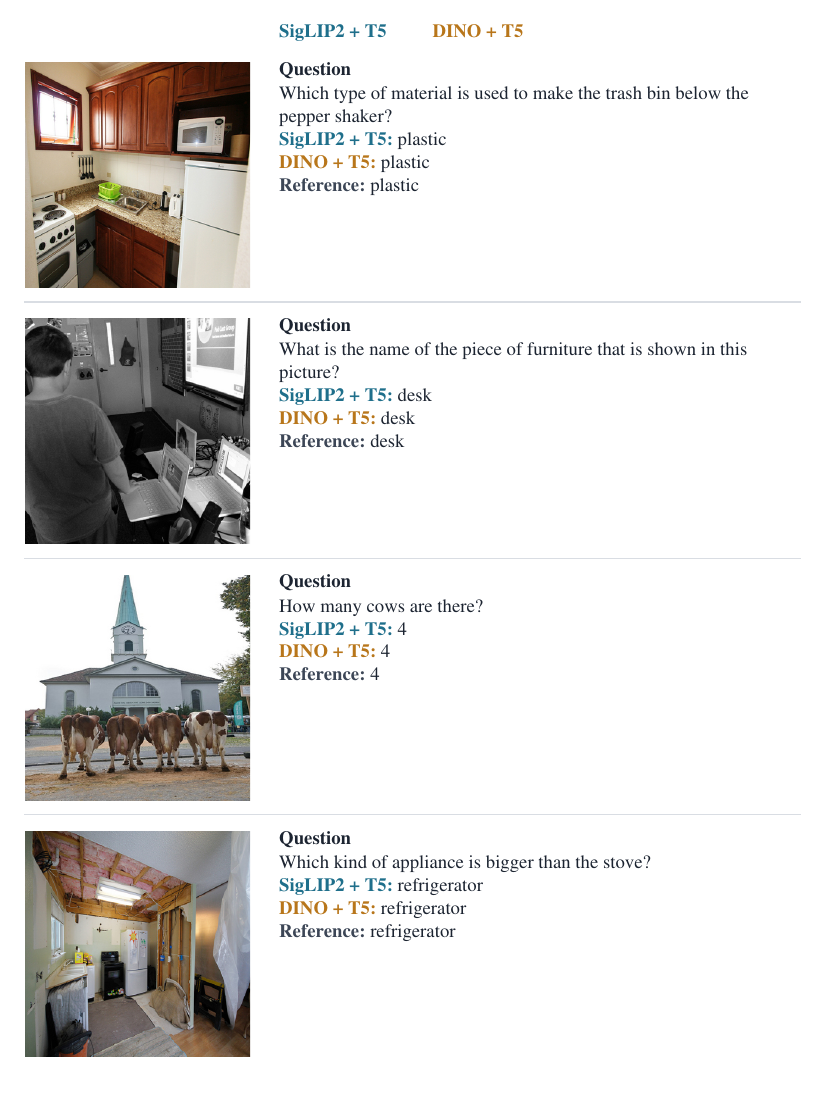}
\caption{\textbf{Visual question answering: attributes and quantities.} Questions concern material, furniture identity, counting, and relative appliance size.}
\label{fig:qual-vqa-1}
\end{figure}

\begin{figure}[!htbp]
\centering
\includegraphics[width=\linewidth,height=.85\textheight,keepaspectratio]{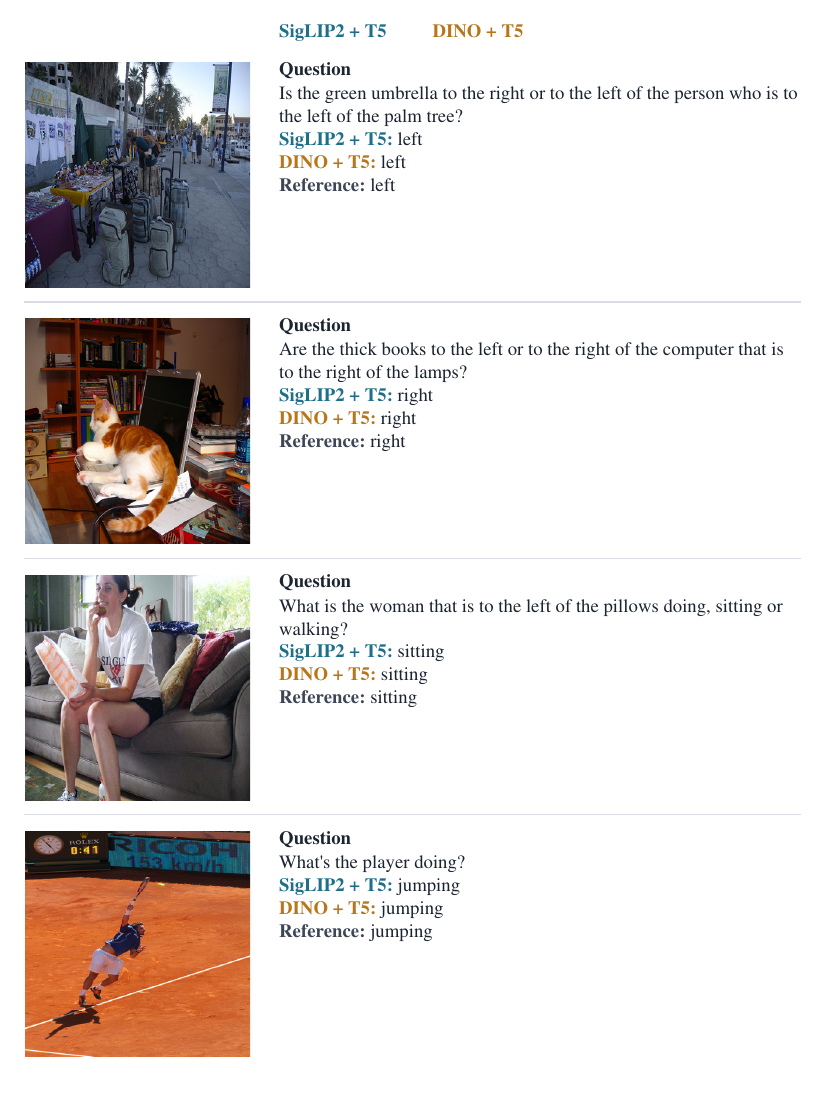}
\caption{\textbf{Visual question answering: spatial relations and actions.} Questions concern left/right relations and the activities of people in the image.}
\label{fig:qual-vqa-2}
\end{figure}

\begin{figure}[!htbp]
\centering
\includegraphics[width=\linewidth]{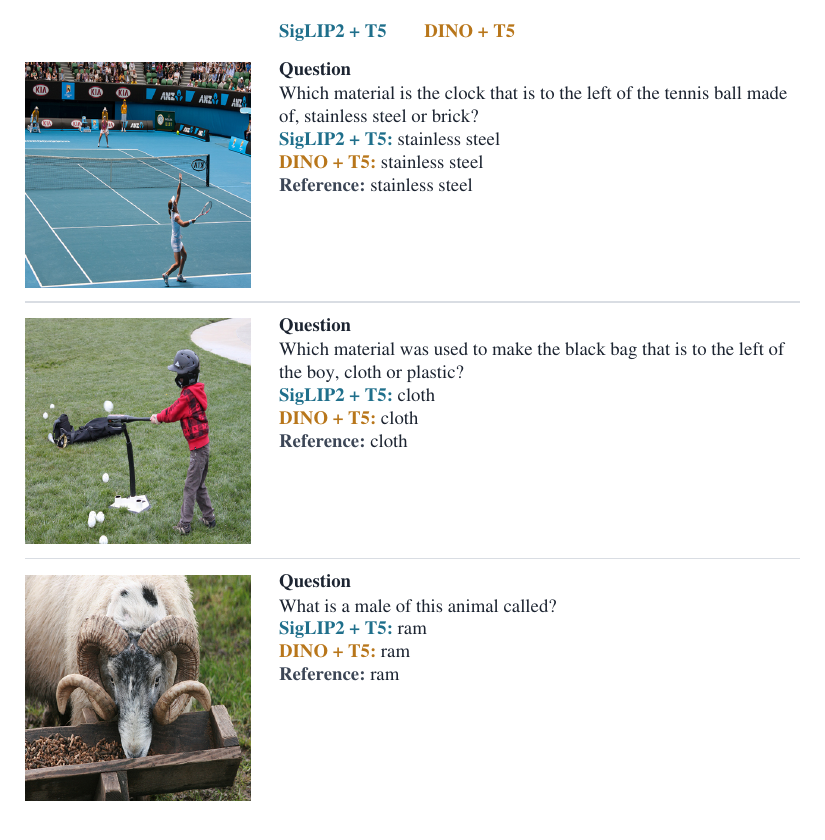}
\caption{\textbf{Visual question answering: compositional and outside knowledge queries.} Questions identify object materials using relational descriptions and ask for outside knowledge about the animal shown.}
\label{fig:qual-vqa-3}
\end{figure}

\begin{figure}[!htbp]
\centering
\includegraphics[width=\linewidth]{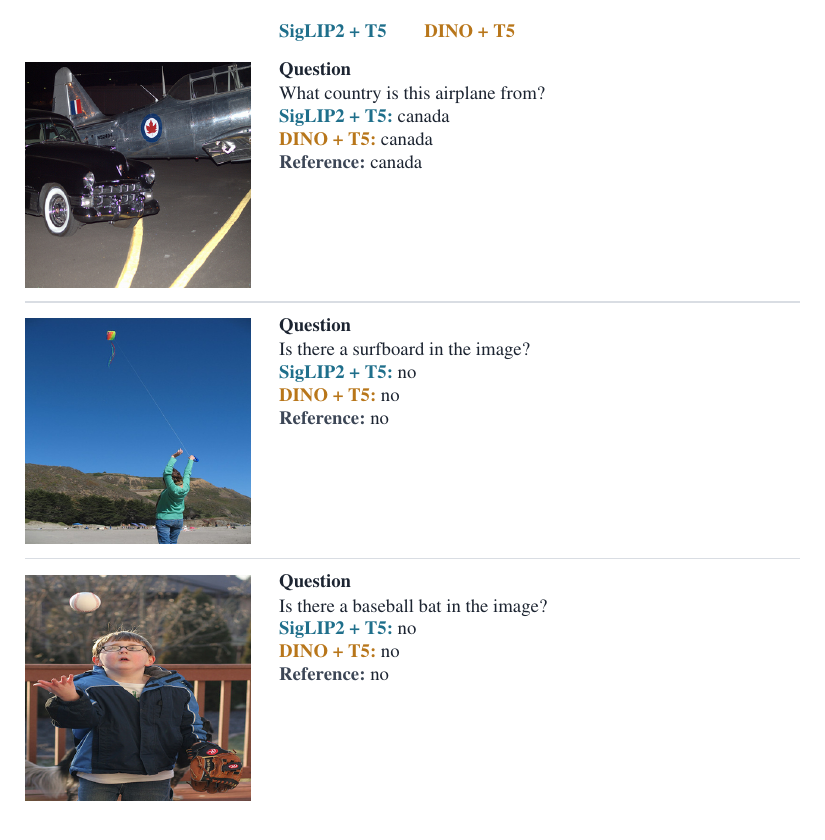}
\caption{\textbf{Visual question answering: outside knowledge and object existence.} Questions ask where an airplane comes from and whether specified objects are present.}
\label{fig:qual-vqa-4}
\end{figure}

We present selected qualitative examples from five generation and understanding settings, complementing the aggregate evaluations in Section~\ref{sec:unified}. Figure~\ref{fig:qual-understanding-main} in the main text draws its questions from Figures~\ref{fig:qual-vqa-1} to~\ref{fig:qual-vqa-4}, and Figure~\ref{fig:qual-joint-main} shows two pairs each from Figures~\ref{fig:qual-joint-1} and~\ref{fig:qual-joint-2}.

\subsection{Image generation conditioned on text}
\label{app:qualitative-generation}

Each pair uses the same prompt for SigLIP2+T5 and DINO+T5 and guidance $w{=}5$ with the generic caption reference, with outputs selected independently from the available samples; some of these prompts also appear in Figure~\ref{fig:qualitative-gen}, which shows further prompts.

\begin{figure}[!htbp]
\centering
\includegraphics[width=.9\linewidth]{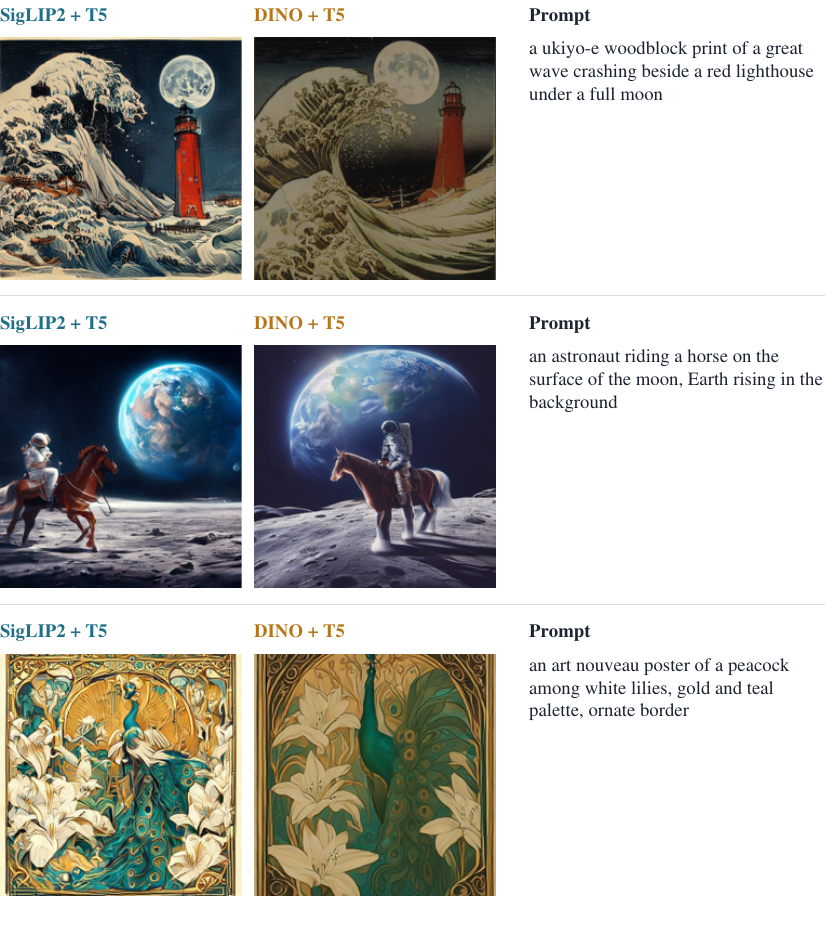}
\caption{\textbf{Image generation conditioned on text: illustration and imagined scenes.} Outputs for the same prompts show a wave beside a lighthouse, an astronaut riding a horse on the Moon, and an art nouveau peacock poster.}
\label{fig:qual-t2i-1}
\end{figure}
\clearpage

\begin{figure}[!htbp]
\centering
\includegraphics[width=.9\linewidth]{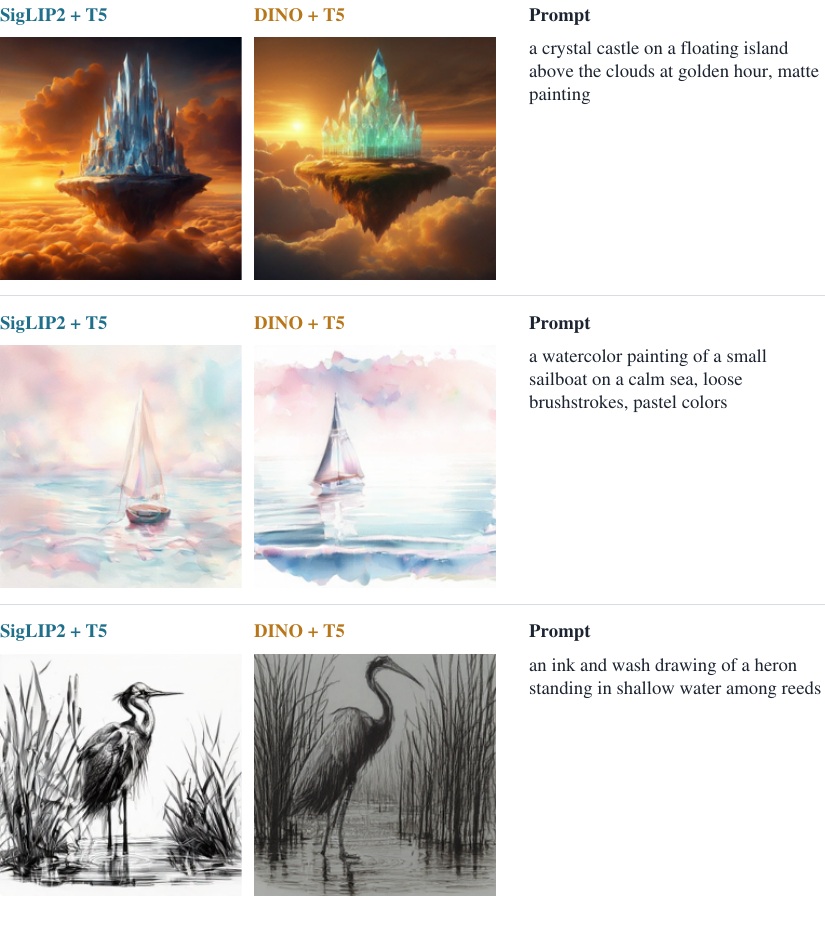}
\caption{\textbf{Image generation conditioned on text: varied visual styles.} The prompts specify a fantasy matte painting, a watercolor sailboat, and an ink wash heron. Complete prompts accompany each pair.}
\label{fig:qual-t2i-2}
\end{figure}
\clearpage

\begin{figure}[!htbp]
\centering
\includegraphics[width=.9\linewidth]{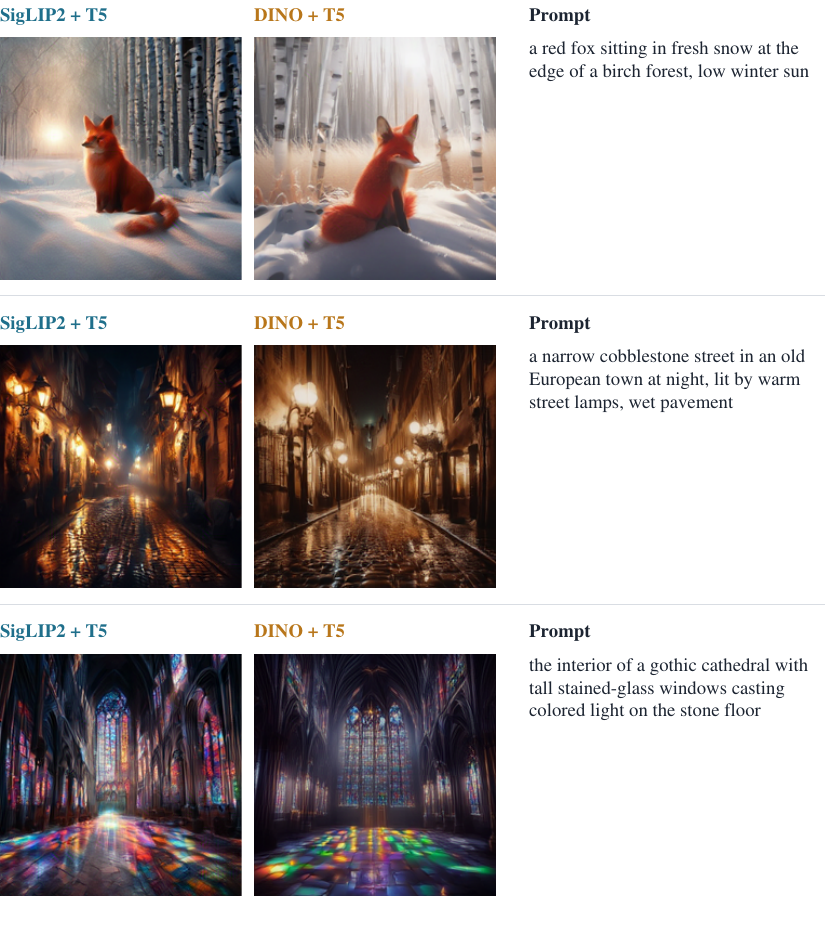}
\caption{\textbf{Image generation conditioned on text: natural and architectural scenes.} A fox in a snowy forest, a lamplit street, and a cathedral interior illustrate differences in composition, lighting, and texture under shared prompts.}
\label{fig:qual-t2i-3}
\end{figure}
\clearpage

\subsection{Image captioning}
\label{app:qualitative-captioning}

Each input image is shown with the caption of each model, and both models get the prompt ``Describe the image concisely.'' In these examples, the two encoder configurations mention different entities, attributes, and scene details.

\begin{figure}[!htbp]
\centering
\includegraphics[width=\linewidth]{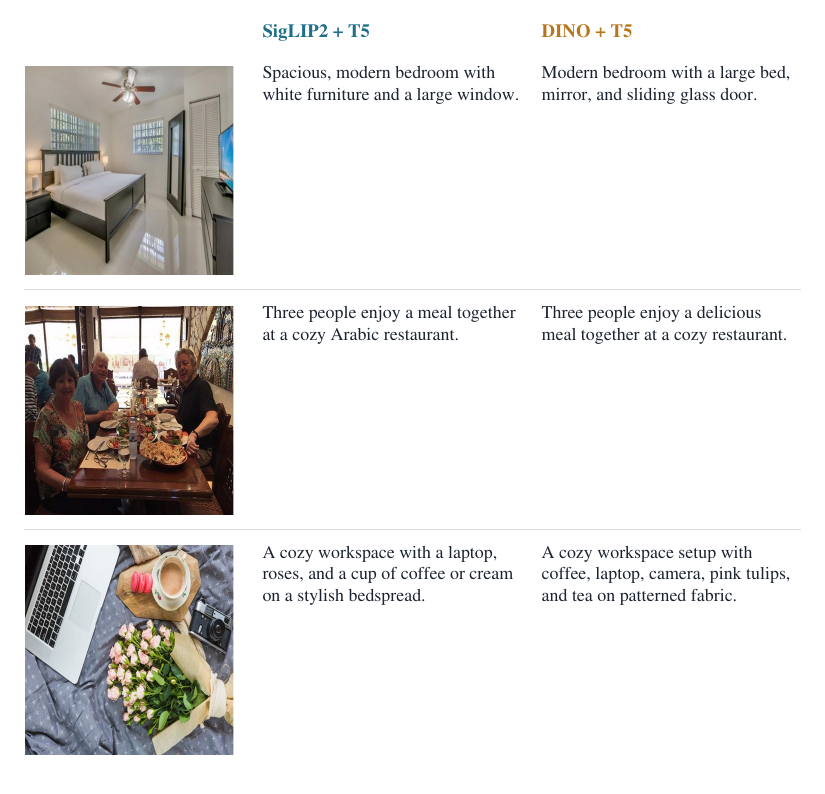}
\caption{\textbf{Image captioning with shared visual inputs.} SigLIP2+T5 and DINO+T5 describe a bedroom, a restaurant meal, and a workspace.}
\label{fig:qual-caption-1}
\end{figure}
\clearpage

\begin{figure}[!htbp]
\centering
\includegraphics[width=\linewidth]{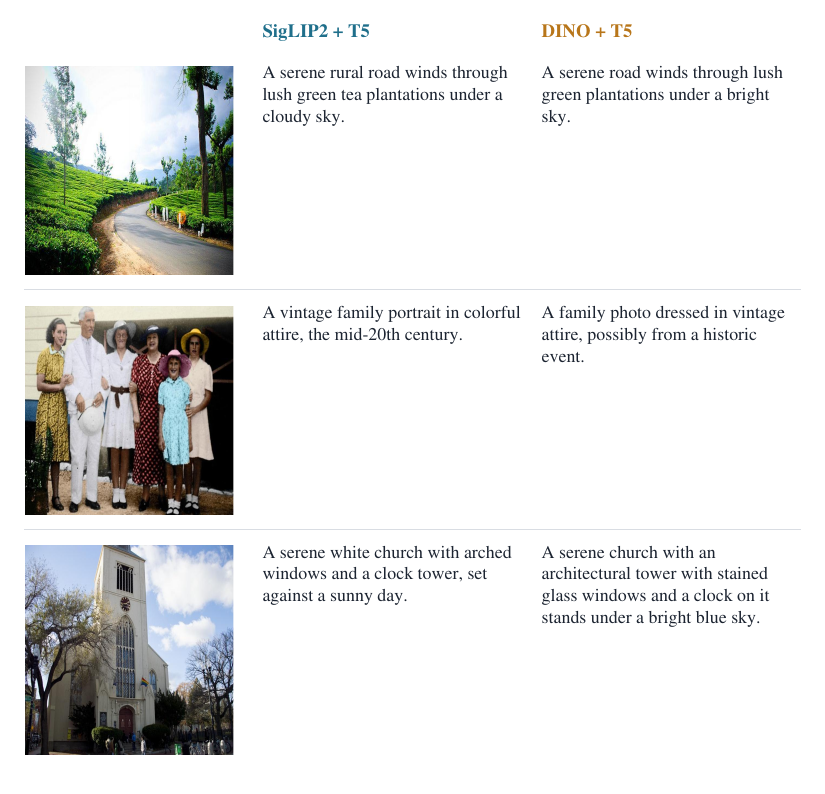}
\caption{\textbf{Image captioning with shared visual inputs (continued).} Both models describe a road through tea plantations, a vintage family portrait, and a church with a clock tower.}
\label{fig:qual-caption-2}
\end{figure}
\clearpage

\subsection{Joint generation}
\label{app:qualitative-joint}

\begin{figure}[!htbp]
\centering
\includegraphics[width=\linewidth,height=.80\textheight,keepaspectratio]{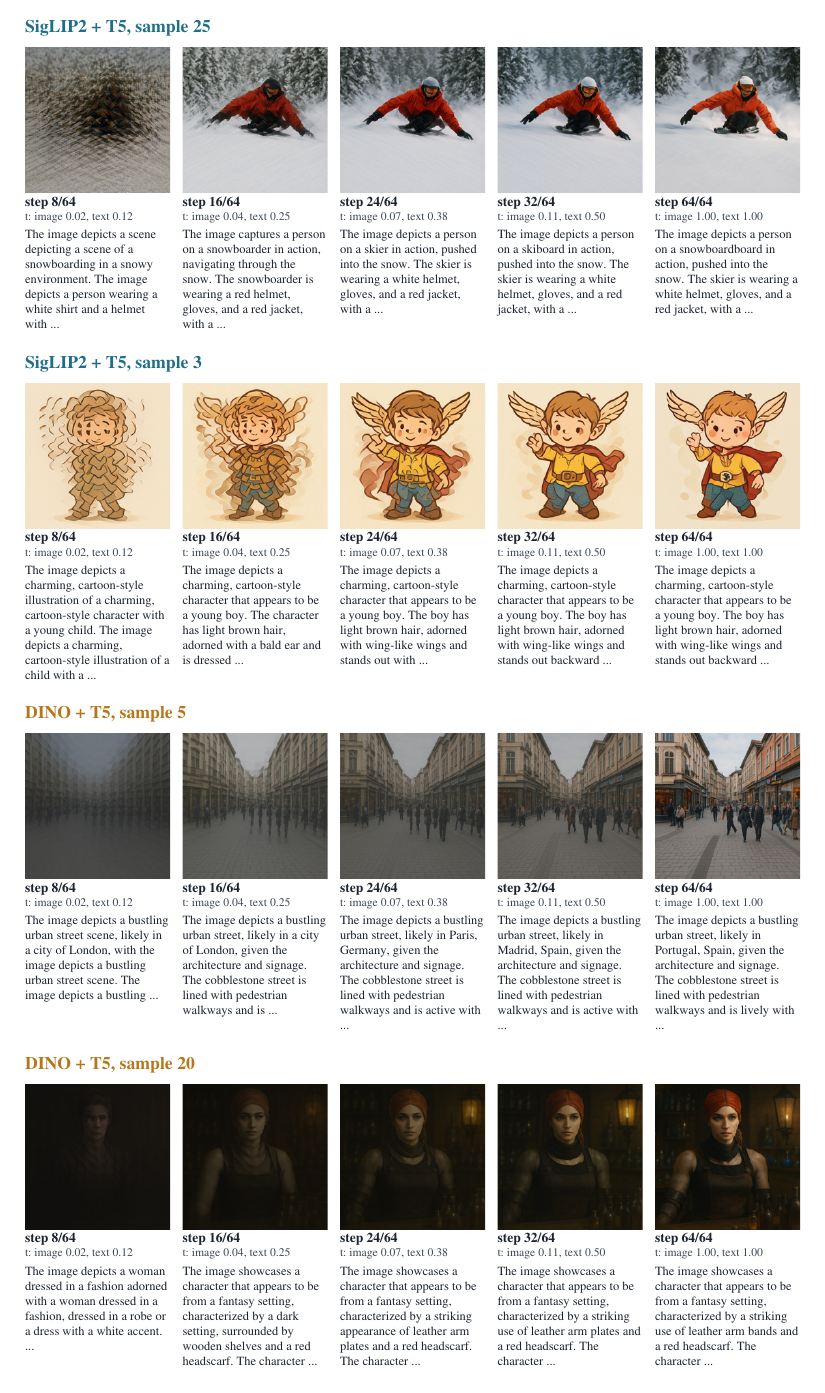}
\caption{\textbf{How a joint sample forms.} For four unconditional joint samples (64 steps), each column shows, at one step, the model's current estimate of the clean image (decoded) and of the clean text (read out with the autoregressive caption decoder, first 28 words). The two streams follow different time grids, given under each step, and the text is further along at a given step: in SigLIP2 sample 25 it already names a snowboarding scene while the image has not yet formed. Later steps revise both streams; in DINO sample 5 the guessed location moves from London to Paris, Madrid, and Portugal as the facades sharpen. The second and fourth rows end in samples shown in Figure~\ref{fig:qual-joint-main}.}
\label{fig:qual-joint-trajectory}
\end{figure}

Both image and text streams start from noise, producing paired visual and textual outputs without a text prompt. We show three independently sampled pairs from each encoder configuration, drawn from the finetuned checkpoints. Text is read out with the autoregressive caption decoder (Appendix~\ref{app:text-decoder}) and shown up to the point where it starts to repeat. Figure~\ref{fig:qual-joint-trajectory} shows how four samples form during sampling.

\clearpage

\clearpage

\clearpage

\subsection{Captioning with long answers}
\label{app:qualitative-long-captioning}

For each image, both models receive the prompt ``Describe this image in detail.'' The first text column is SigLIP2+T5 and the second is DINO+T5.

\begin{figure}[!htbp]
\centering
\includegraphics[width=\linewidth,height=.76\textheight,keepaspectratio]{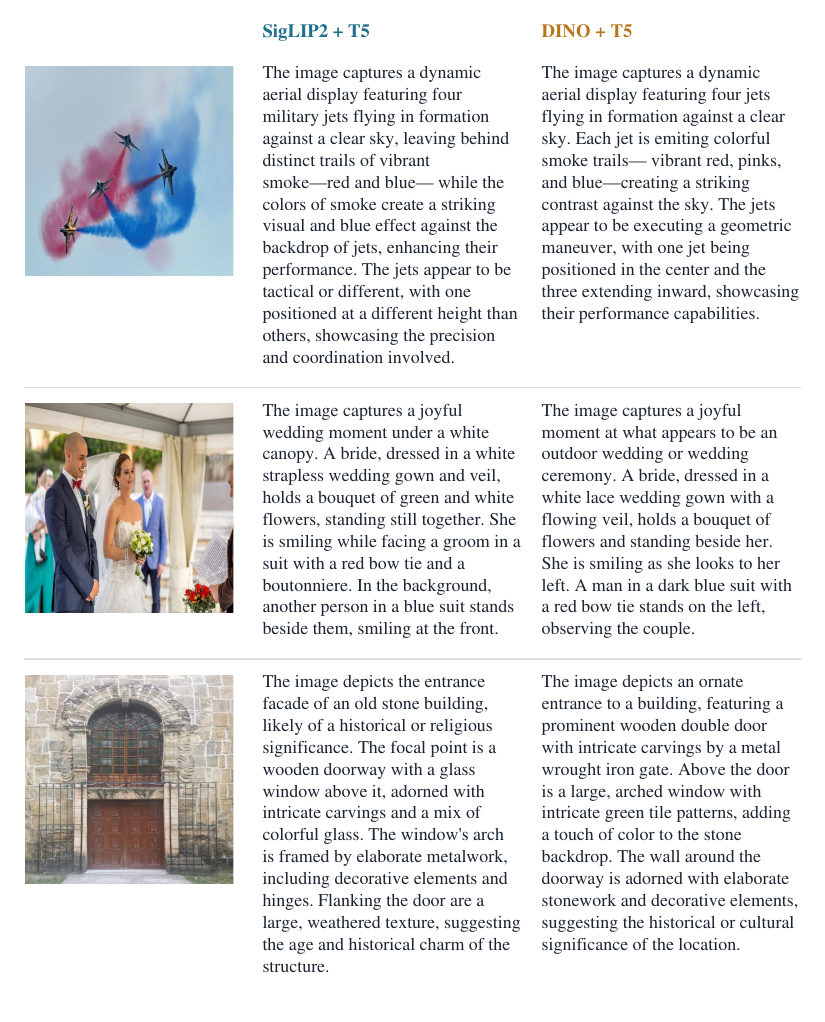}
\caption{\textbf{Image captions with long answers.} Detailed descriptions of jets trailing colored smoke, a wedding under a canopy, and a stone doorway.}
\label{fig:qual-long-1}
\end{figure}
\clearpage

\begin{figure}[!htbp]
\centering
\includegraphics[width=\linewidth]{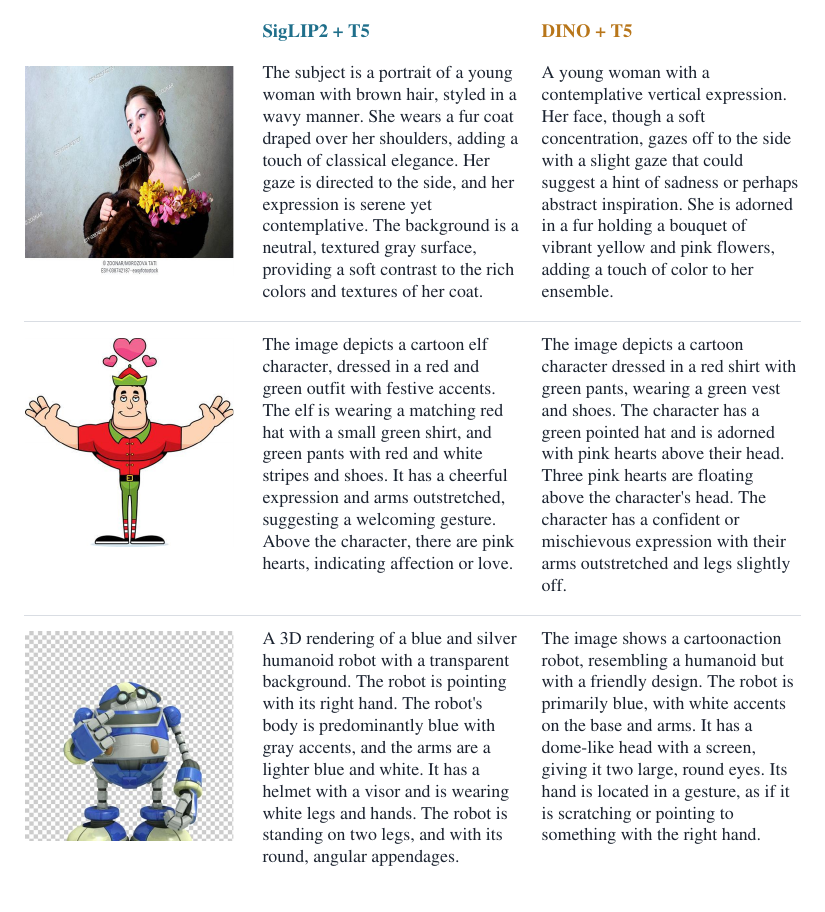}
\caption{\textbf{Image captions with long answers (continued).} Detailed descriptions of a portrait in a fur coat, a cartoon elf, and a 3D robot.}
\label{fig:qual-long-2}
\end{figure}
\clearpage

\begin{figure}[!htbp]
\centering
\includegraphics[width=\linewidth]{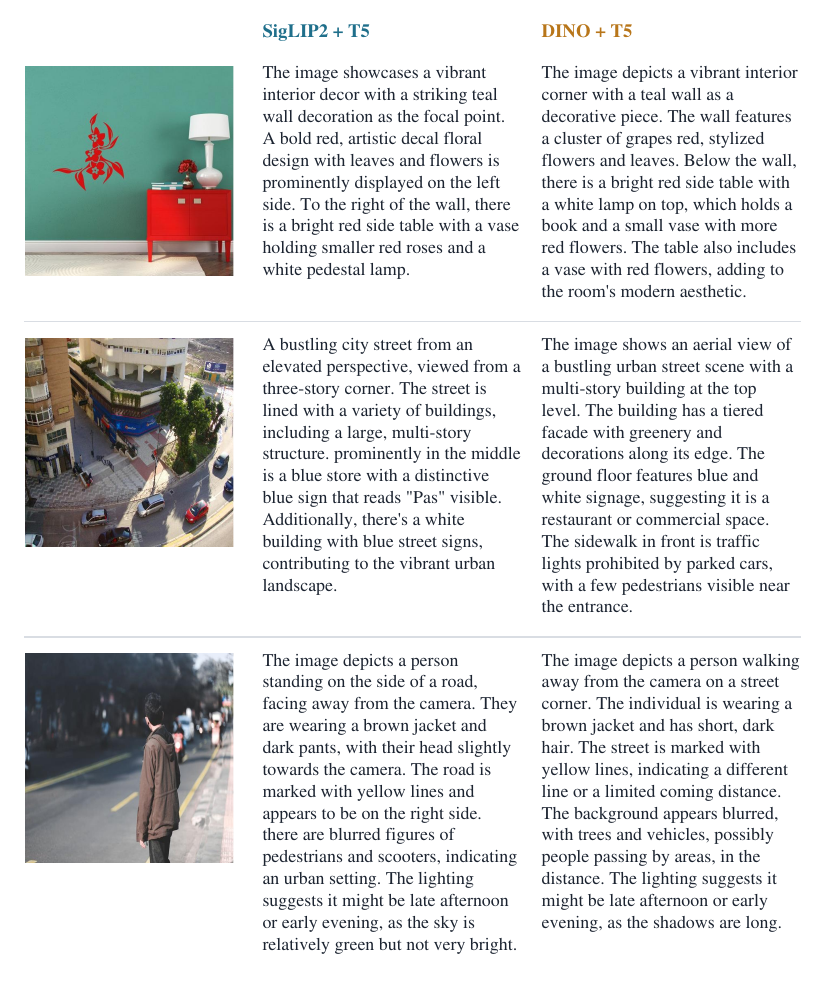}
\caption{\textbf{Image captions with long answers (continued).} Detailed descriptions of a wall with a red side table, a street seen from above, and a person beside a road.}
\label{fig:qual-long-3}
\end{figure}
\clearpage

\begin{figure}[!htbp]
\centering
\includegraphics[width=\linewidth]{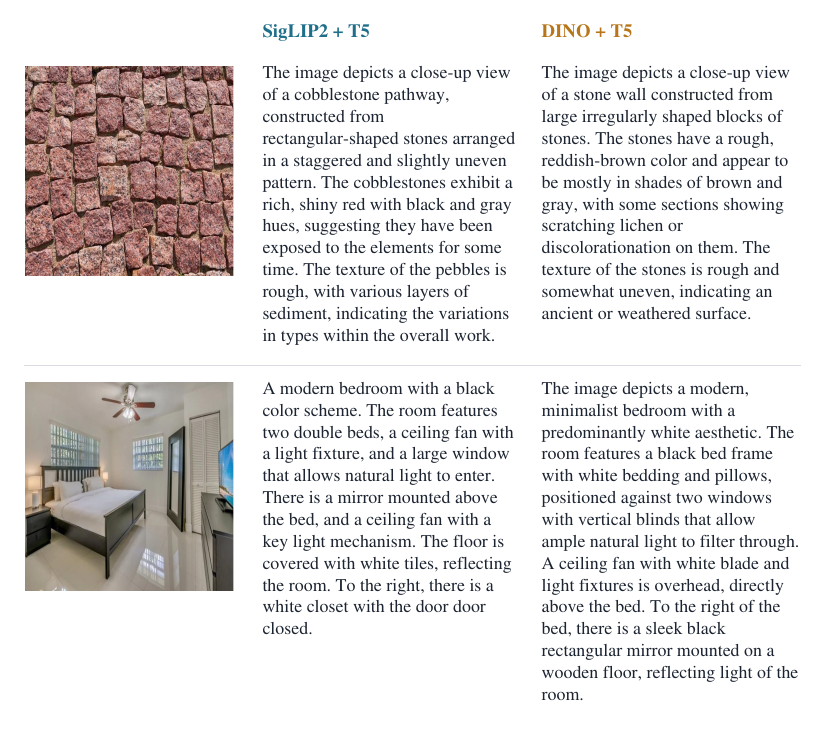}
\caption{\textbf{Image captions with long answers (continued).} Detailed descriptions of a red stone texture and a bedroom. Each row compares the two models on the same image and prompt.}
\label{fig:qual-long-4}
\end{figure}
\clearpage

\subsection{Visual question answering}
\label{app:qualitative-vqa}

Each case includes the input image, question, and both model answers. Reference answers are included where provided. The selected examples span attributes, object size and count, spatial relations, actions, compositional queries, outside knowledge, and object existence.

\clearpage

\clearpage

\clearpage

\clearpage

\begin{figure}[!htbp]
\centering
\includegraphics[width=\linewidth,height=.85\textheight,keepaspectratio]{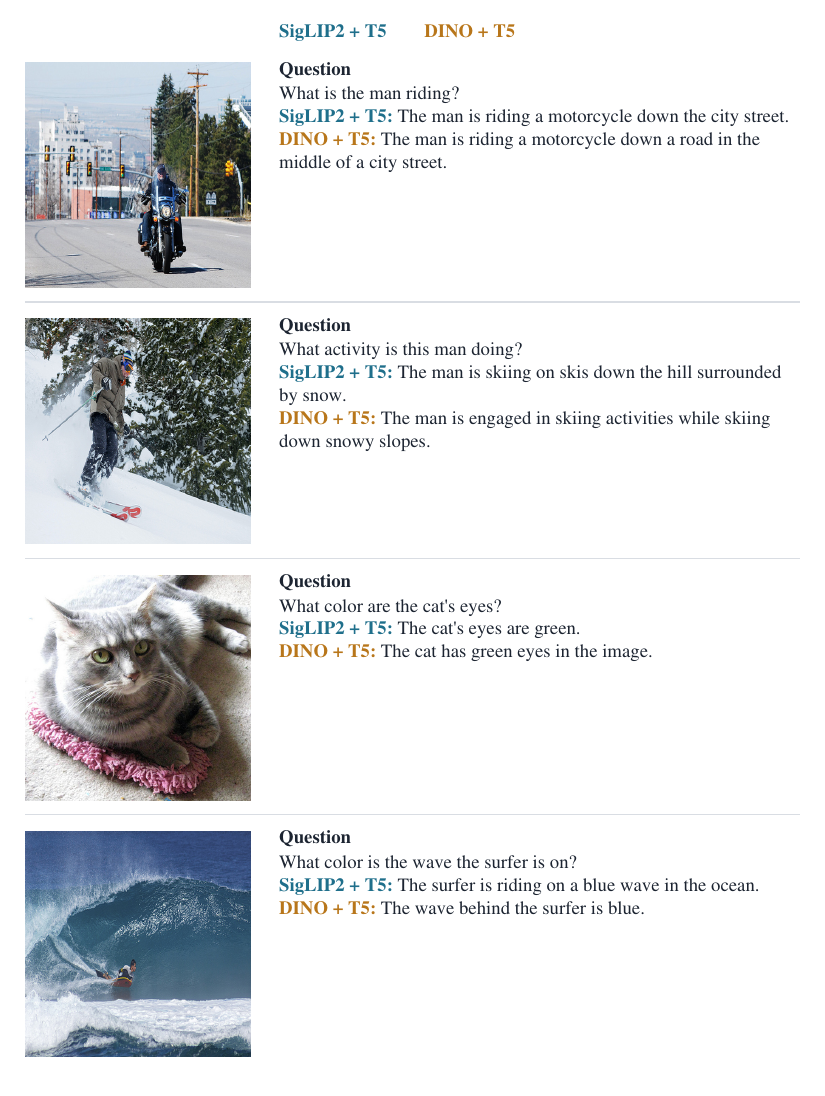}
\caption{\textbf{Visual question answering: additional object, action, and color examples.} Questions ask what a man is riding, what a man is doing, and the color of a cat's eyes and of a wave, without the instruction to answer briefly. Model responses retain their original wording.}
\label{fig:qual-vqa-5}
\end{figure}
\clearpage

\begin{figure}[!htbp]
\centering
\includegraphics[width=\linewidth,height=.85\textheight,keepaspectratio]{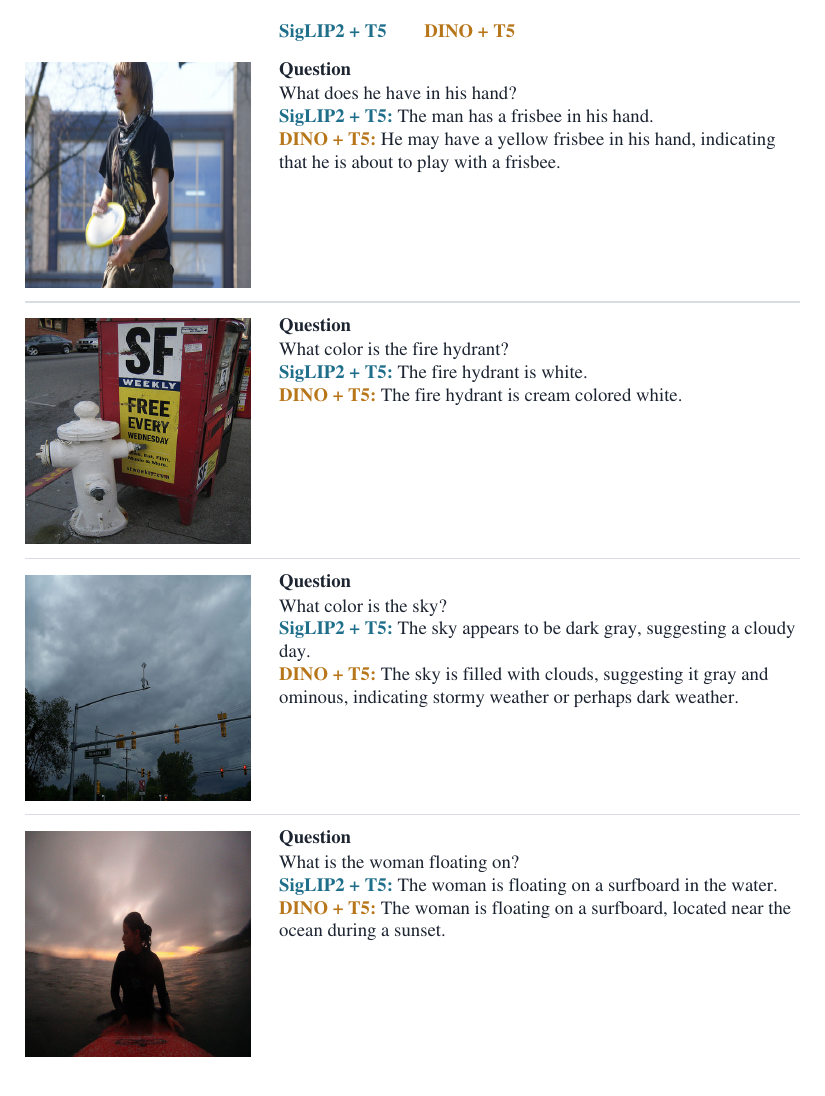}
\caption{\textbf{Visual question answering: additional object and color examples.} Questions ask what a man holds in his hand, the color of a fire hydrant and of the sky, and what a woman is floating on. Model responses retain their original wording.}
\label{fig:qual-vqa-6}
\end{figure}

\end{document}